\documentclass[pdflatex,sn-nature]{sn-jnl}

\usepackage[acronym,style=super]{glossaries}
\usepackage{graphicx}%
\usepackage{multirow}%
\usepackage{amsmath,amssymb,amsfonts}%
\usepackage{amsthm}%
\usepackage{mathrsfs}%
\usepackage[title]{appendix}%
\usepackage{xcolor}%
\usepackage{textcomp}%
\usepackage{manyfoot}%
\usepackage{booktabs}%
\usepackage{algorithm}%
\usepackage{algorithmicx}%
\usepackage{algpseudocode}%
\usepackage{listings}%
\usepackage{rotating}
\usepackage{lscape}
\usepackage{subcaption}
\usepackage{siunitx}
\usepackage{booktabs}
\usepackage{array}

\theoremstyle{thmstyleone}%
\theoremstyle{thmstyletwo}%

\theoremstyle{thmstylethree}%

\newacronym{cur}{CUR}{circadian and ultradian rhythms}
\newacronym{svm}{SVM}{support vector machines}
\newacronym{rfe}{RFE}{recursive feature elimination}
\newacronym{seid}{SeiD}{seizure detection}
\newacronym{edf}{EDF}{European Data Format}
\newacronym{bdf}{BDF}{BioSemi Data Format}
\newacronym{esd}{ESD}{emotion state detection}
\newacronym{ssd}{SSD}{sleep state detection}
\newacronym{sudep}{SUDEP}{sudden unexpected death in epilepsy}
\newacronym{ml}{ML}{machine learning}
\newacronym{dl}{DL}{deep learning}
\newacronym{sl}{SL}{shallow learning}
\newacronym{sgd}{SGD}{stochastic gradient descent}
\newacronym{tusz}{TUSZ}{Temple University Hospital EEG Seizure Corpus}
\newacronym{chb-mit}{CHB-MIT}{Children's Hospital Boston - Massachusetts Institute of Technology Scalp EEG Database}
\newacronym{chb}{CHB}{Children’s Hospital Boston}
\newacronym{mit}{MIT}{Massachusetts Institute of Technology}
\newacronym{deap}{DEAP}{dataset for emotion analysis using EEG, physiological and video signals}
\newacronym{sleep-edf}{Sleep-EDF}{sleep-edf dataset extended}
\newacronym{gssr}{GSSR}{Gonzales-Sulser SYNGAP1 Rodent Dataset}
\newacronym{ai}{AI}{artificial intelligence}
\newacronym{xai}{XAI}{explainable artificial intelligence}
\newacronym{ar}{AR}{Averaged Reference}
\newacronym{nn}{NN}{neural network}
\newacronym{cnn}{CNN}{convolutional neural network}
\newacronym{rnn}{RNN}{recurrent neural network}
\newacronym{fnn}{FNN}{feedforward neural network}
\newacronym{tb}{TB}{tree-based}
\newacronym{lb}{LB}{linear-based}
\newacronym{hpt}{hpt}{hyperparameter tuning}
\newacronym{bci}{BCI}{brain-computer interfaces}
\newacronym{lr}{LR}{logistic regression}
\newacronym{dt}{DT}{decision tree}
\newacronym{rf}{RF}{random forest}
\newacronym{xgb}{XGB}{XGBoost}
\newacronym{mlp}{MLP}{multilayer perceptron}
\newacronym{eegnet}{EEGNet}{compact convolutional network for EEG-based brain-computer interfaces}
\newacronym{lstm}{LSTM}{long short-term memory}
\newacronym{san}{SAN}{self-attention network}
\newacronym{convlstm}{ConvLSTM}{convolutional neural network with long short-term memory}
\newacronym{convtransformer}{ConvTransformer}{convolutional neural network with a self-attention-based transformer classification head}
\newacronym{seq2seq}{Seq2seq}{Sequence-to-Sequence}
\newacronym{tpe}{TPE}{tree-structured parzen estimator}
\newacronym{iea}{IEA}{interictal epileptiform activity}
\newacronym{eeg}{EEG}{electroencephalography}
\newacronym{roc}{ROC AUC}{area under the receiver operating characteristic curve}
\newacronym{prauc}{PRAUC}{area under the precision recall curve}
\newacronym{auc}{AUC}{area under the curve}
\newacronym{mse}{MSE}{mean square error}
\newacronym{shap}{SHAP}{shapley additive explanations}
\newacronym{epoch}{EPOCH}{epoch-based sampling}
\newacronym{ovlp}{OVLP}{any-overlap method}
\newacronym{fir}{FIR}{finite impulse response}
\newacronym{gnb}{GNB}{Gaussian Naïve Bayes}
\newacronym{pac}{PAC}{Passive Aggressive Classifier}
\newacronym{cnn-trf-bm}{CNN-TRF-BM}{CNN-transformer with belief matching loss}
\newacronym{cv}{CV}{cross-validation}
\newacronym{fdr}{FDR}{false discovery rate}
\newacronym{std}{STD}{standard deviation}
\makeglossaries
\begin{document}

\title[Article Title]{PySeizure: A single machine learning classifier framework to detect seizures in diverse datasets}


\author*[1,2,3]{\fnm{Bartłomiej} \sur{Chybowski}}\email{b.s.chybowski@sms.ed.ac.uk}

\author[2]{\fnm{Shima} \sur{Abdullateef}}\email{sabdull2@exseed.ed.ac.uk}

\author[3]{\fnm{Hollan} \sur{Haule}}\email{s1607398@ed.ac.uk}

\author[1,4]{\fnm{Alfredo} \sur{Gonzalez-Sulser}}\email{agonzal2@exseed.ed.ac.uk}

\author[1,3]{\fnm{Javier} \sur{Escudero}}\email{javier.escudero@ed.ac.uk}

\affil[1]{\orgdiv{Muir Maxwell Epilepsy Centre}, \orgname{University of Edinburgh}, 
\city{Edinburgh}, \country{Scotland}}

\affil[2]{\orgdiv{School of Medicine, Deanery of Clinical Sciences}, \orgname{University of Edinburgh}, \orgaddress{\street{50 Little France Crescent}, \city{Edinburgh}, \postcode{EH16 4TJ}, \country{Scotland}}}

\affil[3]{\orgdiv{School of Engineering, Institute for Imaging, Data and Communications}, \orgname{University of Edinburgh}, \orgaddress{\street{Alexander Graham Bell Building, Thomas Bayes Road}, \city{Edinburgh}, \postcode{EH9 3FG}, \country{Scotland}}}

\affil[4]{\orgdiv{School of Medicine, Centre for Discovery Brain Sciences}, \orgname{University of Edinburgh}, \orgaddress{\street{1 George Square}, \city{Edinburgh}, \postcode{EH8 9JZ},
\country{Scotland}}}


\abstract{Reliable seizure detection is critical for diagnosing and managing epilepsy, yet clinical workflows remain dependent on time-consuming manual \acrshort{eeg} interpretation. While machine learning has shown promise, existing approaches often rely on dataset-specific optimisations, limiting their real-world applicability and reproducibility. 
Here, we introduce an innovative, open-source machine-learning framework that enables robust and generalisable seizure detection across various clinical datasets. We evaluate our approach on two publicly available independent \acrshort{eeg} datasets that differ in patient populations and electrode configurations. To enhance robustness, the framework incorporates an automated pre-processing pipeline to standardise data and a majority voting mechanism, in which multiple models independently assess each second of \acrshort{eeg} before reaching a final decision. We train, tune, and evaluate models within each dataset, assessing their cross-dataset transferability.

Our models achieve high within-dataset performance (\acrshort{auc} 0.904$\pm$0.059 for \acrshort{chb-mit}, 0.864$\pm$0.060 for \acrshort{tusz}) and strong cross-dataset generalisation despite differing \acrshort{eeg} setups (\acrshort{auc} 0.615$\pm$0.039 and 0.762$\pm$0.175). Mild post-processing further improved both within- and cross-dataset results. These findings highlight the framework’s potential for deployment in diverse clinical environments. By ensuring complete reproducibility, our framework provides a foundation for robust, dataset-agnostic seizure detection that complements clinical expertise.}



\maketitle

\section{Main}\label{sec:main}
Epilepsy is a chronic neurological disorder characterised by recurrent, unprovoked seizures, affecting approximately 51.7 million individuals globally in 2021~\cite{feigin_global_2025}. It is one of the most prevalent neurological disorders worldwide, significantly impacting quality of life by affecting physical health, cognitive abilities, and social engagement~\cite{reilly_factors_2015}. This highlights the urgent need for effective management and treatment strategies~\cite{meisel_machine_2020}.

Accurate seizure identification is crucial for managing and diagnosing epilepsy. Currently, the clinical gold standard relies on the visual inspection of \gls{eeg} recordings by neurophysiologists~\cite{anuragi_epileptic-seizure_2022}. Although this method is highly accurate, it is labour-intensive, time-consuming, and prone to human variability~\cite{van_donselaar_how_2006, zaidi_misdiagnosis_2000, buettner_high-performance_nodate}. Moreover, limited access to trained specialists, particularly in low-resource settings, exacerbates diagnostic delays and care inequalities~\cite{maloney_association_2022, steer_epilepsy_2014, thomas_056_2012}.

Several studies have recently attempted patient-independent and cross-subject seizure detection. 
Ali \textit{et al.}~\cite{ali_epileptic_nodate} employed the \gls{chb-mit} dataset to address key challenges in seizure detection, including class imbalance and intersubject variability. Their approach used a \gls{rf} classifier with 5-second windows, combined with event-level post-processing. The training data were balanced, while the test data remained unaltered. However, they evaluated the seizure detection performance exclusively within the dataset.
Antonoudiou \textit{et al.}~\cite{antonoudiou_seizyml_2025} proposed the SeizyML framework, initially tested on rodent \gls{eeg} data and later on \gls{chb-mit}. Nonetheless, a large number of recordings and seizures were manually excluded based on duration and amplitude thresholds. 
Zhao \textit{et al.}~\cite{zhao_multi-view_2022} evaluated their method separately on \gls{chb-mit} and \gls{tusz} using models trained within each dataset. Despite the strong reported performance (sensitivity of 0.774 and specificity of 0.763), this study was subjected to several limitations, including the use of a small, manually selected subset of patients with long seizures and differences in input dimensions and configuration settings between the datasets.
Abou-Abbas \textit{et al.}~\cite{abou-abbas_generative_2024} evaluated a method on \gls{tusz} only, using a specific montage configuration. While the results were robust (sensitivity of 0.767 and specificity of 0.955), no external dataset was used for validation. 
Peh \textit{et al.}~\cite{peh_six-center_2023} evaluated models across six datasets using different window sizes. For a 3-second window, they reported strong within-dataset performance (sensitivity of 0.847 and specificity of 0.751). They also presented cross-dataset results, where models trained on \gls{tusz} are evaluated among others on \gls{chb-mit}. These findings highlight a trade-off: longer windows improve accuracy but reduce sensitivity to fine-grained temporal patterns.

These diverse approaches reflect ongoing innovation in seizure detection, focusing on enhancing real-life applicability. The integration of patient-specific models and event-based cross-validation continues to enhance predictive accuracy, advancing the potential of these technologies in managing neurological disorders and improving patient care quality~\cite{ren_performance_2022}.

Nevertheless, despite significant progress in \gls{ml}-driven medical seizure identification, clinical adoption remains limited due to several key issues. Many \gls{ml} models for \gls{seid} were developed and evaluated on single, often small datasets, which are further reduced by selecting only the most favourable examples~\cite{ali_epileptic_nodate}. This process made them susceptible to overfitting to specific recording conditions, feature distributions, or patient populations, ultimately limiting their real-world performance~\cite{bradshaw_guide_2023, steyerberg_internal_2003} and rendering them unable to generalise effectively. Furthermore, a lack of reproducibility and interpretability -- due to proprietary datasets, non-transparent preprocessing steps, or reliance on dataset-specific manual optimisations -- prevents effective clinical integration. The absence of standardised evaluation across datasets further hinders adoption, contributing to the persistent gap between research and real-world implementation~\cite{markowetz_all_2024, aggarwal_diagnostic_2021}. 

To address current limitations in seizure detection, we introduce \textbf{PySeizure}, a modular and clinically oriented \gls{ml} framework designed to support epileptic seizure detection, offering broad applicability across \gls{eeg} datasets. In contrast to previous methods that rely on dataset-specific tuning, curated subsets, or extensive manual preprocessing, PySeizure operates with minimal human intervention and prioritises generalisation, reproducibility, and ease of integration into diverse workflows. It standardises preprocessing, automates feature extraction, and evaluates performance across structurally diverse datasets. The modular architecture supports seamless integration of state-of-the-art \gls{ml} models while maintaining interoperability. Rather than proposing a single model, this study presents the complete detection pipeline as a flexible, scalable solution adaptable to various \gls{eeg} systems. We demonstrate a high level of accuracy across datasets using a generalisable pipeline without any dataset-specific manual data curation and tuning.
PySeizure comprises four main modules. \textit{(1) Standardised preprocessing} includes common filtering steps, optional bipolar re-referencing, and resampling to a uniform frequency, ensuring compatibility across heterogeneous EEG datasets without requiring dataset-specific modifications. Artefact-affected segments are automatically marked rather than excluded, enabling transparent and configurable quality control. This module also supports data augmentation strategies to improve model robustness without requiring additional data collection. (\textit{2) Epoch segmentation and feature extraction} support both raw signal- and feature-based models. The feature extraction module computes nearly 40 features per channel, covering temporal, frequency-based, connectivity, and graph-theory-derived domains. \textit{(3) Model selection and optimisation} are fully configurable, allowing users to choose from simple classifiers to complex architectures depending on their specific research or clinical needs. \textit{(4) Seizure detection on an epoch-wise basis} uses an ensemble of seven models, combined through a majority voting mechanism to improve classification robustness. 
This modular design enhances adaptability across datasets and supports the development of clinically viable \gls{ml} models. By ensuring transparency in preprocessing, feature selection, and evaluation, our work aligns with best practices in \gls{xai}~\cite{chaddad_survey_2023}, essential for regulatory approval and real-world deployment.

\section{Results}\label{sec:results}
We evaluate the performance of PySeizure on two key tasks: within-dataset and cross-dataset seizure detection.

We employed three-fold cross-validation with subject-dependent data splits, ensuring that all recordings from a given patient were assigned to a single set, thereby enhancing the reliability of our results. To improve temporal resolution, we segmented the data into 1-second epochs, allowing for more fine-grained predictions. We trained and evaluated seven models of varying complexity and capability, namely \Gls{lr}, \Gls{xgb}, \Gls{mlp}, \Gls{cnn}, \Gls{eegnet}, \Gls{convlstm}, and \Gls{convtransformer}. Additionally, we implemented a voting mechanism that utilises the predictions of these models to further enhance predictive accuracy.

In the following subsections, we first analyse the models' ability to accurately detect seizures within single datasets, assessing their overall performance using a range of metrics. We then evaluate the models' generalisation ability across different datasets, exploring their robustness in handling configurational variations across \gls{eeg} recordings. To preserve clarity, we provide results without any post-processing. Detailed performance metrics, including \gls{roc}, sensitivity, specificity, and other relevant measures, are provided in the supplementary materials (Supplementary Section \ref{sec:supplementary:tables}), with accompanying visualisations in the main text to highlight key findings, as well as in the supplementary materials (Supplementary Section \ref{sec:supplementary:figures}). Finally, we illustrate the impact of straightforward post-processing on the results.

\subsection{Within-dataset results}
\begin{figure}[!h]
    \centering
    \includegraphics[angle=0,origin=c,width=130mm]{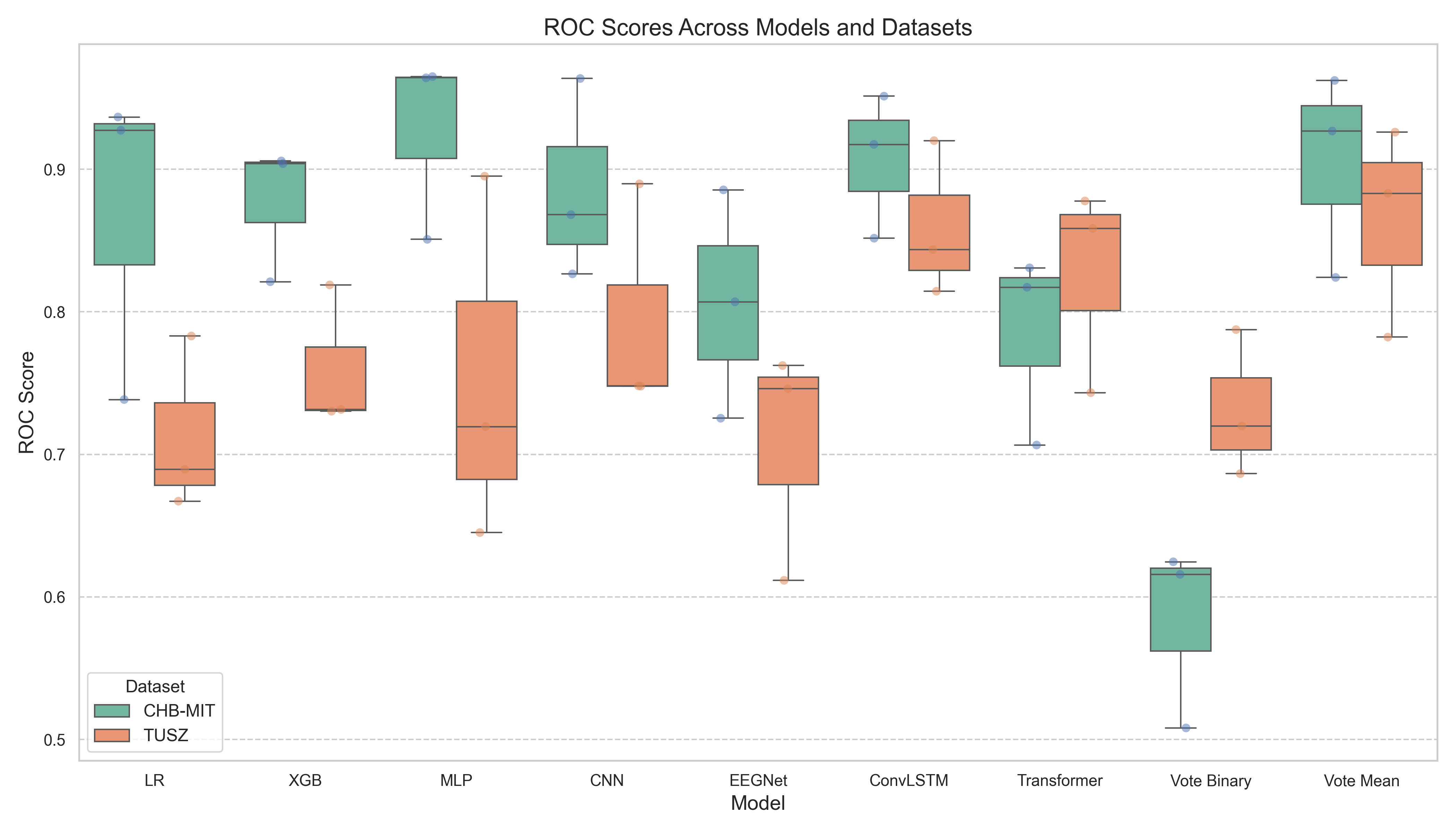}
    \caption{Comparison of \acrfull{roc} scores for \Acrfull{chb-mit} and \Acrfull{tusz} datasets across all the models, including results for binary and mean voting approaches. Dots indicate the scores for individual folds.}
    \label{fig:within_roc}
\end{figure}

We assessed PySeizure on single datasets to evaluate its seizure detection performance. The implemented models demonstrated high accuracy in identifying seizure events within each dataset. These outcomes underscore the model's capacity to differentiate seizures from non-seizure activity across various recording conditions reliably. Figure \ref{fig:within_roc} presents the comparison of \gls{roc} scores across all models for both datasets. 

The remaining metrics are detailed in Supplementary Section \ref{sec:supplementary:figures}. Average performances of selected metrics in individual datasets are provided in Table \ref{tab:tusz-results} and Table \ref{tab:chb-results}, alongside comprehensive results in Supplementary Section \ref{sec:supplementary:tables}

\begin{table}[!h]
\caption{Average performance of proposed models on \Acrfull{tusz} dataset with \acrfull{std}, including results for binary and mean voting approaches.}
\label{tab:tusz-results}
\setlength{\tabcolsep}{3pt}
\begin{tabular}{r|cccccccc}
 &
  \multicolumn{1}{c}{\textbf{ROC}} &
  \multicolumn{1}{c}{\textbf{Accuracy}} &
  \multicolumn{1}{c}{\textbf{Sensitivity}} &
  \multicolumn{1}{c}{\textbf{Specificity}} \\
\hline \\
\textbf{LR} & 0.7132 $\pm$ 0.0502 & 0.6517 $\pm$ 0.0386 & 0.6517 $\pm$ 0.0386 & 0.6574 $\pm$ 0.0404 \\
\textbf{XGB} & 0.7602 $\pm$ 0.0415 & 0.8180 $\pm$ 0.0330 & 0.8180 $\pm$ 0.0330 & 0.7250 $\pm$ 0.0084 \\
\textbf{MLP} & 0.7532 $\pm$ 0.1048 & 0.7282 $\pm$ 0.0727 & 0.7282 $\pm$ 0.0727 & 0.6866 $\pm$ 0.0911 \\
\textbf{CNN} & 0.7952 $\pm$ 0.0668 & 0.7908 $\pm$ 0.0454 & 0.7908 $\pm$ 0.0454 & 0.6923 $\pm$ 0.0832 \\
\textbf{EEGNet} & 0.7067 $\pm$ 0.0675 & 0.6775 $\pm$ 0.1760 & 0.6775 $\pm$ 0.1760 & 0.5683 $\pm$ 0.0478 \\
\textbf{ConvLSTM} & 0.8594 $\pm$ 0.0445 & 0.8549 $\pm$ 0.0342 & 0.8549 $\pm$ 0.0342 & 0.7365 $\pm$ 0.0486 \\
\textbf{ConvTransformer} & 0.8265 $\pm$ 0.0594 & 0.8287 $\pm$ 0.0352 & 0.8287 $\pm$ 0.0352 & 0.6749 $\pm$ 0.0344 \\
\textbf{Binary voting} & 0.7313 $\pm$ 0.0420 & 0.8418 $\pm$ 0.0464 & 0.8418 $\pm$ 0.0464 & 0.7313 $\pm$ 0.0420 \\
\textbf{Mean voting} & 0.8638 $\pm$ 0.0603 & 0.8531 $\pm$ 0.0425 & 0.8531 $\pm$ 0.0425 & 0.7413 $\pm$ 0.0378
\end{tabular}
\end{table}

\begin{table}[!h]
\caption{Average performance of proposed models on \Acrfull{chb-mit} dataset with \acrfull{std}, including results for binary and mean voting approaches.}
\label{tab:chb-results}
\setlength{\tabcolsep}{3pt}
\begin{tabular}{r|cccc}
 &
  \multicolumn{1}{c}{\textbf{ROC}} &
  \multicolumn{1}{c}{\textbf{Accuracy}} &
  \multicolumn{1}{c}{\textbf{Sensitivity}} &
  \multicolumn{1}{c}{\textbf{Specificity}} \\
\hline \\
\textbf{LR} & 0.8675 $\pm$ 0.0913 & 0.8163 $\pm$ 0.0907 & 0.8163 $\pm$ 0.0907 & 0.7990 $\pm$ 0.0471 \\
\textbf{XGB} & 0.8769 $\pm$ 0.0395 & 0.9695 $\pm$ 0.0114 & 0.9695 $\pm$ 0.0114 & 0.8605 $\pm$ 0.0627 \\
\textbf{MLP} & 0.9266 $\pm$ 0.0536 & 0.8906 $\pm$ 0.0260 & 0.8906 $\pm$ 0.0260 & 0.8574 $\pm$ 0.0640 \\
\textbf{CNN} & 0.8861 $\pm$ 0.0574 & 0.8025 $\pm$ 0.2322 & 0.8025 $\pm$ 0.2322 & 0.7655 $\pm$ 0.0789 \\
\textbf{EEGNet} & 0.8059 $\pm$ 0.0654 & 0.9023 $\pm$ 0.1018 & 0.9023 $\pm$ 0.1018 & 0.7245 $\pm$ 0.0660 \\
\textbf{ConvLSTM} & 0.9067 $\pm$ 0.0414 & 0.9286 $\pm$ 0.0694 & 0.9286 $\pm$ 0.0694 & 0.7570 $\pm$ 0.0984 \\
\textbf{ConvTransformer} & 0.7848 $\pm$ 0.0557 & 0.8306 $\pm$ 0.1691 & 0.8306 $\pm$ 0.1691 & 0.6823 $\pm$ 0.0149 \\
\textbf{Binary voting} & 0.5829 $\pm$ 0.0530 & 0.9684 $\pm$ 0.0235 & 0.9684 $\pm$ 0.0235 & 0.5829 $\pm$ 0.0530 \\
\textbf{Mean voting} & 0.9044 $\pm$ 0.0586 & 0.9752 $\pm$ 0.0142 & 0.9752 $\pm$ 0.0142 & 0.6309 $\pm$ 0.0998
\end{tabular}
\end{table}

\subsection{Cross-dataset results}
\begin{figure}[!h]
    \centering
    \includegraphics[angle=0,origin=c,width=130mm]{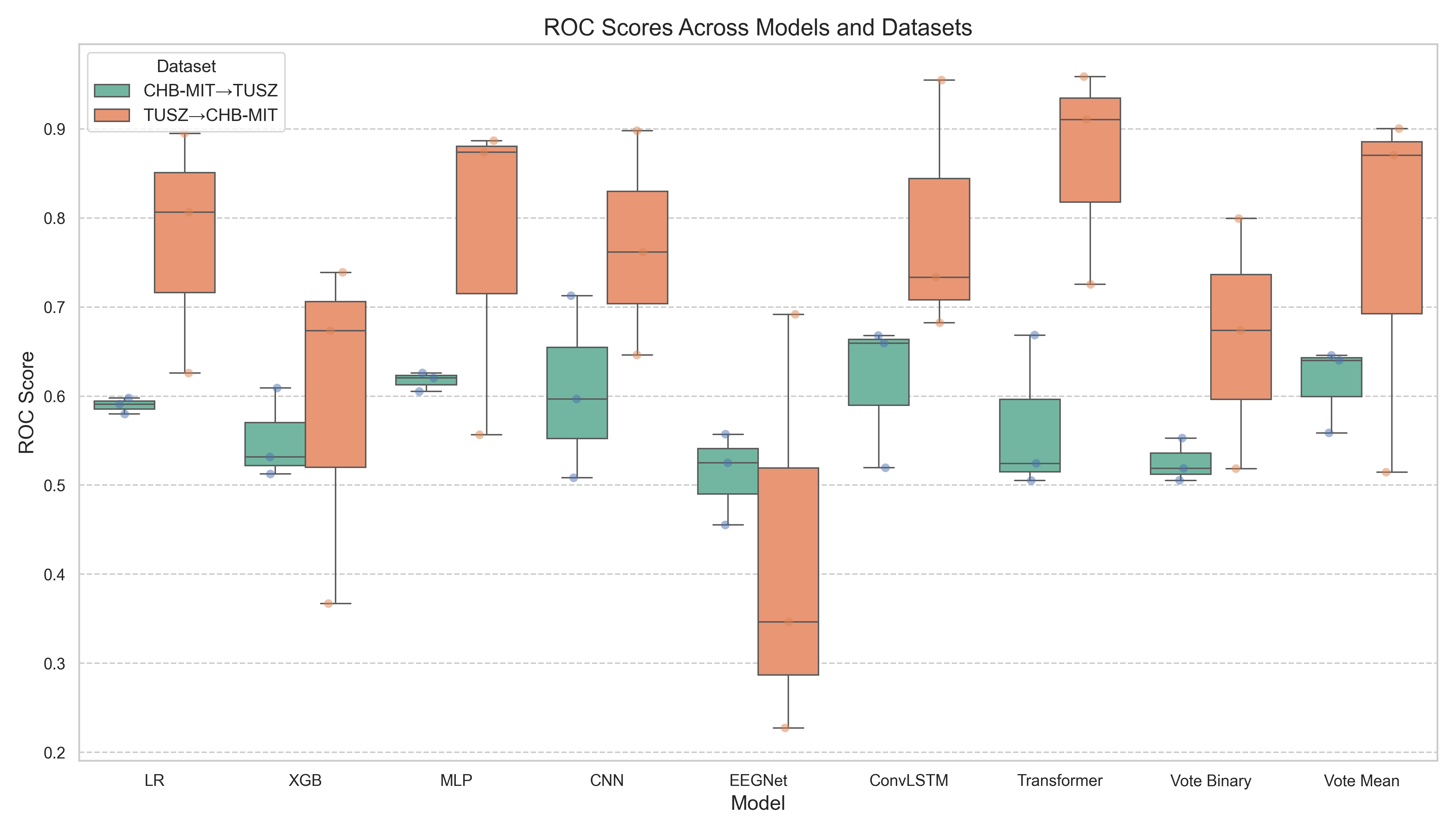}
    \caption{Comparison of \acrfull{roc} scores for all the models trained on the \Acrfull{chb-mit} and evaluated on the \Acrfull{tusz} datasets, including results for binary and mean voting approaches. In the legend, the arrow symbol ($\rightarrow$) denotes that models were trained on the dataset indicated before the arrow and evaluated on the dataset indicated after the arrow.}
    \label{fig:cross_roc}
\end{figure}

To assess the generalisation capability of PySeizure, we tested the framework with all models across multiple datasets with diverse configurational characteristics. Despite variations in recording conditions and patient populations, the models exhibited strong performance, consistently achieving comparable seizure detection accuracy across datasets. Specifically, results for models trained on \gls{chb-mit} and evaluated on \gls{tusz} are presented in Table \ref{tab:chb-on-tusz-results}, while models trained on \gls{tusz} and evaluated on \gls{chb-mit} are shown in Table \ref{tab:tusz-on-chb-results}. Figure \ref{fig:cross_roc} presents the comparison of \gls{roc} scores across all models. 
The remaining metrics are detailed in Supplementary Section \ref{sec:supplementary:figures} and Supplementary Section \ref{sec:supplementary:tables}

This evaluation underscores the framework's ability to adapt to various data sources and manage variations without extensive retraining or post-processing.

\begin{table}[!h]
\caption{The average performance of proposed models trained on \Acrfull{chb-mit} and tested on \Acrfull{tusz} dataset, including results for binary and mean voting approaches.}
\label{tab:chb-on-tusz-results}
\setlength{\tabcolsep}{3pt}
\begin{tabular}{r|cccc}
 &
  \multicolumn{1}{c}{\textbf{ROC}} &
  \multicolumn{1}{c}{\textbf{Accuracy}} &
  \multicolumn{1}{c}{\textbf{Sensitivity}} &
  \multicolumn{1}{c}{\textbf{Specificity}} \\
\hline \\
\textbf{LR} & 0.5895 $\pm$ 0.0074 & 0.6590 $\pm$ 0.0791 & 0.6590 $\pm$ 0.0791 & 0.5511 $\pm$ 0.0129 \\
\textbf{XGB} & 0.5510 $\pm$ 0.0417 & 0.7660 $\pm$ 0.0267 & 0.7660 $\pm$ 0.0267 & 0.5231 $\pm$ 0.0080 \\
\textbf{MLP} & 0.6171 $\pm$ 0.0088 & 0.6930 $\pm$ 0.0457 & 0.6930 $\pm$ 0.0457 & 0.5648 $\pm$ 0.0162 \\
\textbf{CNN} & 0.6059 $\pm$ 0.0837 & 0.6704 $\pm$ 0.0835 & 0.6704 $\pm$ 0.0835 & 0.5383 $\pm$ 0.0287 \\
\textbf{EEGNet} & 0.5125 $\pm$ 0.0426 & 0.7609 $\pm$ 0.0307 & 0.7609 $\pm$ 0.0307 & 0.5015 $\pm$ 0.0057 \\
\textbf{ConvLSTM} & 0.6157 $\pm$ 0.0681 & 0.7118 $\pm$ 0.0690 & 0.7118 $\pm$ 0.0690 & 0.5372 $\pm$ 0.0475 \\
\textbf{ConvTransformer} & 0.5660 $\pm$ 0.0729 & 0.6233 $\pm$ 0.1209 & 0.6233 $\pm$ 0.1209 & 0.5021 $\pm$ 0.0026 \\
\textbf{Binary voting} & 0.5257 $\pm$ 0.0200 & 0.7552 $\pm$ 0.0256 & 0.7552 $\pm$ 0.0256 & 0.5257 $\pm$ 0.0200 \\
\textbf{Mean voting} & 0.6148 $\pm$ 0.0398 & 0.7562 $\pm$ 0.0274 & 0.7562 $\pm$ 0.0274 & 0.5244 $\pm$ 0.0200            
\end{tabular}
\end{table}

\begin{table}[!h]
\caption{The average performance of proposed models trained on \Acrfull{tusz} and tested on \Acrfull{chb-mit} dataset, including results for binary and mean voting approaches.}
\label{tab:tusz-on-chb-results}
\setlength{\tabcolsep}{3pt}
\begin{tabular}{r|cccc}
 &
  \multicolumn{1}{c}{\textbf{ROC}} &
  \multicolumn{1}{c}{\textbf{Accuracy}} &
  \multicolumn{1}{c}{\textbf{Sensitivity}} &
  \multicolumn{1}{c}{\textbf{Specificity}} \\
\hline \\
\textbf{LR} & 0.6610$\pm$0.1712 & 0.6002$\pm$0.2003 & 0.6002$\pm$0.2003 & 0.6102$\pm$0.1527 \\
\textbf{XGB} & 0.5358$\pm$0.1573 & 0.5778$\pm$0.1870 & 0.5778$\pm$0.1870 & 0.5501$\pm$0.1414 \\
\textbf{MLP} & 0.6734$\pm$0.1541 & 0.6503$\pm$0.1292 & 0.6503$\pm$0.1292 & 0.6226$\pm$0.1393 \\
\textbf{CNN} & 0.5502$\pm$0.1515 & 0.5704$\pm$0.1176 & 0.5704$\pm$0.1176 & 0.5427$\pm$0.0608 \\
\textbf{EEGNet} & 0.4630$\pm$0.1839 & 0.3480$\pm$0.2644 & 0.3480$\pm$0.2644 & 0.4208$\pm$0.0978 \\
\textbf{ConvLSTM} & 0.5656$\pm$0.1188 & 0.6402$\pm$0.1017 & 0.6402$\pm$0.1017 & 0.5531$\pm$0.0788 \\
\textbf{ConvTransformer} & 0.6186$\pm$0.0870 & 0.7310$\pm$0.0259 & 0.7310$\pm$0.0259 & 0.5932$\pm$0.0717 \\
\textbf{Binary voting} & 0.5478$\pm$0.0903 & 0.5964$\pm$0.1088 & 0.5964$\pm$0.1088 & 0.5519$\pm$0.0868 \\
\textbf{Mean voting} & 0.6438$\pm$0.1673 & 0.5614$\pm$0.1551 & 0.5614$\pm$0.1551 & 0.5800$\pm$0.1276
\end{tabular}
\end{table}

\subsection{Feature importance}

\begin{figure}[!h]
    \centering
    \begin{subfigure}[b]{0.48\textwidth}
        \caption{}
        \centering
        \includegraphics[width=\linewidth]{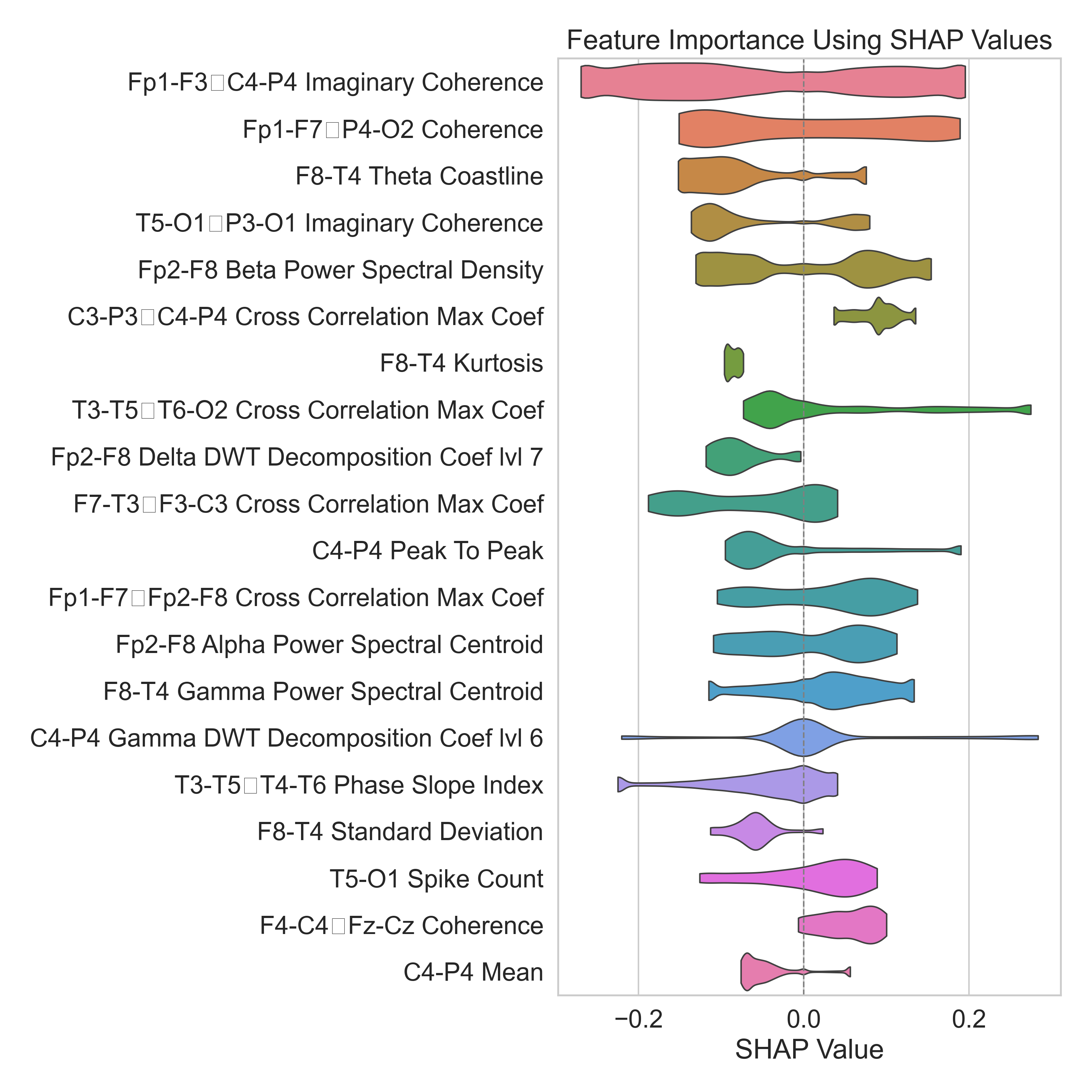}
        \label{fig:feature-shap-calc-a}
    \end{subfigure}
    \hfill
    \begin{subfigure}[b]{0.48\textwidth}
        \caption{}
        \centering
        \includegraphics[width=\linewidth]{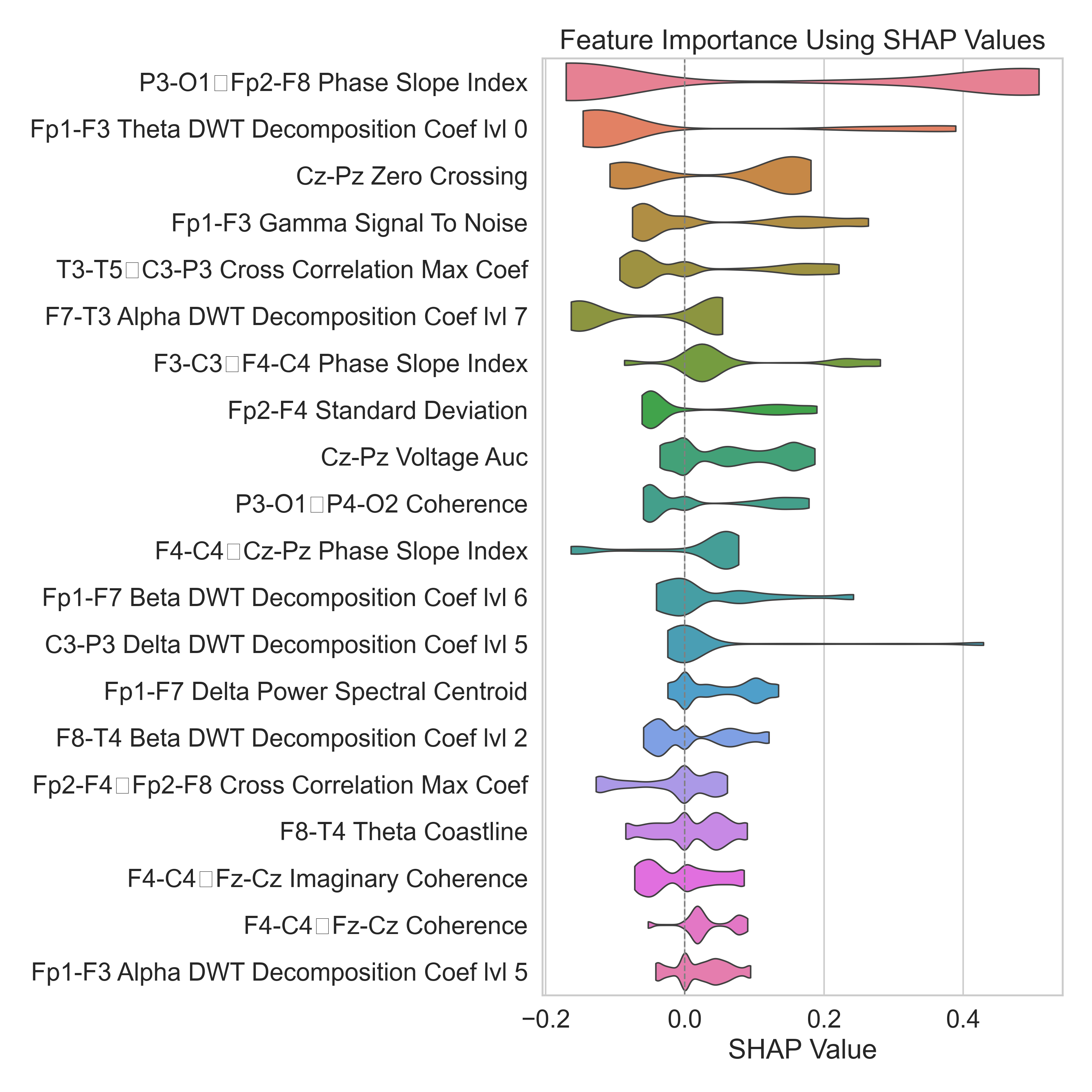}
        \label{fig:feature-shap-calc-b}
    \end{subfigure}

    \begin{subfigure}[b]{0.48\textwidth}
        \caption{}
        \centering
        \includegraphics[width=\linewidth]{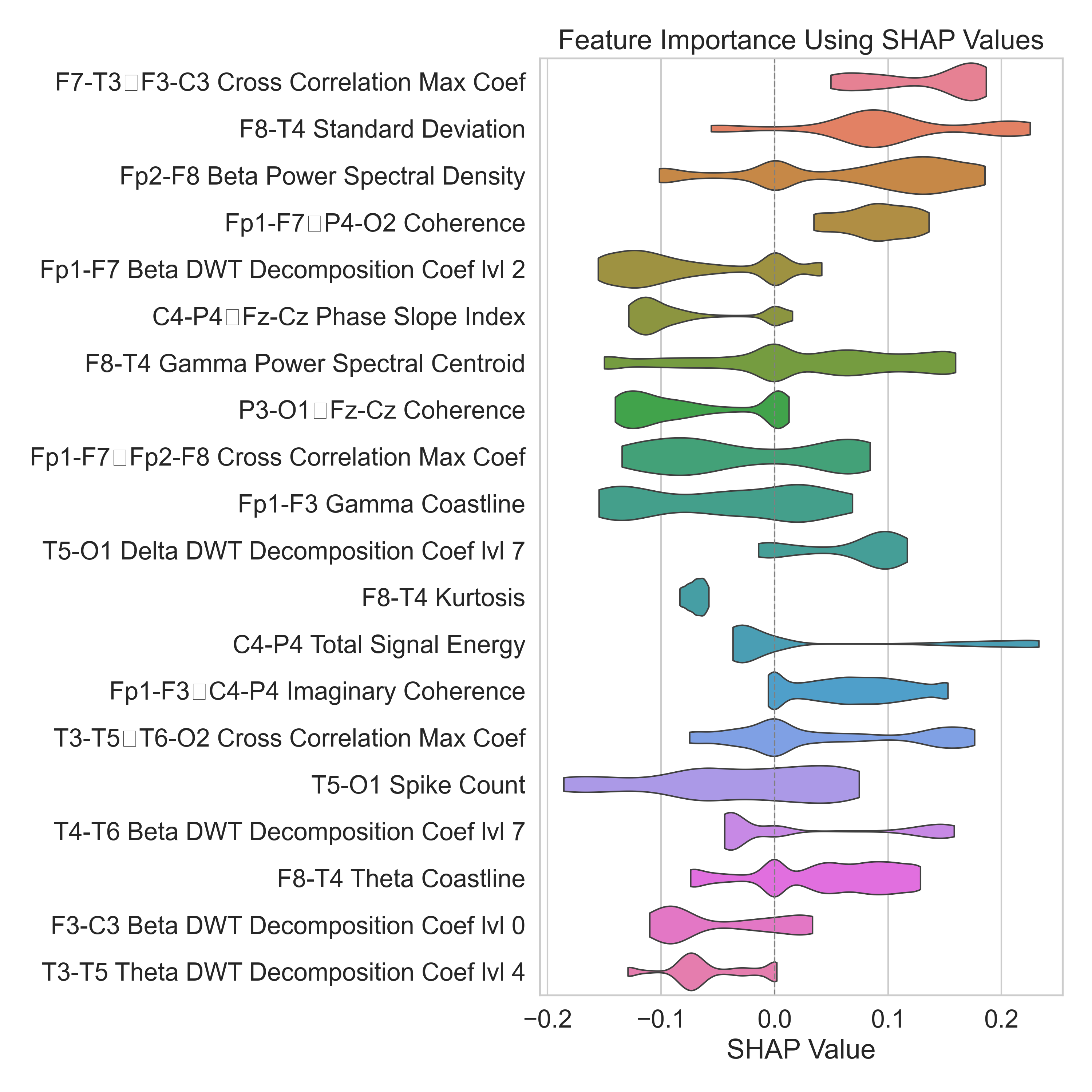}
        \label{fig:feature-shap-calc-c}
    \end{subfigure}
    \hfill
    \begin{subfigure}[b]{0.48\textwidth}
        \caption{}
        \centering
        \includegraphics[width=\linewidth]{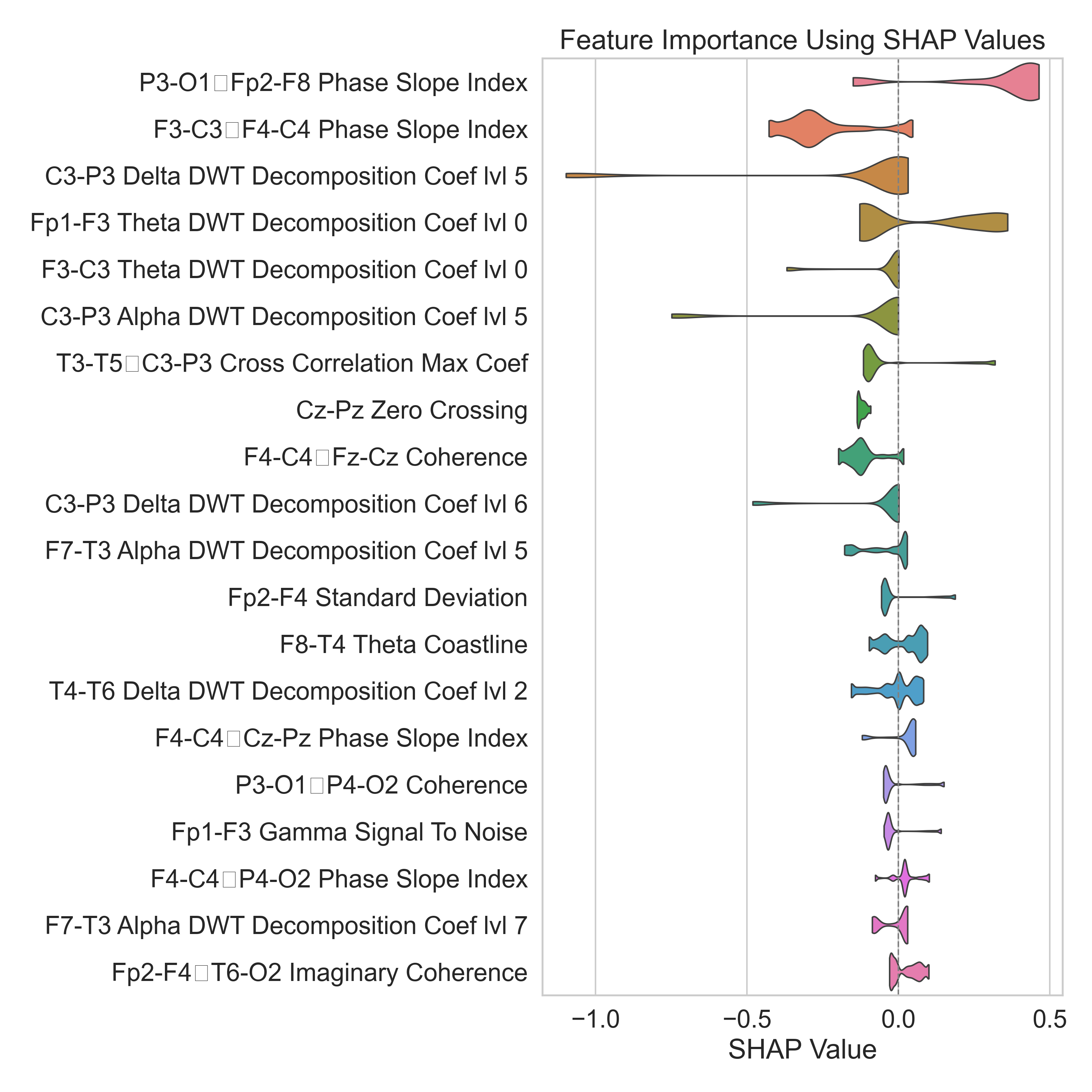}
        \label{fig:feature-shap-calc-d}
    \end{subfigure}
    \caption{Global feature importance derived from \acrshort{shap} values, showing the top twenty most influential features across all models using engineered features within a) the \Acrfull{tusz}, b) the \Acrfull{chb-mit} dataset, and cross-dataset evaluation: c) trained on \acrshort{tusz} and evaluated on \acrshort{chb-mit}, and d) trained on \acrshort{chb-mit} and evaluated on \acrshort{tusz}. Feature importance was aggregated over all evaluation folds to reflect consistent patterns across diverse clinical settings.}
    \label{fig:feature-shap-calc}
\end{figure}

\begin{figure}[!h]
    \centering
    \begin{subfigure}[b]{0.48\textwidth}
        \caption{}
        \centering
        \includegraphics[width=\linewidth]{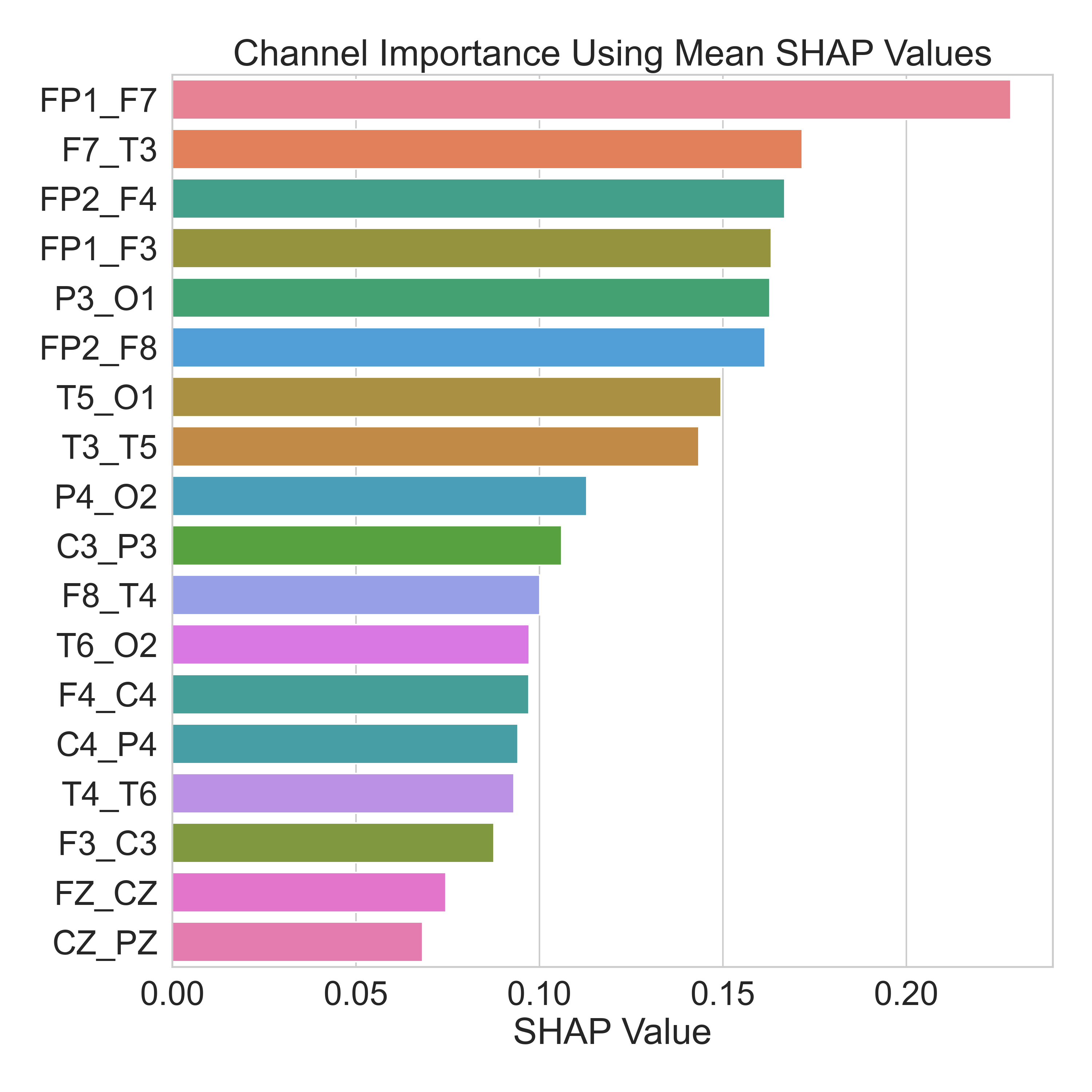}
        \label{fig:feature-shap-raw-a}
    \end{subfigure}
    \hfill
    \begin{subfigure}[b]{0.48\textwidth}
        \caption{}
        \centering
        \includegraphics[width=\linewidth]{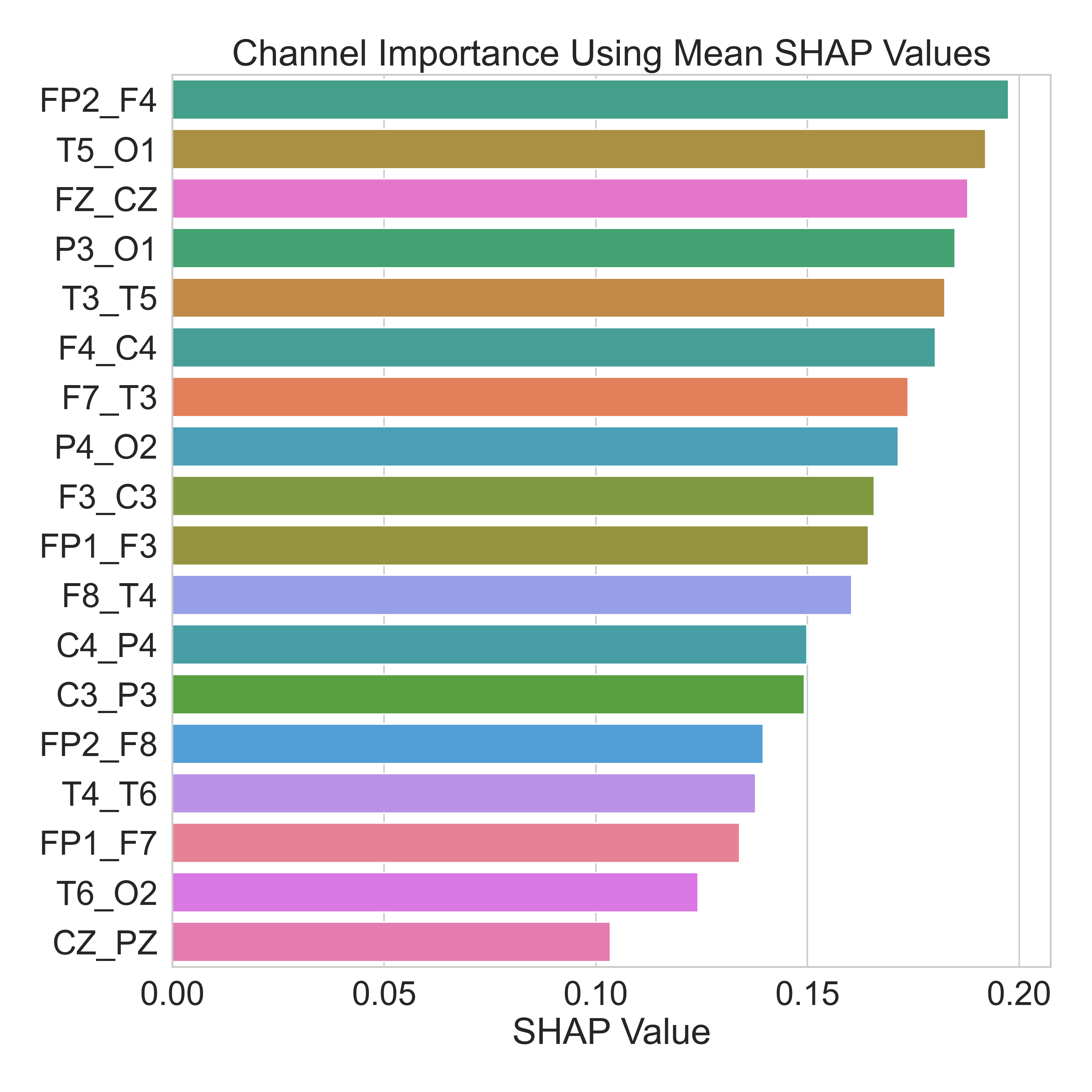}
        \label{fig:feature-shap-raw-b}
    \end{subfigure}

    \begin{subfigure}[b]{0.48\textwidth}
        \caption{}
        \centering
        \includegraphics[width=\linewidth]{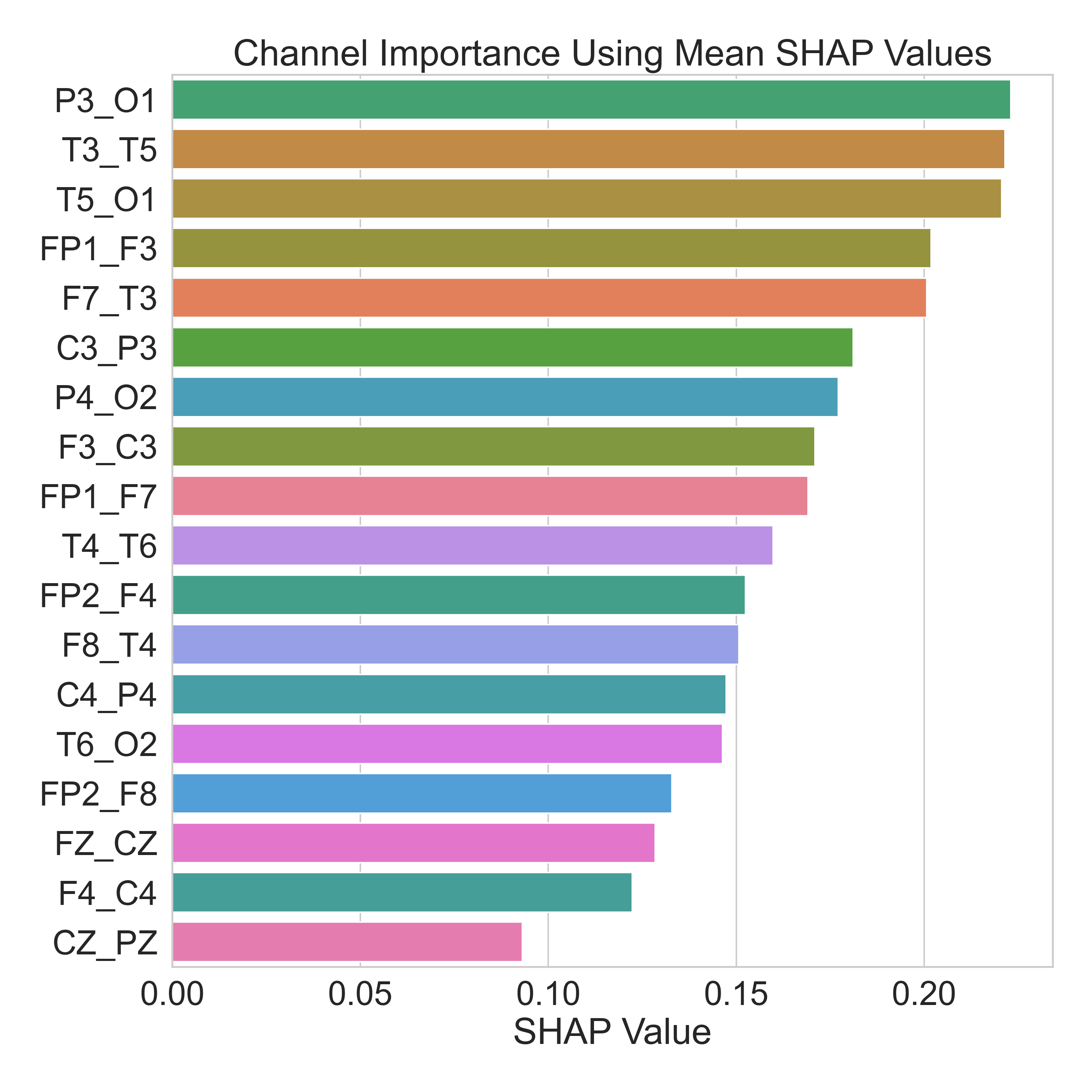}
        \label{fig:feature-shap-raw-c}
    \end{subfigure}
    \hfill
    \begin{subfigure}[b]{0.48\textwidth}
        \caption{}
        \centering
        \includegraphics[width=\linewidth]{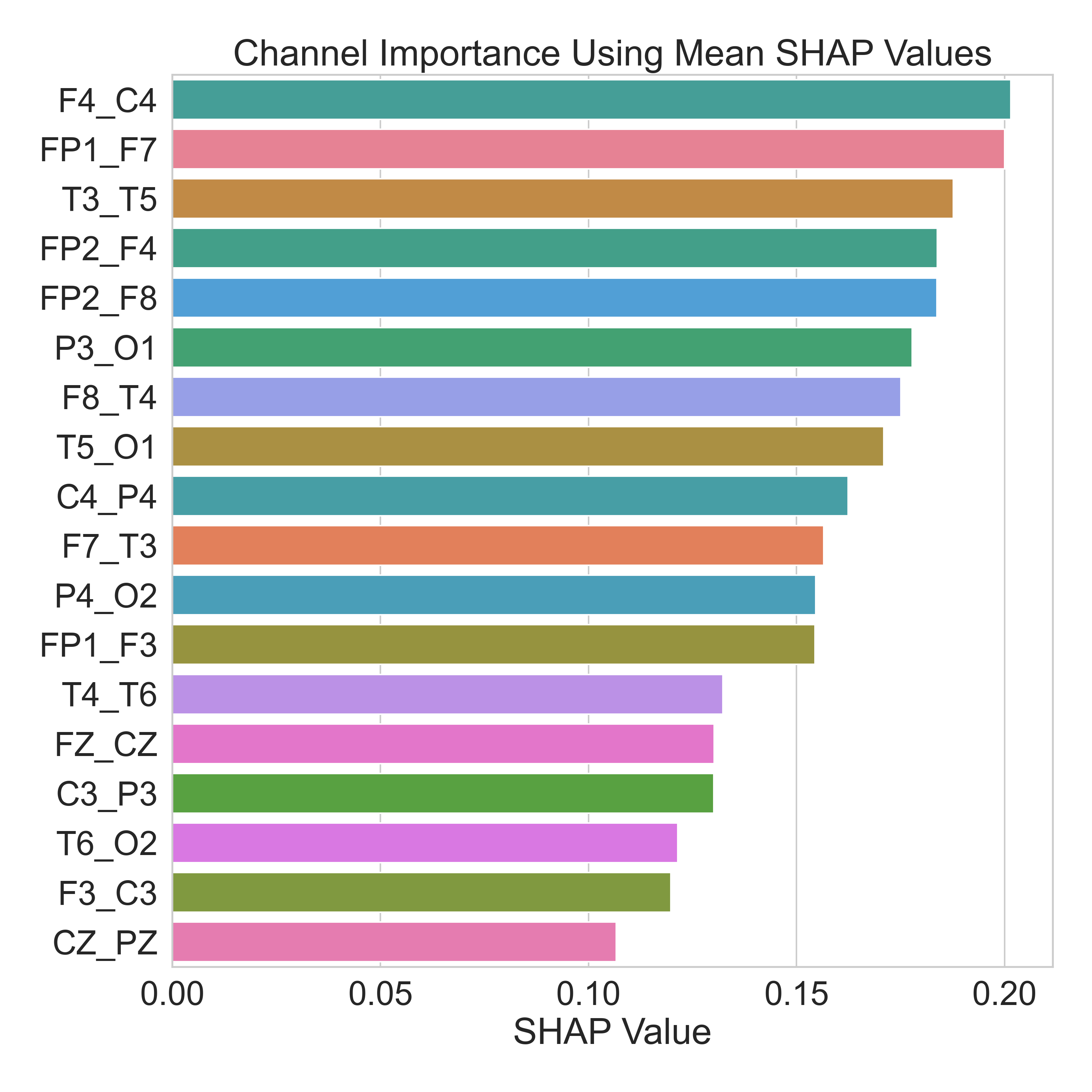}
        \label{fig:feature-shap-raw-d}
    \end{subfigure}
    \caption{Channel importance using \acrshort{shap} values for models using unprocessed data within a) the \Acrfull{tusz}, b) the \Acrfull{chb-mit} dataset, and cross-dataset evaluation: c) trained on \acrshort{tusz} and evaluated on \acrshort{chb-mit}, and d) trained on \acrshort{chb-mit} and evaluated on \acrshort{tusz}.}
    \label{fig:feature-shap-raw}
\end{figure}

Using the Boruta feature eliminator~\cite{kursa_feature_2010}, one of the two methods available in our framework, we selected the best-performing features and conducted \Gls{shap} analysis to identify the most important features for models using these features for seizure detection, as shown in Figure \ref{fig:feature-shap-calc}. For models using raw data, we analysed the channels with the greatest impact (Figure \ref{fig:feature-shap-raw}), where impact refers to the ability to predict whether the analysed segment indicates a seizure.

The \gls{shap} analysis provided insight into the relative importance of individual features by quantifying their contribution to the model's prediction. The most influential variables included a range of features related to power and energy across different frequency bands, such as \textit{Power Spectral Density}, \textit{Power Spectral Centroid}, or \textit{Total Signal Energy}. Temporal features such as \textit{Coastline}, \textit{Zero Crossing}, or \textit{Signal To Noise} defining the shape of the signal and the amount of noise in it, as well as connectivity properties such as the \textit{Phase Slope Index}, \textit{Coherence}, \textit{Imaginary Coherence}, and \textit{cross-correlation maximum coefficient}, also contributed substantially.

For models trained on unprocessed \gls{eeg} data, the channel-level \gls{shap} analysis (Figure \ref{fig:feature-shap-raw}) revealed that frontal-central and temporal channels had the greatest impact on predictions. Detailed \gls{shap} analyses of individual models across datasets are provided in Supplementary Section \ref{sec:supplementary:figures}.

\subsection{Post-processing}

\begin{figure}[h]
    \centering
    \includegraphics[angle=0,origin=c,width=130mm]{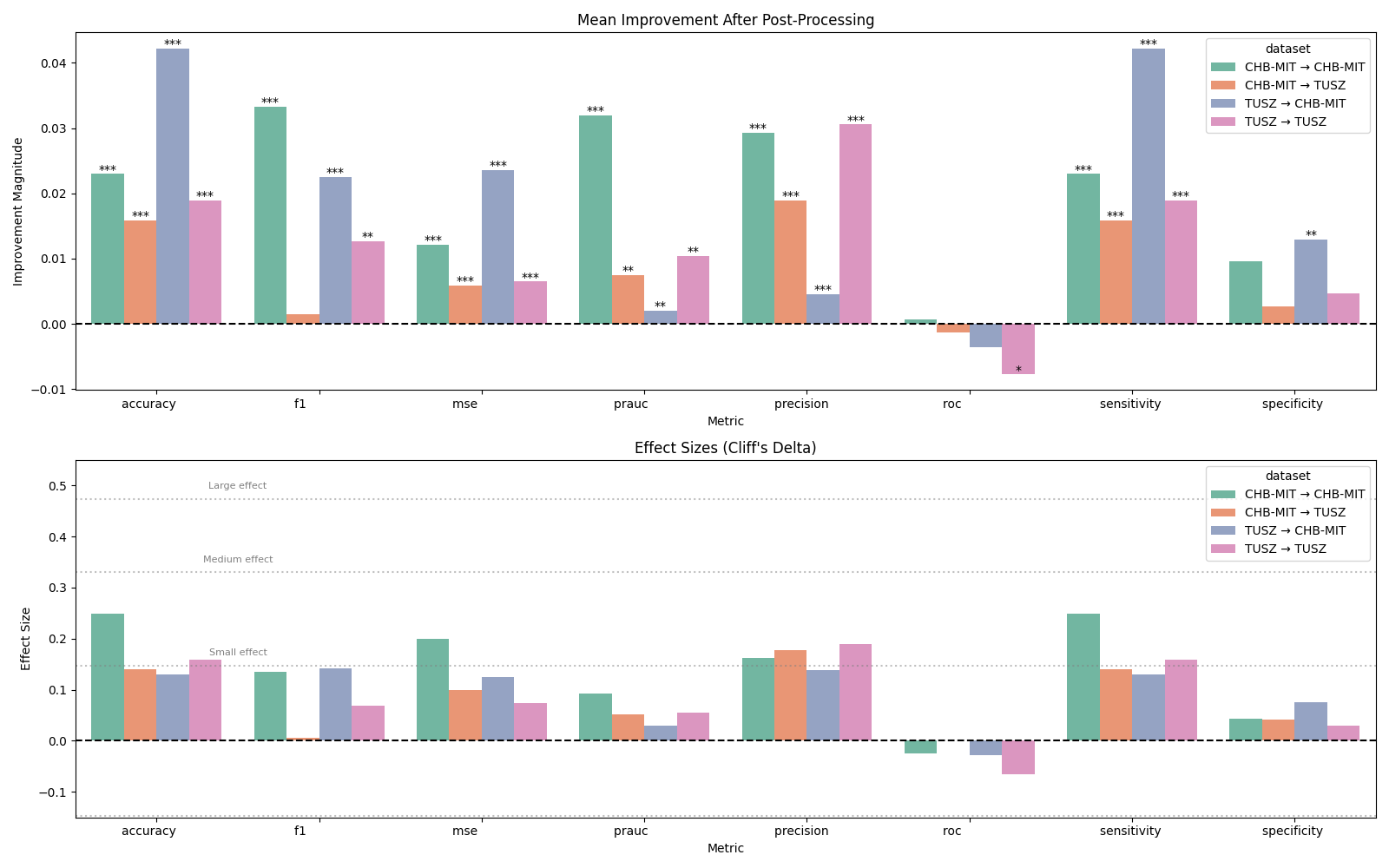}
    \caption{The average improvement on the metric after post-processing. The top plot shows the average improvement and indicator of statistical significance after multiple tests correction using the \Acrfull{fdr} test (Benjamini/Hochberg). The bottom plot shows the effect size calculated using Cliff's Delta. In the legend, the arrow symbol ($\rightarrow$) denotes that models were trained on the dataset indicated before the arrow and evaluated on the dataset indicated after the arrow.}
    \label{fig:post-processing-pval}
\end{figure}

Thus far, we have reported results without any post-processing to provide a clear view of PySeizure's baseline performance -- that is, all results were reported considering an epoch assessment. Here, we apply a mild post-processing (Section \ref{sec:post-processing}), which combines \gls{epoch} and \gls{ovlp} methods~\cite{obeid_objective_2021} to evaluation cross-validation folds. Table \ref{tab:post-processing} reports the average improvement for each metric, along with p-values from the Wilcoxon signed-rank test~\cite{wilcoxon_individual_1945}, corrected for multiple comparisons using the \gls{fdr} method via the Benjamini-Hochberg procedure~\cite{benjamini_controlling_1995}. The results are also visualised in Figure \ref{fig:post-processing-pval}, which additionally shows the effect size calculated using Cliff’s Delta~\cite{macbeth_cliffs_2011}. These results indicate statistically significant improvements across datasets and experimental variations, including models trained and evaluated on the \gls{tusz} and \gls{chb-mit} datasets, as well as cross-dataset evaluations (i.e., trained on \gls{chb-mit} and tested on \gls{tusz}, and vice versa). Detailed results for individual datasets are provided in Supplementary Table \ref{tab:post-processing-all}. Overall, the analysis shows that most post-processing improvements are statistically significant, underscoring the potential of post-processing to enhance model performance and reliability.

\begin{table}[!h]
\caption{The average improvement in the metric after post-processing, with the p-value indicating whether the difference is statistically significant. Asterisks denote significance level: * $p<0.05$; ** $p<0.01$; *** $p<0.001$. Values are aggregated for each metric across all folds and all models. The arrow symbol ($\rightarrow$) denotes that models were trained on the dataset indicated before the arrow and evaluated on the dataset indicated after the arrow.}
\label{tab:post-processing}
\setlength{\tabcolsep}{3pt}
\begin{tabular}{r|lS[table-format=-1.1e-1]l}
    & \textbf{Dataset} & \textbf{Mean improvement} &\textbf{ p-value} \\
    \hline\\
    \multirow{4}{*}{\textbf{ROC}} & CHB-MIT $\rightarrow$ CHB-MIT & 7.30e-04 & 8.70e-02 \\
     & CHB-MIT $\rightarrow$ TUSZ & -1.27e-03 & 3.92e-01 \\
     & TUSZ $\rightarrow$ CHB-MIT & -3.63e-03 & 8.56e-01 \\
     & TUSZ $\rightarrow$ TUSZ & -7.63e-03 & 4.08e-02 \\
     \rule{0pt}{1em}
    \multirow{4}{*}{\textbf{Accuracy}} & CHB-MIT $\rightarrow$ CHB-MIT & 2.30e-02 & 3.32e-05$^{***}$ \\
     & CHB-MIT $\rightarrow$ TUSZ & 1.59e-02 & 6.47e-05$^{***}$ \\
     & TUSZ $\rightarrow$ CHB-MIT & 4.22e-02 & 1.19e-07$^{***}$ \\
     & TUSZ $\rightarrow$ TUSZ & 1.89e-02 & 3.86e-05$^{***}$ \\
     \rule{0pt}{1em}
    \multirow{4}{*}{\textbf{Sensitivity}} & CHB-MIT $\rightarrow$ CHB-MIT & 2.30e-02 & 3.32e-05$^{***}$ \\
     & CHB-MIT $\rightarrow$ TUSZ & 1.59e-02 & 6.47e-05$^{***}$ \\
     & TUSZ $\rightarrow$ CHB-MIT & 4.22e-02 & 1.19e-07$^{***}$ \\
     & TUSZ $\rightarrow$ TUSZ & 1.89e-02 & 3.86e-05$^{***}$ \\
     \rule{0pt}{1em}
    \multirow{4}{*}{\textbf{Specificity}} & CHB-MIT $\rightarrow$ CHB-MIT & 9.66e-03 & 7.20e-02 \\
     & CHB-MIT $\rightarrow$ TUSZ & 2.65e-03 & 2.62e-01 \\
     & TUSZ $\rightarrow$ CHB-MIT & 1.30e-02 & 2.39e-03$^{**}$ \\
     & TUSZ $\rightarrow$ TUSZ & 4.67e-03 & 1.22e-01
\end{tabular}
\end{table}

\section{Discussion}\label{sec:discussion}
\subsection{Clinical relevance and real-world impact}
Reliable seizure detection is critical for both diagnosis and long-term patient management, yet current clinical workflows rely heavily on manual EEG classification, a time-intensive and highly variable process. The proposed framework addresses these challenges by providing an automated and generalisable solution designed for real-world clinical deployment. Unlike traditional approaches that require dataset-specific tuning~\cite{probst_tunability_nodate}, this framework is inherently cross-compatible, standardising \gls{eeg} processing across diverse datasets by automatically handling variations in sampling rates, referencing schemes, and signal artefacts. This adaptability will facilitate future seamless integration into different clinical environments, research studies, and ambulatory monitoring systems, supporting broader clinical adoption of \gls{ai}-driven seizure detection.

A key advantage of this framework is its flexibility. Users can configure pre-processing parameters, artefact handling, feature extraction, and model selection to suit specific clinical or research objectives. For instance, instead of excluding noisy epochs outright, the system allows them to be marked, enabling users to make informed decisions based on data quality rather than applying rigid exclusion criteria. Another example is the framework’s support for both raw signal-based and feature-based models, allowing integration into a wide range of analytical workflows. In this work, we evaluated models spanning the spectrum from simple classifiers, such as logistic regression -- which is unlikely to fully capture the complexity of EEG signals -- to deep learning architectures designed to learn directly from raw data. This range was chosen deliberately to demonstrate the versatility of the framework across different modelling paradigms.
This adaptability aligns with precision medicine initiatives by supporting individualised analyses and enabling rapid, data-driven clinical decision-making.
While our experiments report results for a single configuration, many of the framework’s parameters -- such as the epoch length or downsampling -- are user-adjustable. We established default parameters empirically, for example, a 1-second epoch length to maximise temporal resolution and improve the granularity of predictions. The sampling frequency of 256~Hz was chosen to ensure consistency across datasets, as it was the lowest common frequency available. These choices are not fixed, and users may tailor them to better suit their specific data or application needs.

In addition to evaluating the current version on publicly available datasets, we also tested early versions of the framework~\cite{35iec_abstracts_2023} using clinical data~\cite{15eec_abstracts_2024}. These findings further underscore the potential impact and real-world applicability of our approach in clinical environments. 
Our results already demonstrate strong performance across diverse datasets, even when using general-purpose models. This suggests that PySeizure provides a robust foundation for seizure detection. We hypothesise that incorporating state-of-the-art architectures specifically tailored to EEG or seizure detection may further enhance performance, particularly in clinical applications.
Automated seizure monitoring could enable earlier intervention, reduce misdiagnoses, and support continuous assessment of treatment efficacy~\cite{rai_automated_2024}. Furthermore, by structuring EEG data in a standardised format, the framework facilitates large-scale, multi-centre validation studies -- a critical step towards regulatory approval and clinical integration. By bridging the gap between \gls{ai} research and real-world neurology practice, this framework paves the way for more scalable, efficient, and accessible seizure detection in clinical care. Moreover, the framework aligns with the broader trend of \gls{ai} systems designed to support, rather than replace, clinical decision-making.

\subsection{Alignment with trends in \gls{ai} and healthcare}
The proposed framework aligns with the growing trend of leveraging \gls{ai} to improve healthcare outcomes, particularly in the domain of neurology. As healthcare increasingly adopts \gls{ai}-driven tools for diagnostics and decision-making, the need for reliable, generalisable, and scalable solutions becomes paramount. Our framework directly addresses this need by offering an adaptable, cross-dataset solution for seizure detection that can seamlessly integrate into diverse clinical environments, ensuring that \gls{ai} models can be reliably applied across multiple hospital systems, patient populations, and \gls{eeg} recording configurations.

A key trend in \gls{ai} healthcare applications is the emphasis on generalisability, allowing \gls{ai} models to be effective not only on the data they were trained on but also across different datasets~\cite{yang_generalizability_2024}. The framework’s ability to handle varying signal quality, electrode configurations, and artefacts across datasets ensures its robustness in real-world clinical scenarios, positioning it as a strong candidate for widespread adoption. This emphasis on cross-dataset generalisation is crucial for \gls{ai} solutions that need to function in a variety of clinical settings, as datasets are often heterogeneous, especially in multi-centre studies and global healthcare systems.

Our results demonstrate strong generalisation across heterogeneous datasets using a fixed, short 1-second window and minimal pre-processing. In contrast, several prior studies rely on dataset-specific manual adjustments, hand-picked examples or larger window sizes. Here, we compare our work with studies mentioned in the Main Section (Section \ref{sec:main}), emphasising the differences between solutions. Ali \textit{et al.}~\cite{ali_epileptic_nodate} report sensitivities of 0.726 (subject-wise 5-fold) and 0.753 (Leave One Out) using a \gls{rf} classifier on 5-second segments from \gls{chb-mit}, with event-level post-processing. The training data are balanced while the test data remain unmodified. Their results are limited to \gls{chb-mit} only.
Antonoudiou \textit{et al.}~\cite{antonoudiou_seizyml_2025} evaluate four classical classifiers -- \Gls{gnb}, \Gls{dt}, \Gls{sgd} classifier, and \Gls{pac} -- on \Gls{chb-mit} after excluding seizures shorter than 30 seconds or with low amplitude. This filtering removes 66 recordings and 99 seizures, leaving 86 seizures across 24 subjects. Their best F1 score for \gls{chb-mit} is approximately 0.2, and they do not evaluate on other \gls{eeg} datasets. The code is publicly available.
Zhao \textit{et al.}~\cite{zhao_multi-view_2022} train and test on balanced data from manually selected patients -- 9 from \gls{chb-mit} and 12 from \gls{tusz} -- choosing cases with longer seizures. Their within-dataset performance is high, reporting accuracy: 0.767, specificity: 0.763, sensitivity: 0.774, AUC: 0.826, and F1: 0.761. However, they do not report any cross-dataset evaluations, and the work is not publicly available.
Abou-Abbas \textit{et al.}~\cite{abou-abbas_generative_2024} focus exclusively on \gls{tusz} with an \gls{ar} montage. They report accuracy: 0.917, recall: 0.767, precision: 0.808, and specificity: 0.955. Their evaluation is limited to a single dataset and montage. The code is not available publicly. Peh \textit{et al.}~\cite{peh_six-center_2023} present a study across six datasets using their \gls{cnn-trf-bm} model. They demonstrate that performance improves with larger windows, peaking at a window size of 20 seconds. However, this comes at the cost of temporal granularity. For a 3-second window -- their shortest evaluated and closest to PySeizure's 1-second setup -- the \gls{cnn-trf-bm} model achieves accuracy: 0.823, sensitivity: 0.885, specificity: 0.616, and F1: 0.824 on \gls{tusz}; and accuracy: 0.833, sensitivity: 08080, specificity: 0.886, and F1: 0.837 on \gls{chb-mit}. When trained on \gls{tusz} and tested on \gls{chb-mit}, the performance drops to accuracy: 0.584, sensitivity: 0.365, specificity: 0.959, and F1: 0.547. The implementation is not publicly available. These findings highlight the trade-off between performance and temporal resolution: while longer windows yield higher accuracy, they may reduce the model’s ability to capture fine-grained temporal patterns, which PySeizure targets more directly using 1-second windows, although with slightly lower overall metrics.

\subsection{Limitations and challenges}
We acknowledge several limitations and challenges in our work. A key limitation is the reliance on publicly available datasets, which may not fully capture the diversity of clinical data. To address this, future efforts should include validation with hospital-acquired \gls{eeg} data to ensure robustness in diverse clinical settings. Incorporating advanced artefact suppression techniques could further mitigate noise impacts, enhancing model reliability.

Another challenge is the handling of artefacts and noisy epochs. The framework automatically marks artefact-affected segments rather than discarding them, allowing users to make informed decisions about data inclusion. However, this approach requires careful consideration, as excessive noise may still impact model performance. Future improvements could incorporate advanced artefact suppression techniques or adaptive weighting strategies to mitigate the influence of low-quality recordings.

Finally, the computational demands of feature extraction and the model ensemble present an additional challenge. While the system is optimised for scalability, training and deploying multiple models require substantial computational resources. 
Running the complete PySeizure pipeline on large datasets such as \gls{tusz} or \gls{chb-mit} requires several days on a high-performance computing cluster~\cite{eddie_edinburgh_2024}, primarily due to time-consuming steps such as data preprocessing, feature extraction, and model training. While inference is fast, especially for short recordings, comprehensive evaluations with \gls{shap} analysis can still take hours for entire datasets. This limits deployment in real-time or resource-constrained settings. To address this, techniques such as model pruning or knowledge distillation could improve efficiency with minimal performance loss. Future work may also focus on reducing ensemble size by selecting either the highest-performing models or those meeting predefined performance thresholds.

\section{Methods}\label{sec:methods} 
In this section, we detail the methodologies employed in our study, beginning with a presentation of the datasets (Section \ref{sec:datasets}). We then provide an overview of the architecture (Section \ref{sec:architecture}), followed by a detailed description of the pre-processing steps (Section \ref{sec:pre-processing}), models (Section \ref{sec:models}) and post-processing (Section \ref{sec:post-processing}. For complete transparency and to enable replication, the code is available in the GitHub repository~\cite{pyseizure_code}.

\subsection{Datasets}\label{sec:datasets}
In this section, we will discuss the datasets chosen for this study: the \gls{tusz} and \gls{chb-mit}. These datasets have been selected due to their large size and variability, offering both intra- and inter-dataset diversity. The rich variability found within each dataset, alongside the differences between them, provides a comprehensive framework for evaluating the robustness and generalisability of algorithms. 
Datasets used in this study are publicly available and were accessed in accordance with their respective data use agreements. Ethical approval for this work was obtained from the University of Edinburgh School of Engineering.

\subsubsection{\Acrlong{tusz}}\label{sec:tusz}
The \acrfull{tusz} is a comprehensive dataset widely utilised in seizure detection research, comprised of a large collection of annotated \gls{eeg} recordings~\cite{obeid_temple_2016}. The dataset includes 1,493 \gls{eeg} sessions from 613 patients, with a total of 2,981 seizure events. It covers eight distinct seizure types, with expert-verified annotations detailing the precise onset and offset of each event. \gls{eeg} recordings are sampled at a minimum rate of 250 Hz per second. Annotations are available in two formats: per channel, which provides event details for individual \gls{eeg} channels, and for all channels, offering an aggregated view of the events. Additionally, the dataset includes recordings with one of three reference point types: linked ears, average reference, or an alternative version of the average reference. \gls{tusz} is continually updated, with ongoing improvements to annotations that enhance its clinical relevance. Reflecting real-world clinical conditions, the dataset incorporates inter-patient variability and variations in seizure manifestations. For this study, we conducted experiments using version 2.0.0 of the dataset.

\subsubsection{\Acrlong{chb-mit}}\label{sec:chb}
The \acrfull{chb-mit} dataset, developed by the \gls{mit} and \gls{chb}, is a widely recognised resource for seizure detection and prediction research~\cite{shoeb_application_2009}. It contains 915 hours of EEG recordings from 23 paediatric subjects, with a total of 198 episodes, including 84 seizures across 5 seizure types. The dataset adopts the International 10-20 System’s bipolar montage method, capturing \gls{eeg} signals from 22 electrodes at a 256~Hz sampling rate. In some cases, recordings are made with 18 channels. Each \gls{eeg} recording file typically lasts for one hour, with most subjects having between 9 and 42 consecutive \gls{eeg} files. The dataset includes annotations specifying the precise onset and offset times for each seizure event. Additionally, data from \textit{CHB01} and \textit{CHB21} were collected from the same patient, 1.5 years apart, providing an opportunity for studying the evolution of seizures over time. \Gls{chb-mit}’s expert-verified annotations, detailing seizure events across all channels. The dataset reflects real-world clinical conditions, with inter-patient variability and challenges in detecting seizures in paediatric populations. 

\subsection{Architecture overview}\label{sec:architecture}

\begin{figure}[h]
    \centering
    \includegraphics[angle=0,origin=c,width=130mm]{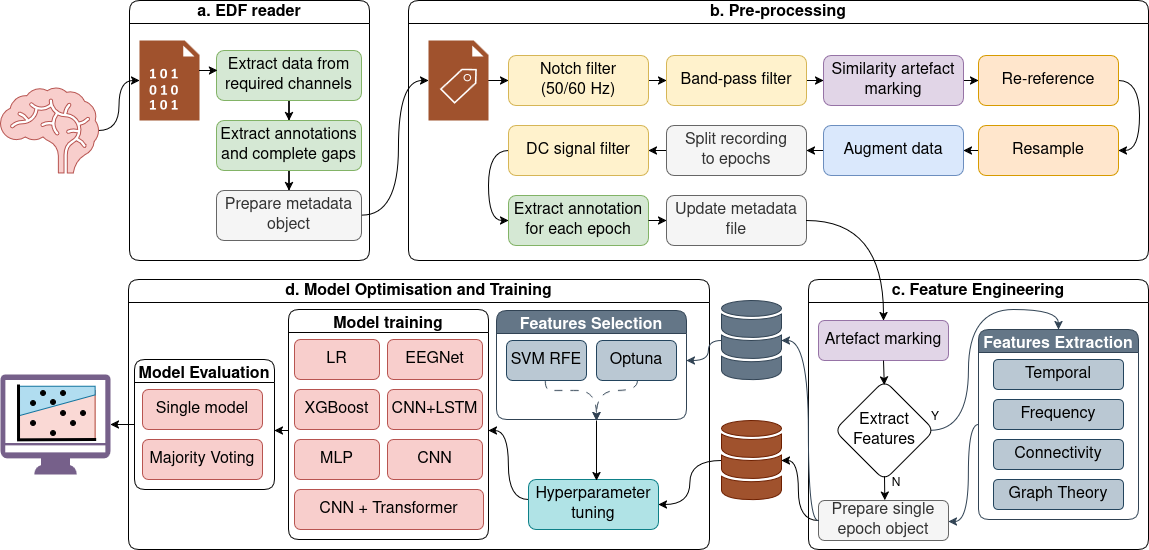}
    \caption{Diagram of the modules and components of the framework. a) Data reader module responsible for reading the data from \gls{edf} files and extracting the annotation either from external files or annotations embedded in data files; b) Pre-processing module liable for filtering and marking common artefacts as well as for resampling, re-referencing, and augmenting data; c) Feature engineering module responsible for calculating features and preparing final shape of the data; d) Model optimisation and training module managing the optimisation and training of the models along with their evaluation.}
    \label{fig:pipeline}
\end{figure}

Figure \ref{fig:pipeline} provides an overview of the architecture, divided into modules that will be described in more detail in the following subsections.

The proposed framework processes \gls{eeg} data from \gls{edf} files through a structured pipeline designed for robust seizure detection. The preprocessing stage includes automatic frequency filtering to reduce artefacts, re-referencing signals to a bipolar montage, resampling to a predefined (in configuration, by user, default value is 256~Hz) frequency, and automatically marking epochs that are considered noisy or artefact-contaminated, allowing the user to decide whether to include them in training (by default excluded from training). The data is then segmented into epochs, serving as the basis for further analysis.

To improve the generalisability of the model, data augmentation is applied using sign flipping, time reversal~\cite{abdallah_cross-site_2023}, and their combination, effectively quadrupling available data. For models employing feature-based learning, a feature extraction step derives nearly 40 unique features per channel, encompassing temporal and frequency-domain characteristics, inter-channel connectivity, and graph-theoretic properties. The full list of available features is presented in the Supplementary Table \ref{tab:features-list}.

A feature selection step follows, ensuring that only the most relevant features are retained for classification. The framework offers two configurable options: the Boruta feature elimination algorithm~\cite{kursa_feature_2010}, which identifies all relevant features by comparing them to randomised shadow features using a random forest classifier, and Cross-Validated Support Vector Machine Recursive Feature Elimination (\acrshort{svm}-\acrshort{rfe}\acrshort{cv}), which recursively removes the least informative features based on their weights in a linear \gls{svm} model~\cite{guyon_gene_2002}. The default framework's feature selector is Boruta, as it exhibits a higher stability in feature selection compared to \gls{rfe}~\cite{degenhardt_evaluation_2017}. Furthermore, Boruta, with the number of iterations set at 20, has been faster in comparison to \acrshort{svm}-\acrshort{rfe}\acrshort{cv} with 3 cross-validation rounds for both tested datasets. The results presented in the paper were computed with Boruta for the reasons mentioned earlier.

Hyperparameter tuning is individually performed for each of the seven models to optimise their configurations, using the Optuna~\cite{akiba2019optuna} framework. This approach ensures that each model is precisely tailored for optimal performance. 

During the evaluation, each model independently classifies every second of the \gls{eeg} data, determining whether the segment indicates a seizure. We also implemented a majority voting mechanism, in which each model's prediction contributes to the final decision. 

The framework addresses class imbalance in two ways. Users can optionally provide custom class weights to rebalance the loss function during training. However, by default, the framework selects all seizure epochs and randomly samples an equal number of background epochs for training, feature selection, and hyperparameter tuning. This balanced subset ensures robust model development. For final evaluation, all available epochs (typically highly imbalanced) are used without shuffling, thereby preserving the original temporal order of the recordings.

The framework is designed to be highly flexible and scalable, allowing researchers to customise input data, parameters, feature sets, and processing steps to suit different datasets and clinical requirements, enhancing adaptability across various applications.

\subsection{Preprocessing}\label{sec:pre-processing}
To ensure consistency and quality in \gls{eeg} signal processing, which is crucial for the analysis of such signals~\cite{del_pup_more_2025}, the framework implements a structured, automated pre-processing pipeline. Initially, \gls{eeg} data are imported from \gls{edf} files with accompanying annotations, which are standardised into a structured table based on the predefined configuration. This supports both generalised and per-channel annotations, while allowing for the optional inclusion of artefact markers, facilitating flexible downstream analysis.

\begin{figure}[h]
    \centering
    \includegraphics[angle=0,origin=c,width=130mm]{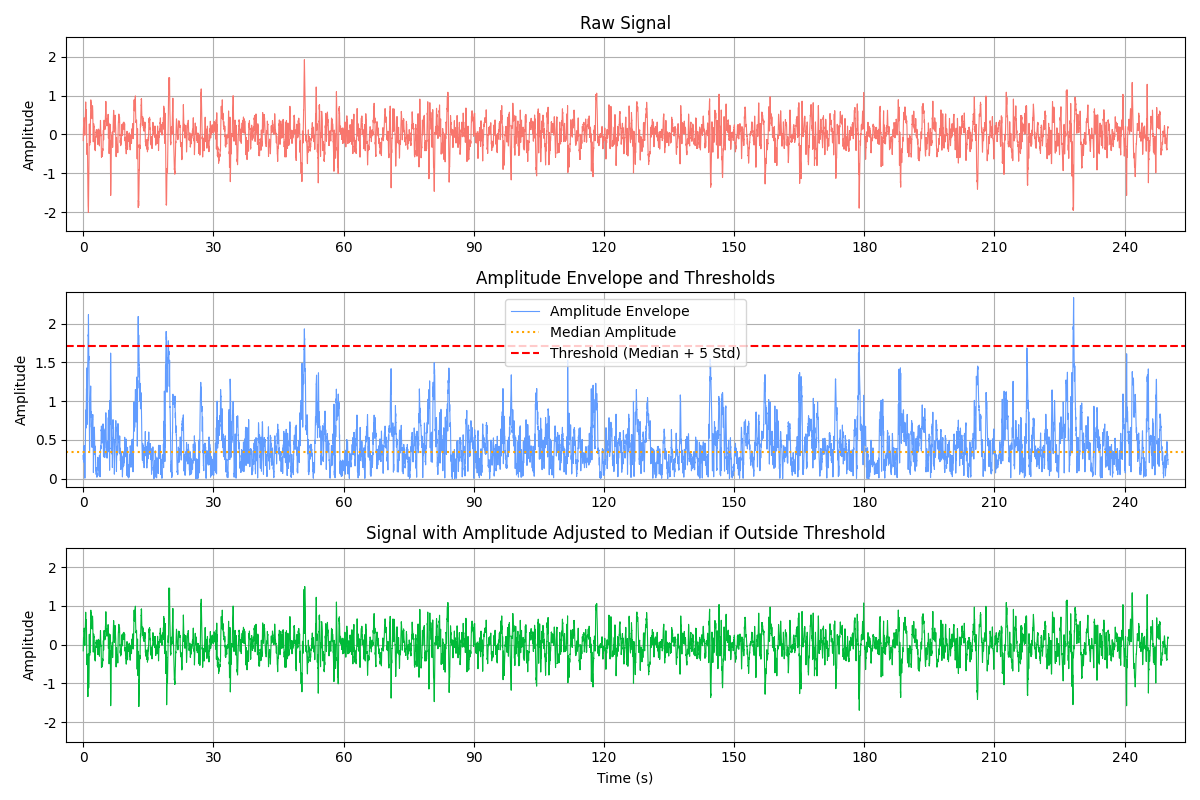}
    \caption{Threshold-based smoothing of the signal. The top plot shows the raw, unprocessed signal. The middle plot shows the amplitude envelope with empirically derived thresholds. The bottom plot shows the processed signal where out-of-threshold amplitudes are normalised to the median value, preserving in-range dynamics.}
    \label{fig:smoothing}
\end{figure}

To improve signal integrity and enhance cross-dataset compatibility, the framework applies automatic filtering to remove common artefacts, including power line interference at 50 and 60~Hz and the Hanning window \gls{fir} dual high-band filter of 0.6~Hz. We also propose a smoothing mechanism that detects unnaturally high amplitudes -- defined as those exceeding $\pm 5$ times the \gls{std} of the median amplitude -- thereby addressing the problem of changes in signal gain (Figure \ref{fig:smoothing}). We also mark sections of the recording where individual signals are extremely similar (cosine similarity above 0.95) as artefacts. The signals are then resampled to a uniform frequency (default: 256~Hz), ensuring comparability across recordings acquired at different sampling rates. If the data are recorded in a unipolar montage, an optional re-referencing step converts them to a bipolar configuration, reducing inter-electrode variability and aligning the signal representation across datasets. These standardisation steps are critical for enabling robust model performance across datasets.

The data are then segmented into non-overlapping epochs (default: 1 second), which serve as the fundamental units for analysis. The default duration of 1 second reflects a practical compromise -- short enough to limit background activity in brief seizures, yet sufficiently long to capture ictal features -- while remaining adaptable to other epoch lengths depending on clinical or computational requirements.
Each epoch is assessed for artefacts using a combined slope-based approach, which identifies segments with exceptionally steep slopes~\cite{fasol_single-channel_2023} and includes our implementation of flatlining detection to capture signal loss or amplifier saturation (i.e., when the signal plateaus due to disconnection or hardware limits). Rather than excluding noisy epochs, the framework flags them, allowing researchers to determine their inclusion based on study requirements.

The framework enables two analytical workflows. In the first case, raw EEG epochs are provided for models that leverage direct signal analysis. In the second case, a comprehensive feature extraction process derives nearly 40 unique features per channel. These features capture temporal and frequency characteristics, inter-channel connectivity, and properties derived from graph theory~\cite{mozafari_automatic_2019, zabihi_patient-specific_2013, guo_epileptic_2022, boonyakitanont_review_2020, networkx_2025}. We present the full list of features in the Supplementary Section \ref{sec:supplementary:tables}. This approach ensures the retention of clinically and computationally relevant information, enhancing the interpretability of the models.

All processed epochs, whether in raw or feature-extracted form, are stored as structured records in an SQLite database, facilitating efficient retrieval, reproducibility, and integration into large-scale clinical studies.

\subsection{Models}\label{sec:models} 
Here, we present the models implemented within our framework. All deep learning models incorporating normalisation layers utilise Batch Normalisation, which was selected as the default during development. Additionally, Leaky ReLU activation functions and Kaiming normal initialisation were employed where appropriate to enhance model convergence and stability~\cite{he_delving_2015}. Hyperparameter optimisation is performed independently for each cross-validation fold and each dataset as part of the processing pipeline. As a result, we do not use a fixed architecture. Instead, we define an optimisation search space for each model and report the range of selected values across folds to guide future uses of our pipeline in similar problems. The complete search space is presented in the Supplementary Table \ref{tab:search-space}\\

\textbf{\Acrlong{lr}} is a fundamental statistical model used for binary classification tasks~\cite{harris_primer_2021}. It estimates the probability of an outcome by applying the logistic function to a linear combination of input features. An architecture diagram of this model is presented in Supplementary Figure~\ref{fig:feature_models}a).
The most common parameters across all folds were: learning rate of 0.01, Adam optimiser, weight decay of 0.01 and 0.00001, and batch size of 64.\\

\textbf{\Acrlong{xgb}} is a powerful ensemble learning algorithm based on gradient boosting decision trees~\cite{chen_xgboost_2016}. It employs advanced techniques such as regularisation, tree pruning, and parallel computation to enhance performance and mitigate over-fitting. An architecture diagram of this model is presented in the Supplementary Figure \ref{fig:feature_models} b). The most common parameters across all folds were: learning rate of 0.03, number of estimators of 300 and 400, number of parallel trees of 900, max delta step of 0.1668, gamma of 0.0 or 0.0002, lambda of 0.0215, minimum child weight of 278 and 1668, subsample of 0.55, and colsample by tree of 0.4481 and 0.5123, and batch size of 160 and 32. Other parameters were non-conclusive.\\

\textbf{\Acrlong{mlp}} is an artificial neural network composed of multiple layers of interconnected neurons~\cite{fully_connected_deep_network}. It utilises nonlinear activation functions and back-propagation for training, enabling it to capture complex patterns in data. An architecture diagram of this model is presented in the Supplementary Figure \ref{fig:feature_models} c). 
The most common parameters across all folds were: learning rate of 0.0001, Adam optimiser, weight decay of 0.001 and 0.0000001, batch size of 64 and 128, and hidden dimensions 1, 2, and 3 of 512, 256, 128 or 64, respectively.

\textbf{\Acrlong{cnn}} are deep learning models designed to process spatially structured data, particularly images~\cite{lecun_gradient-based_1998}. They employ convolutional layers to extract hierarchical features, followed by pooling layers to reduce dimensionality and fully connected layers for classification. An architecture diagram of this model is presented in the Supplementary Figure \ref{fig:cnn_model}. The most common parameters across all folds were: learning rate of 0.00001, Adam optimiser, weight decay of 0.0000001, batch size of 128, the first convolution layer of 512 and kernel of 4, max pool of 2, the second convolution layer of 256 with kernel of 3, the third convolution later of 128 and 64 with kernel of 3, and the fully connected layer of 128. \\

\textbf{\Acrshort{eegnet}} is a compact and efficient deep learning architecture tailored for \gls{eeg} signal analysis~\cite{lawhern_eegnet_2018}. It incorporates depthwise and separable convolutions to capture both spatial and temporal features while maintaining low computational complexity. \gls{eegnet} has demonstrated strong performance in \gls{bci} applications and \gls{eeg}-based classification tasks, offering a balance between accuracy and model efficiency. An architecture diagram of this model is presented in the Supplementary Figure \ref{fig:eegnet_model}. This model's hyperparameters are defined by the authors and therefore are not tunable.\\

\textbf{ConvLSTM} model integrates \gls{cnn} with \gls{lstm} networks to leverage both spatial and temporal dependencies in sequential data~\cite{liu_automatic_2020}. \Gls{cnn}s extract high-level features, which are subsequently processed by \gls{lstm}s to capture long-term dependencies. This hybrid approach is particularly effective for time series analysis, including medical signal classification and speech recognition~\cite{zhao_convolutional_2017}. An architecture diagram of this model is presented in the Supplementary Figure \ref{fig:cnnlstm_model}. The most common parameters across all folds were: Adam optimiser, weight decay of 0.0001, batch size of 64, the first convolution layer of 512 (kernel size inconclusive), max pool of 2, the second convolution layer of 256 with kernel of 5, the third convolution later of 128 and 64 with kernel of 5, the number of \gls{lstm} layers of 2, the hidden dimension of \gls{lstm} of 128, the first fully connected layer of 128, and the second fully connected layer of 64 and 32. The value of the learning rate was inconclusive.\\

\textbf{ConvTransformer} model combines the feature extraction capabilities of \gls{cnn}s with the self-attention mechanism of transformers~\cite{li_eeg-based_2022}. \Gls{cnn}s encode local spatial patterns, while the transformer component captures long-range dependencies, enhancing the model’s ability to process complex sequential data~\cite{bougourzi_d-trattunet_2024}. An architecture diagram of this model is presented in the Supplementary Figure \ref{fig:cnntransformer_model}. The most common parameters across all folds were: Adam optimiser, weight decay of 0.000001, batch size of 32, the learning rate of 0.0001 and 0.00001, the first convolution layer of 1024 with kernel size of 3, max pool of 2, the second convolution layer of 256 and 128 with kernel of 3 and 4, the third convolution later of 64 with kernel of 5, the vocabulary size of 5500, the feed forward layer of 1024, number of heads of 6, number of encoder layers of 4 and 5, and model dimension of 300.\\

\subsection{Post-processing}\label{sec:post-processing}
\begin{figure}[h]
    \centering
    \includegraphics[angle=0,origin=c,width=130mm]{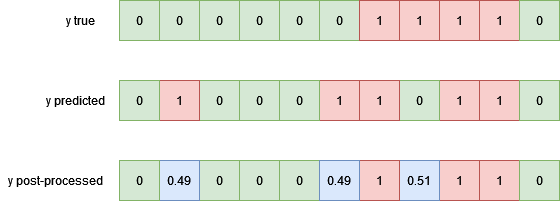}
    \caption{Diagram illustrating the effect of post-processing on a signal. The figure shows three signals: the true signal (y true), the predicted signal (y predicted), and the post-processed result (y post-processed). In the post-processed signal, highlighted regions show the effect of post-processing on an isolated epoch, a drift, and a gap.  Labels ``1'' and ``0'' denote an event and background, respectively.}
    \label{fig:post-processing}
\end{figure}

Our framework incorporates a post-processing module designed to make minimal adjustments, targeting single-epoch artefacts such as drift, isolated epochs, or gaps in consecutive series. This method emulates human correction by addressing minor misalignments and clear errors without significantly altering the predictions, which comprise a mixture of \gls{epoch} and \gls{ovlp}~\cite{obeid_objective_2021}. We continue to analyse the recording at high granularity, treating each epoch as a standalone signature rather than considering the entire event as a single unit. However, we permit minimal misalignment or gaps where prediction overlap is high. For example, if a single-epoch disruption has a predicted value of 0.23 but the true label is 1 and the neighbouring epochs are confidently positive, the post-processing module may raise it to 0.51. This results in a correct binary classification without overstating the model's certainty, thus preserving metric integrity. The goal is to improve temporal consistency while ensuring that adjusted values remain close to the decision threshold, reflecting a nuanced correction rather than artificial performance inflation. Figure \ref{fig:post-processing} illustrates these subtle but meaningful refinements. Importantly, the module remains optional and is disabled by default to maintain analytical transparency.

These refinements, illustrated in Figure~\ref{fig:post-processing}, more closely reflect human judgment and support a nuanced enhancement of the overall analysis. The post-processing module operates on an epoch-wise basis, enabling highly granular classification. While this granularity can lead to an increased number of false alarms, it supports more precise estimates of sensitivity and specificity than event-based post-processing methods~\cite{obeid_objective_2021}.

\section{Conclusion}\label{sec:conclusions}
In this work, we introduced \textbf{PySeizure}, a machine learning-based framework for automated epileptic seizure detection. Our models achieved high within-dataset performance (\gls{auc} 0.904$\pm$0.059 and 0.864$\pm$0.060) and demonstrated strong generalisation across datasets, despite differences in experimental set-up and patient populations, with \gls{auc} values of 0.762$\pm$0.175 and 0.615$\pm$0.039. These results were obtained without any post-processing, highlighting the robustness and adaptability of PySeizure in varying clinical settings. The framework’s design prioritises generalisability, reproducibility, and ease of integration into existing workflows, addressing key challenges faced by current automated systems and offering a scalable solution for epilepsy management.
To further enhance performance, we applied mild post-processing based on a combination of \gls{epoch} and \gls{ovlp} strategies~\cite{obeid_objective_2021}, resulting in improved \gls{auc} scores of 0.913$\pm$0.064 and 0.867$\pm$0.058 within-dataset and 0.768$\pm$0.172 and 0.619$\pm$0.036 cross-datasets. Notably, PySeizure employs a voting-based model ensemble, justified by the observed occasional lack of agreement between individual models. In complex classification settings such as seizure detection -- where subtle patterns in \gls{eeg} can lead models to diverge in their predictions -- this disagreement is not a weakness, but rather an opportunity: aggregating predictions through a voting mechanism allows the system to exploit complementary strengths of individual models, improving overall robustness and reducing the risk of overfitting to dataset-specific noise.

Future directions include extending PySeizure's evaluation with hospital-acquired data to enhance its clinical applicability. We also plan to expand testing across more diverse datasets. This will provide further insights into the system’s performance in real-world clinical environments. Future work will also explore real-time performance improvements and the potential integration with wearable devices for continuous monitoring. As PySeizure progresses, it holds promise to bridge the gap between state-of-the-art \gls{ai} research and tangible clinical applications, driving advancements in the management of epilepsy and potentially other neurological disorders.

\backmatter

\clearpage

\newpage

\section*{Declarations}

\begin{itemize}
\item Funding

\textit{Not applicable}
\item Conflict of interest/Competing interests

\textit{Not applicable}
\item Ethics approval and consent to participate

\textit{Datasets used in this study are publicly available and were accessed in accordance with their respective data use agreements. Ethical approval for this work was obtained from the University of Edinburgh School of Engineering.
}
\item Consent for publication

\textit{Not applicable}
\item Data availability 

\textit{Datasets used in this study are publicly available and were accessed in accordance with their respective data use agreements.}

\begin{itemize}
    \item{\href{https://isip.piconepress.com/projects/tuh\_eeg/}{TUH EEG Seizure Detection Corpus}}
    \item{\href{https://physionet.org/content/chbmit/1.0.0/}{CHB-MIT Scalp EEG Database}}
\end{itemize}

\item Materials availability

\textit{Not applicable}
\item Code availability 

\textit{The code for PySeizure framework is publicly available at \href{https://github.com/bartlomiej-chybowski/PySeizure}{https://github.com/bartlomiej-chybowski/PySeizure}}
\item Author contribution

\textit{B.C.: Conceptualisation, Methodology, Software, Validation, Formal analysis, Investigation, Writing – original draft.
S.A., H.H.: Methodology, Validation, Writing – review \& editing.
A.G., J.E.: Supervision, Validation, Methodology, Project administration, Writing – review \& editing.
All authors reviewed and approved the final manuscript.}
\end{itemize}

\section*{Acknowledgements}

For the purpose of open access, the author has applied a Creative Commons Attribution (CC BY) licence to any Author Accepted Manuscript version arising from this submission.

\noindent

\newpage
\bibliography{sn-bibliography}
\newpage

\begin{appendices}

\section{Figures}\label{sec:supplementary:figures}
\begin{figure}[!h]
    \centering
    \includegraphics[angle=0,origin=c,width=130mm]{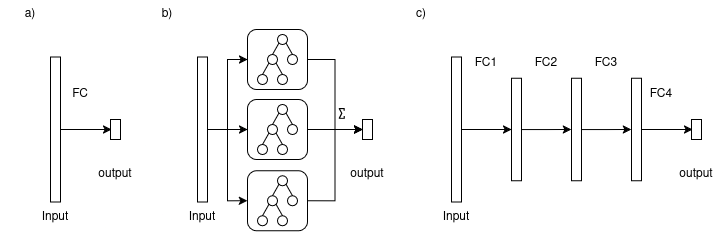}
    \caption{Diagram of the a) \Acrfull{lr}, b) \Acrfull{xgb}, c) \Acrfull{mlp} models. The size of the Fully Connected (FC 1-4) layers and XGBoost parameters are optimised by the hyperparameter optimisation algorithm.}
    \label{fig:feature_models}
\end{figure}

\begin{figure}[!h]
    \centering
    \includegraphics[angle=0,origin=c,width=130mm]{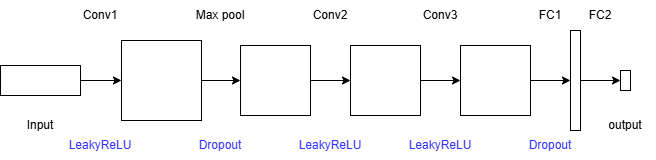}
    \caption{Diagram of the \Acrfull{cnn} model architecture. The size of the convolution (Conv 1-3), Max Pool (MP), and fully connected (FC) layers is optimised by the hyperparameter optimisation algorithm.}
    \label{fig:cnn_model}
\end{figure}

\begin{figure}[h]
    \centering
    \includegraphics[angle=0,origin=c,width=130mm]{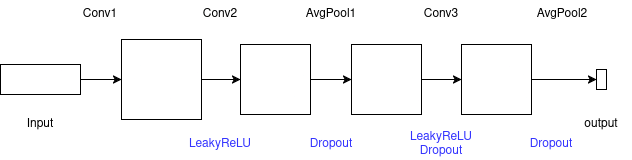}
    \caption{Diagram of the \acrshort{eegnet} model architecture. All parameters are predefined according to the original implementation.}
    \label{fig:eegnet_model}
\end{figure}

\begin{figure}[h]
    \centering
    \includegraphics[angle=0,origin=c,width=130mm]{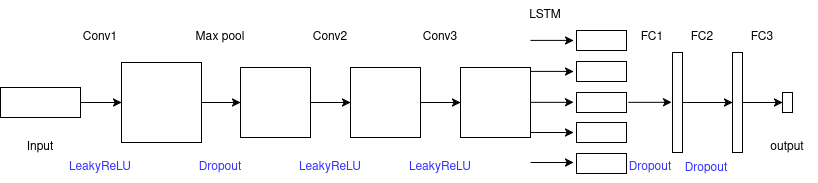}
    \caption{Diagram of the \acrshort{convlstm} model architecture. The size of the convolution (Conv 1-3), Max Pool (MP), LSTM and fully connected (FC 1 and 2) layers is optimised by the hyperparameter optimisation algorithm.}
    \label{fig:cnnlstm_model}
\end{figure}

\begin{figure}[h]
    \centering
    \includegraphics[angle=0,origin=c,width=130mm]{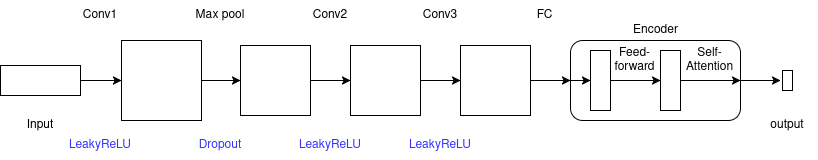}
    \caption{Diagram of the \acrshort{convtransformer} model architecture. The hyperparameter optimisation algorithm optimises the size of the convolution (Conv 1-3), Max Pool (MP) layers, and encoder parameters.}
    \label{fig:cnntransformer_model}
\end{figure}

\begin{figure}[hp]
    \centering
    \includegraphics[angle=0,origin=c,width=130mm]{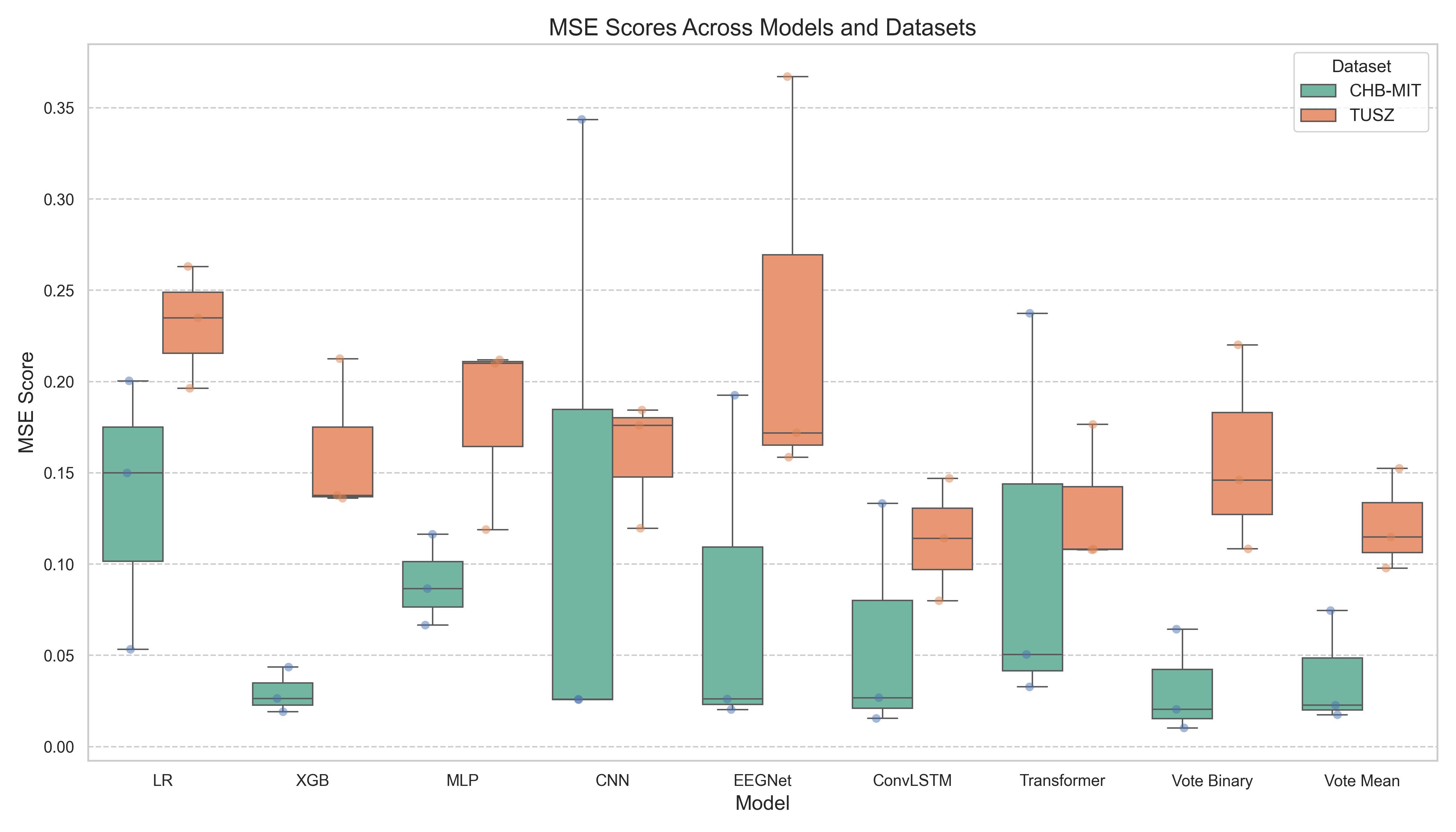}
    \caption{Comparison of \Acrfull{mse} scores for \Acrfull{chb-mit} and \Acrfull{tusz} datasets across all the models.}
    \label{fig:within_mse}
\end{figure}
\begin{figure}[h]
    \centering
    \includegraphics[angle=0,origin=c,width=130mm]{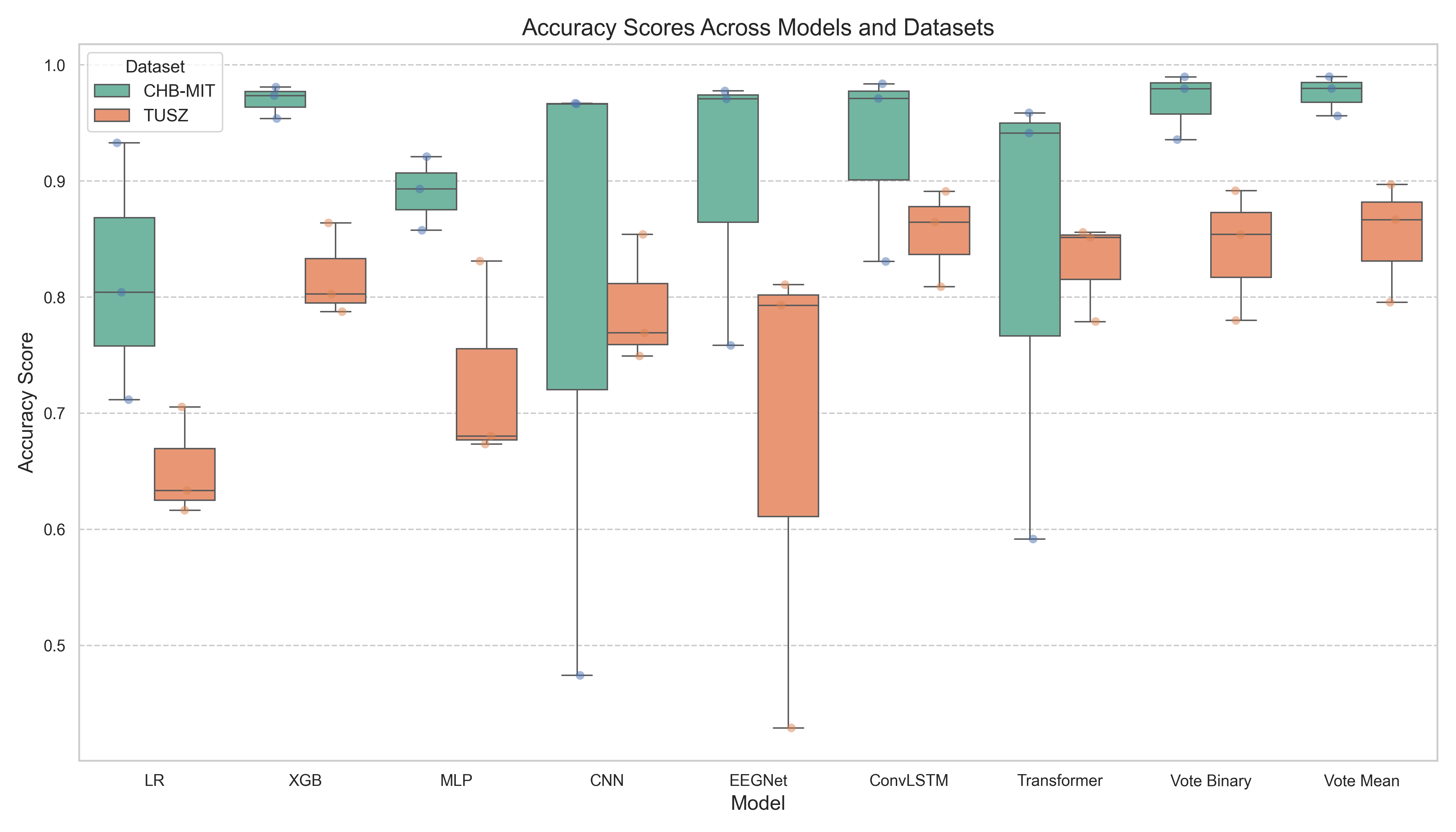}
    \caption{Comparison of Accuracy scores for \Acrfull{chb-mit} and \Acrfull{tusz} datasets across all the models.}
    \label{fig:within_accuracy}
\end{figure}
\begin{figure}[h]
    \centering
    \includegraphics[angle=0,origin=c,width=130mm]{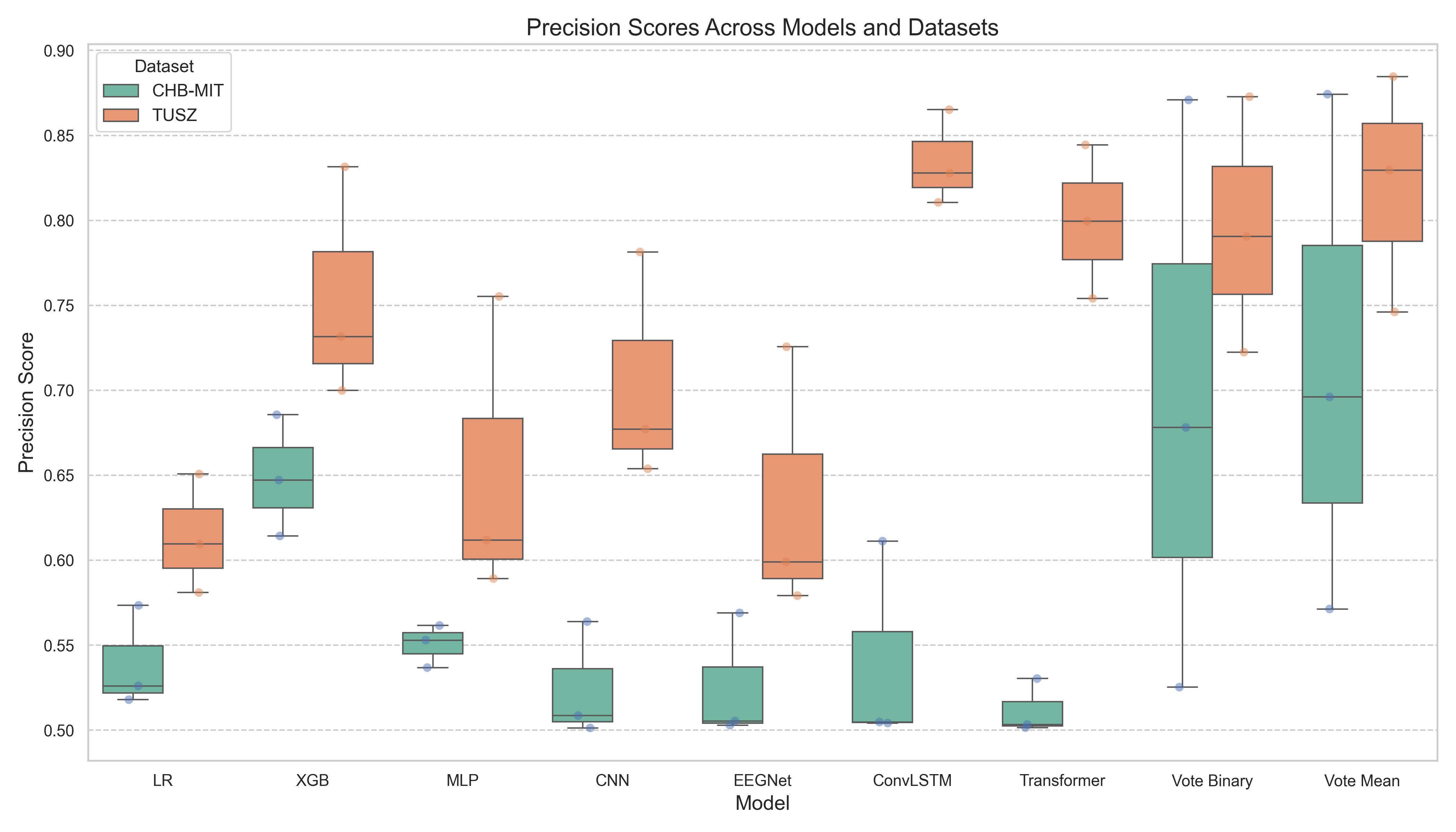}
    \caption{Comparison of Precision scores for \Acrfull{chb-mit} and \Acrfull{tusz} datasets across all the models.}
    \label{fig:within_precision}
\end{figure}
\begin{figure}[h]
    \centering
    \includegraphics[angle=0,origin=c,width=130mm]{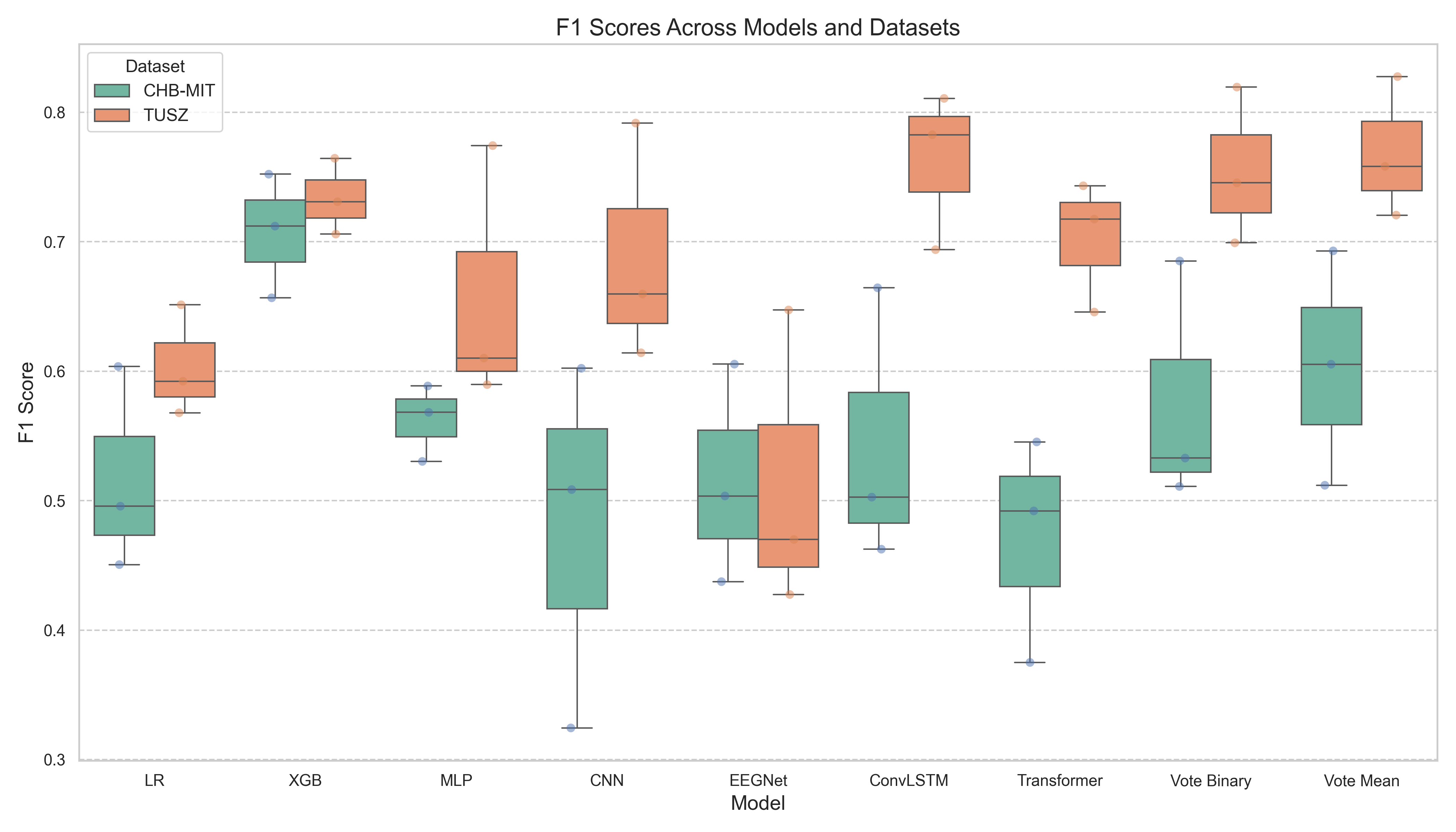}
    \caption{Comparison of F1 scores for \Acrfull{chb-mit} and \Acrfull{tusz} datasets across all the models.}
    \label{fig:within_f1}
\end{figure}
\begin{figure}[h]
    \centering
    \includegraphics[angle=0,origin=c,width=130mm]{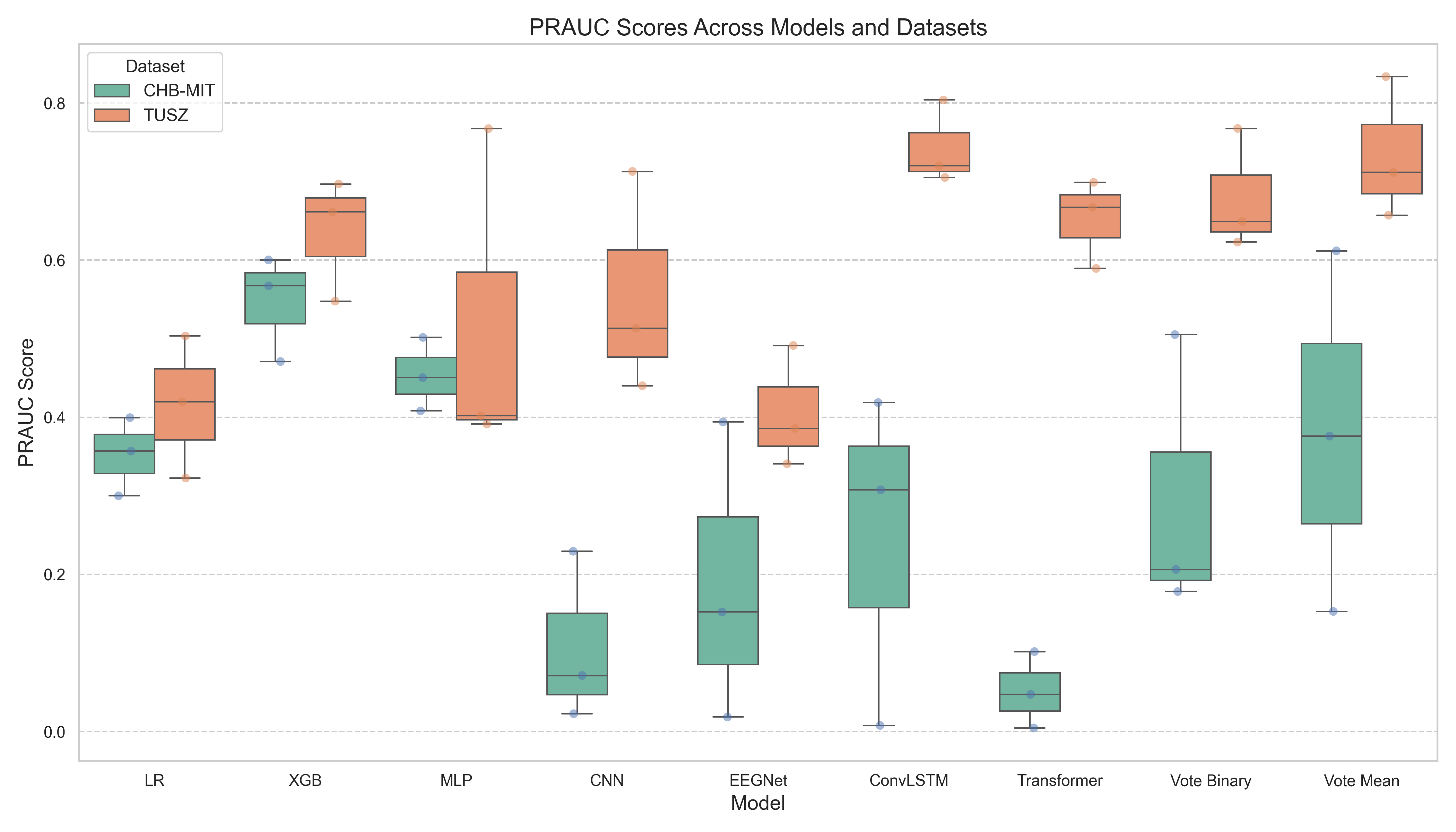}
    \caption{Comparison of \Acrfull{prauc} scores for \Acrlong{chb-mit} and \Acrlong{tusz} datasets across all the models.}
    \label{fig:within_prauc}
\end{figure}
\begin{figure}[h]
    \centering
    \includegraphics[angle=0,origin=c,width=130mm]{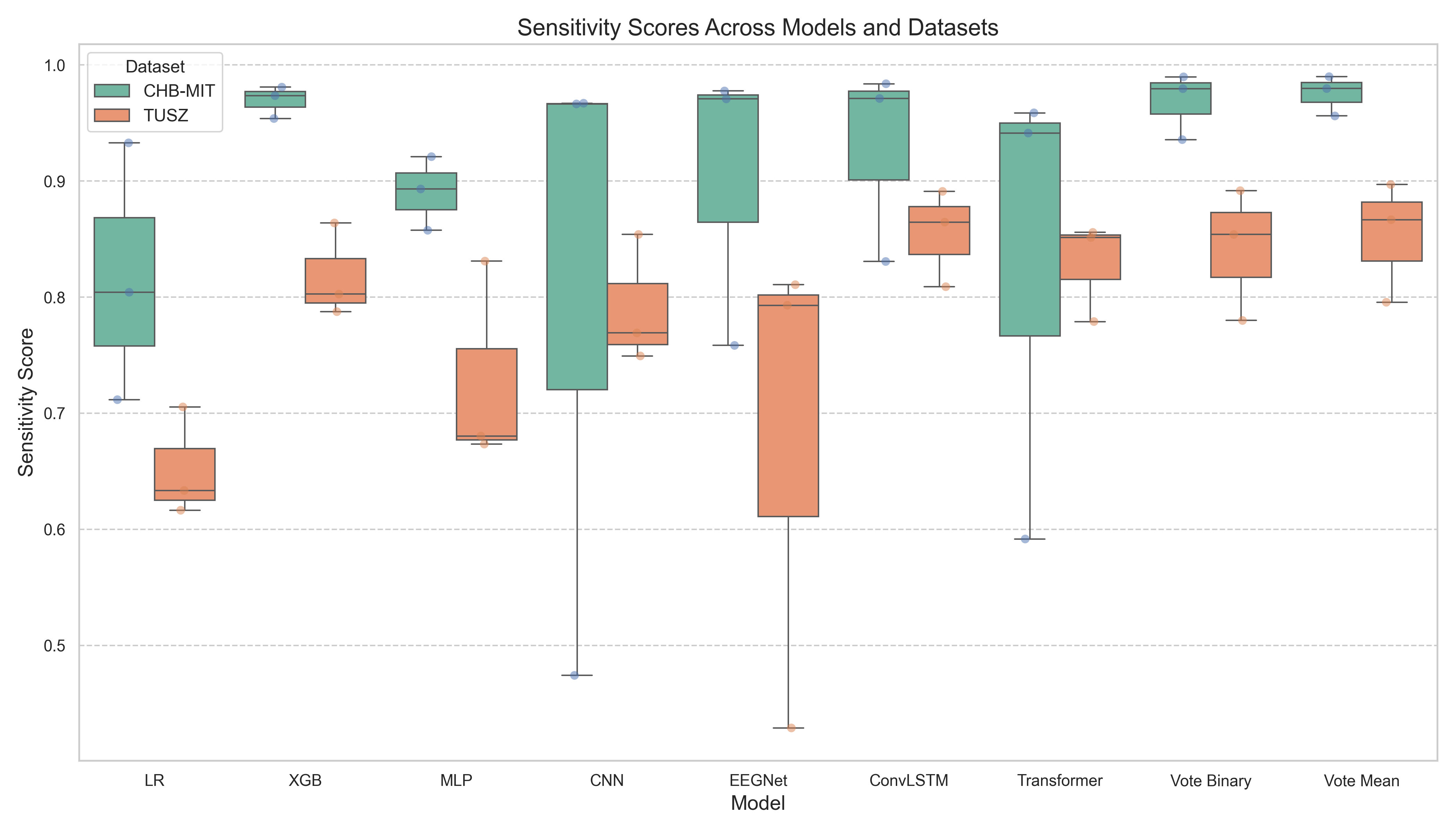}
    \caption{Comparison of Sensitivity scores for \Acrfull{chb-mit} and \Acrfull{tusz} datasets across all the models.}
    \label{fig:within_sensitivity}
\end{figure}
\begin{figure}[h]
    \centering
    \includegraphics[angle=0,origin=c,width=130mm]{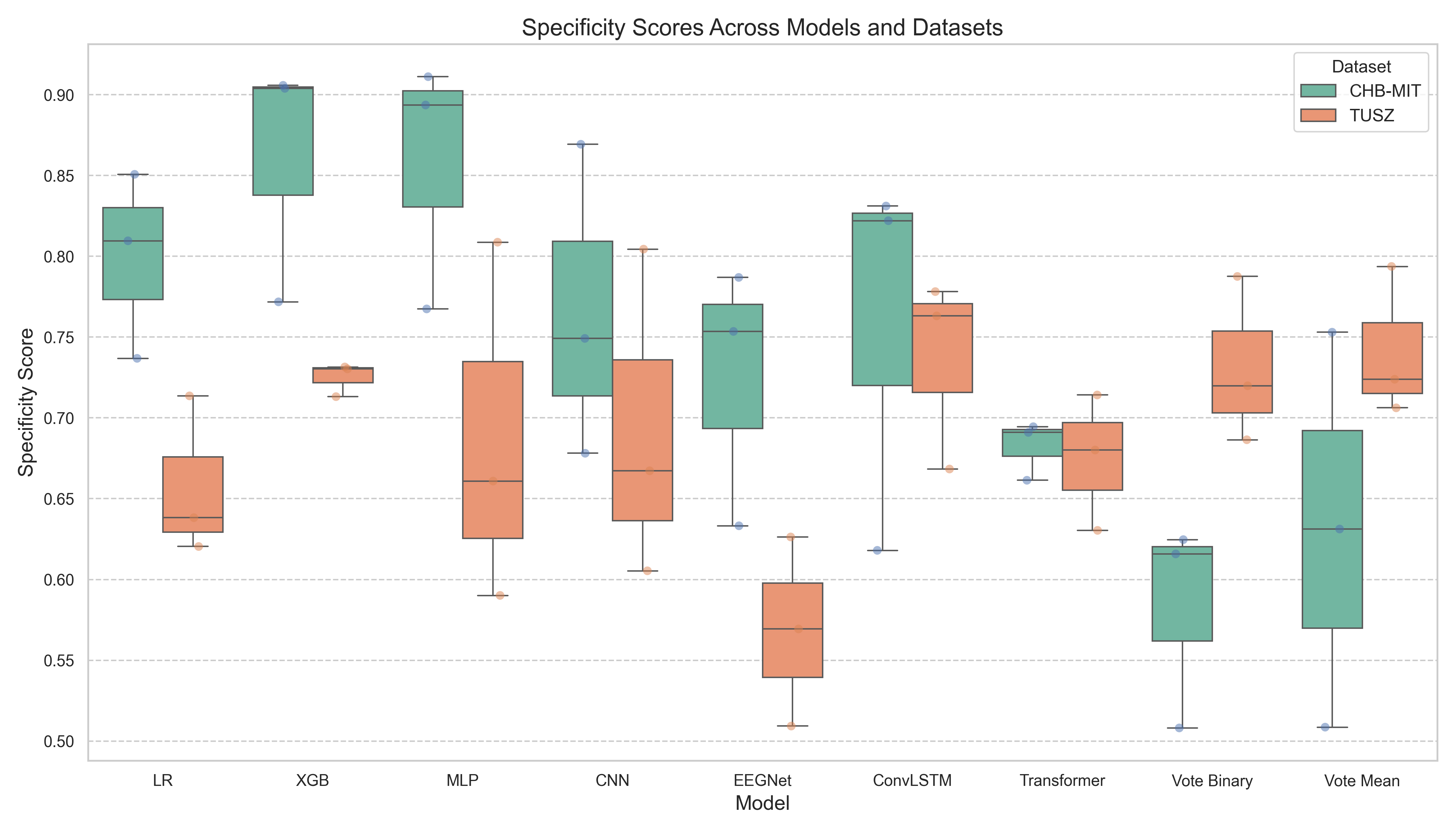}
    \caption{Comparison of Specificity scores for \Acrfull{chb-mit} and \Acrfull{tusz} datasets across all the models.}
    \label{fig:within_specificity}
\end{figure}

\begin{figure}[hp]
    \centering
    \includegraphics[angle=0,origin=c,width=130mm]{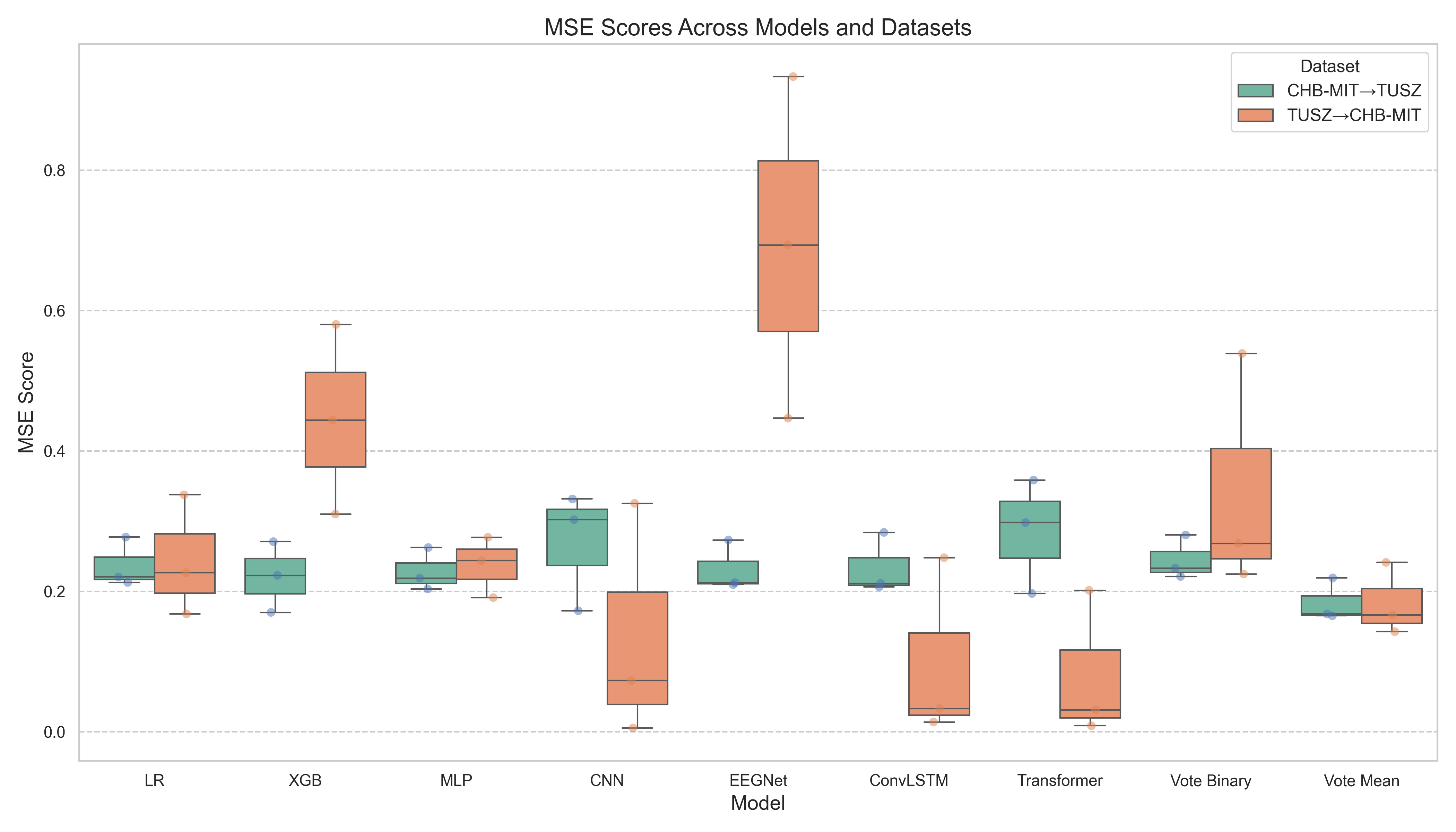}
    \caption{Comparison of \Acrfull{mse} scores for all the models trained on \Acrfull{chb-mit} and evaluated on \Acrfull{tusz} datasets and vice versa.}
    \label{fig:within_mse}
\end{figure}
\begin{figure}[h]
    \centering
    \includegraphics[angle=0,origin=c,width=130mm]{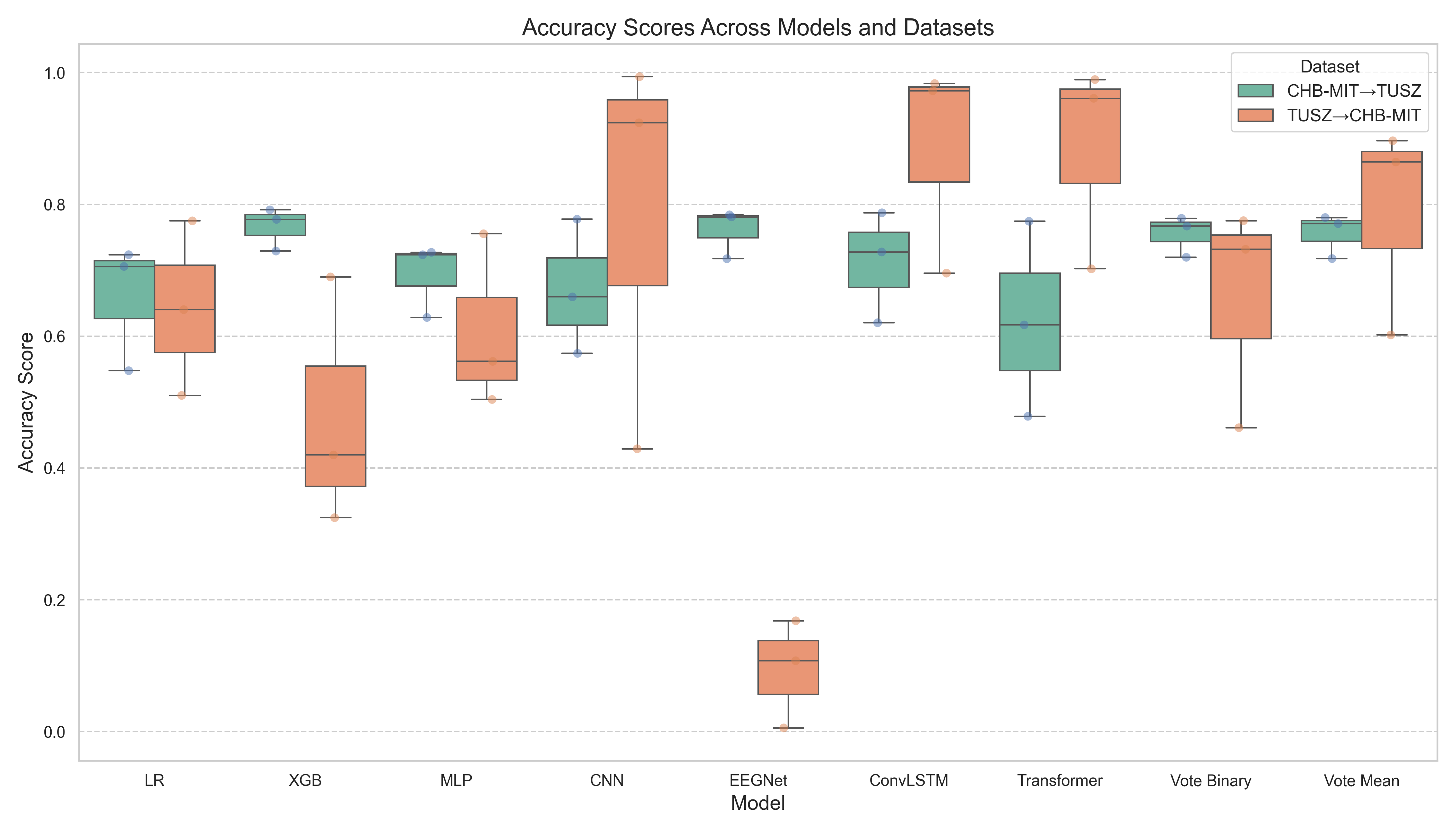}
    \caption{Comparison of Accuracy scores for all the models trained on \Acrfull{chb-mit} and evaluated on \Acrfull{tusz} datasets and vice versa.}
    \label{fig:within_accuracy}
\end{figure}
\begin{figure}[h]
    \centering
    \includegraphics[angle=0,origin=c,width=130mm]{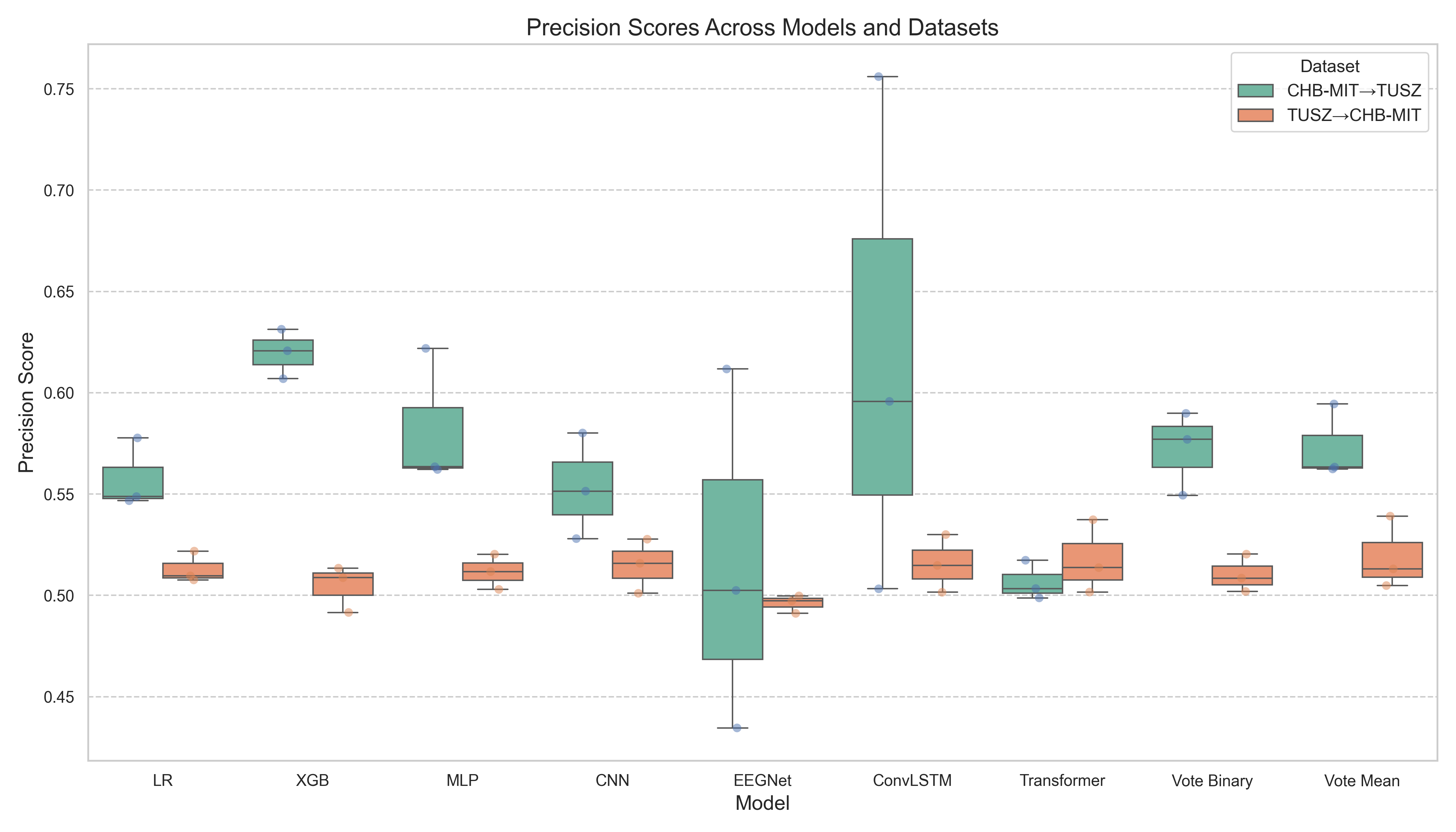}
    \caption{Comparison of Precision scores for all the models trained on \Acrfull{chb-mit} and evaluated on \Acrfull{tusz} datasets and vice versa.}
    \label{fig:within_precision}
\end{figure}
\begin{figure}[h]
    \centering
    \includegraphics[angle=0,origin=c,width=130mm]{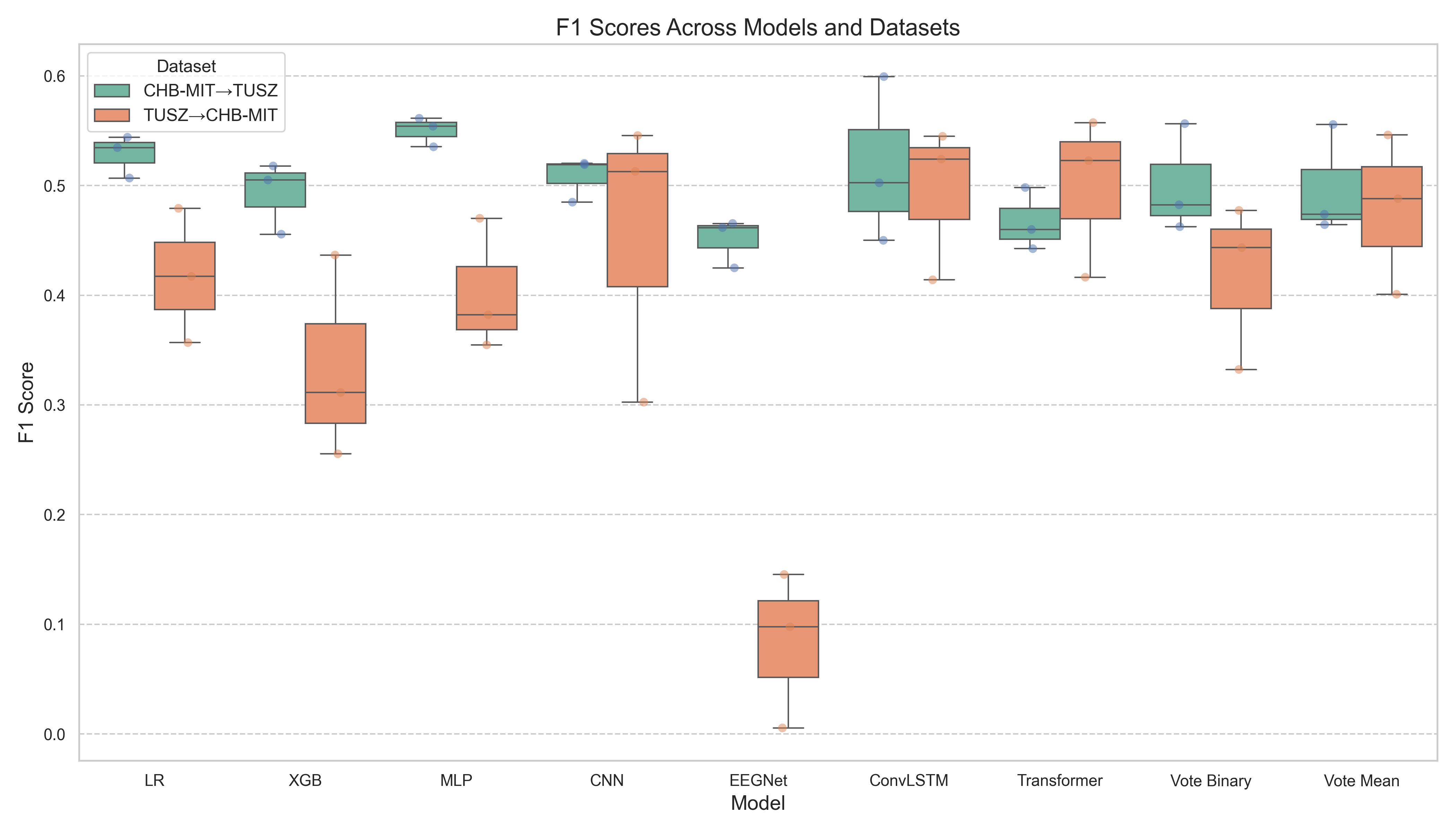}
    \caption{Comparison of F1 scores for all the models trained on \Acrfull{chb-mit} and evaluated on \Acrfull{tusz} datasets and vice versa.}
    \label{fig:within_f1}
\end{figure}
\begin{figure}[h]
    \centering
    \includegraphics[angle=0,origin=c,width=130mm]{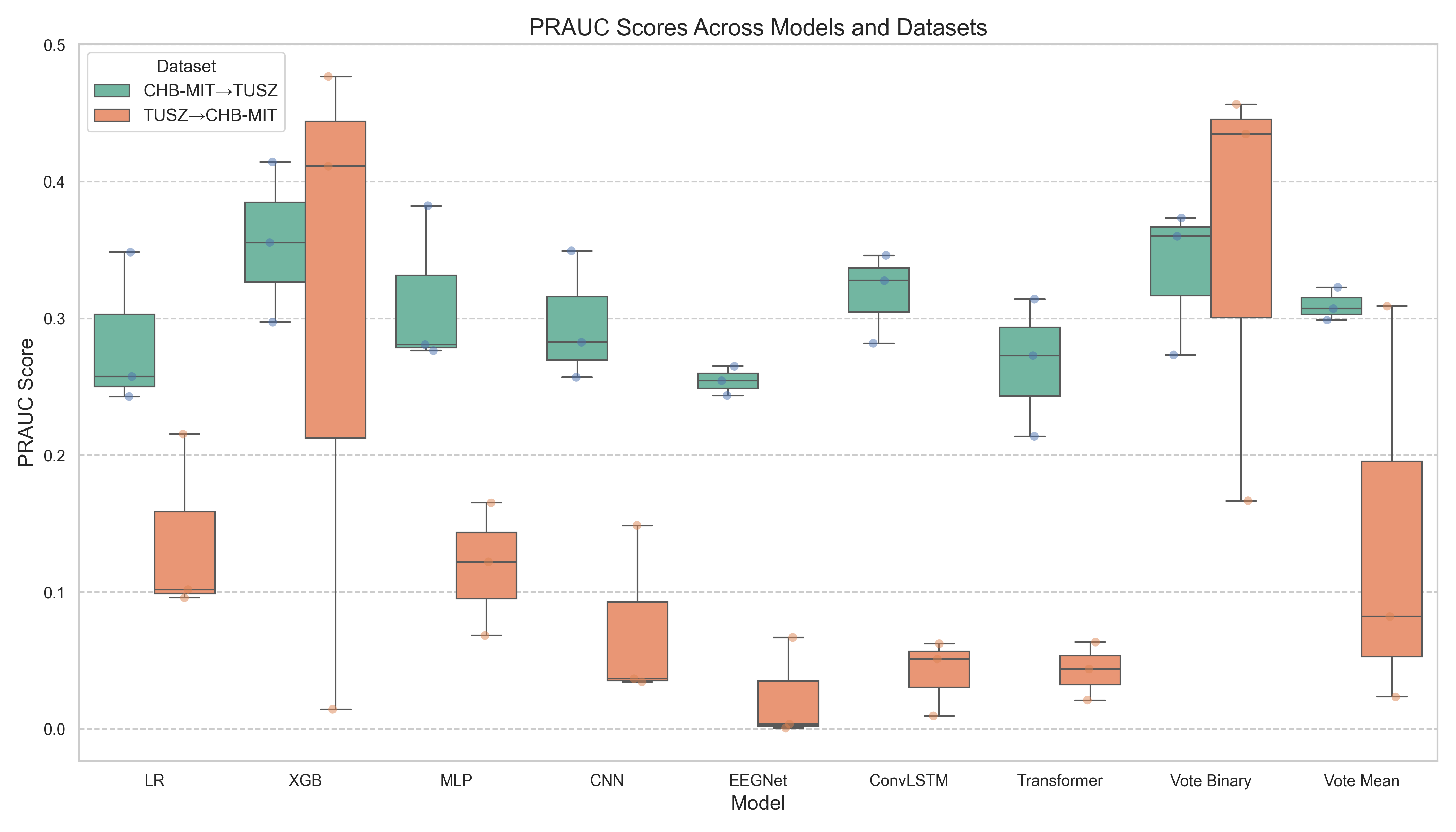}
    \caption{Comparison of \Acrfull{prauc} scores for all the models trained on \Acrfull{chb-mit} and evaluated on \Acrfull{tusz} datasets and vice versa.}
    \label{fig:within_prauc}
\end{figure}
\begin{figure}[h]
    \centering
    \includegraphics[angle=0,origin=c,width=130mm]{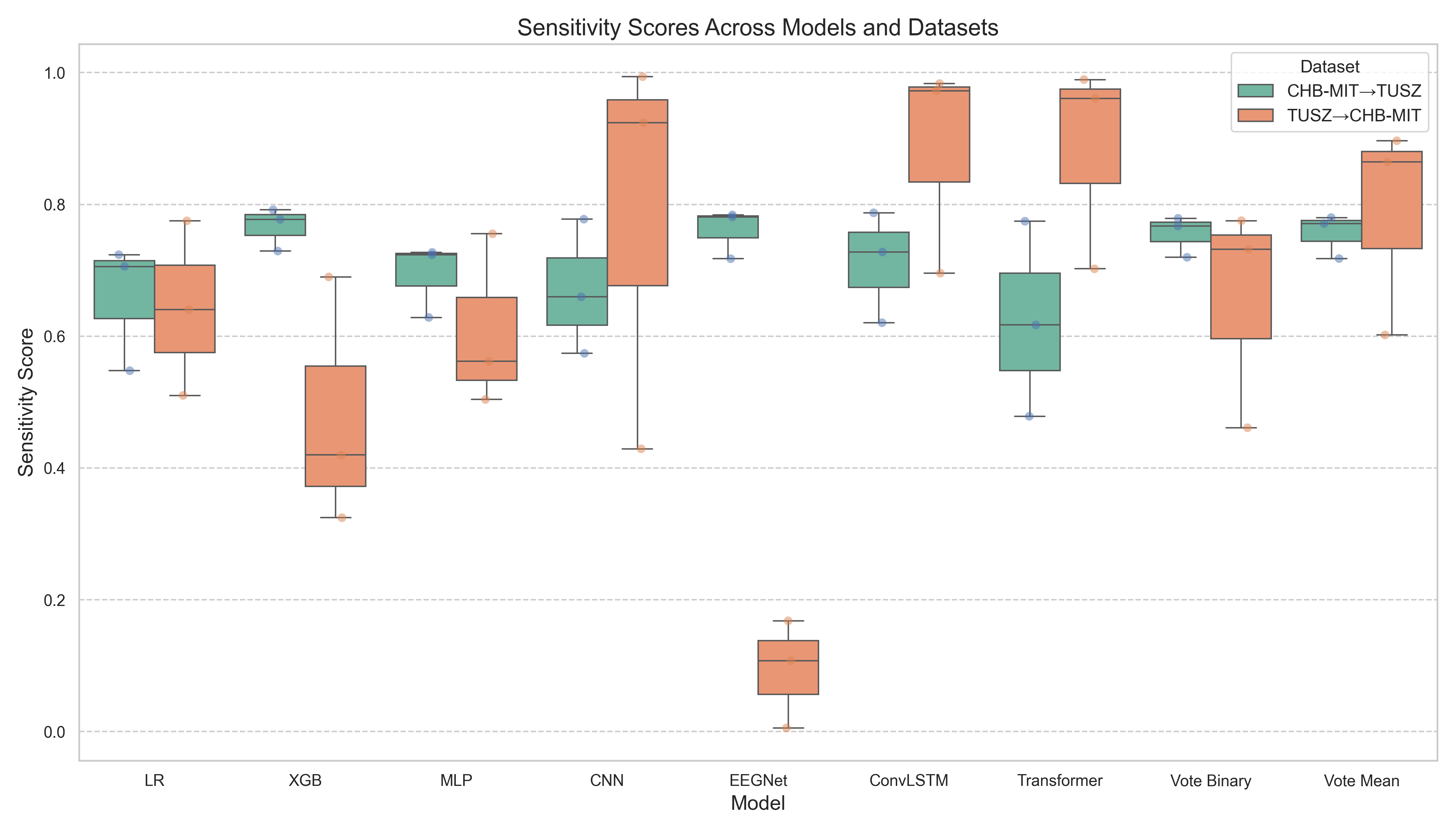}
    \caption{Comparison of Sensitivity scores for all the models trained on \Acrfull{chb-mit} and evaluated on \Acrfull{tusz} datasets and vice versa.}
    \label{fig:within_sensitivity}
\end{figure}
\begin{figure}[h]
    \centering
    \includegraphics[angle=0,origin=c,width=130mm]{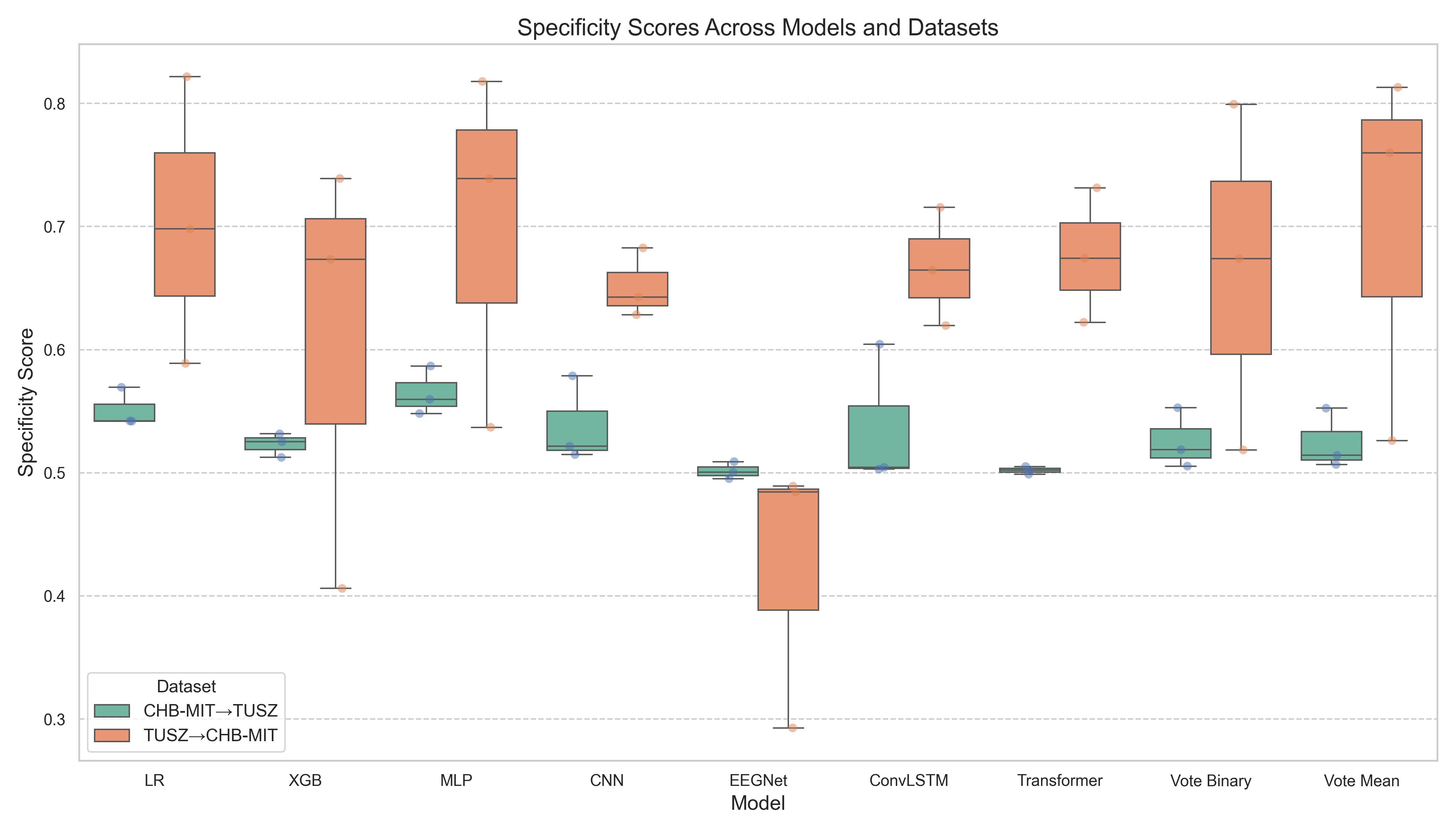}
    \caption{Comparison of Specificity scores for all the models trained on \Acrfull{chb-mit} and evaluated on \Acrfull{tusz} datasets and vice versa.}
    \label{fig:within_specificity}
\end{figure}

\begin{figure}[h]
    \centering
    \includegraphics[angle=0,origin=c,width=130mm]{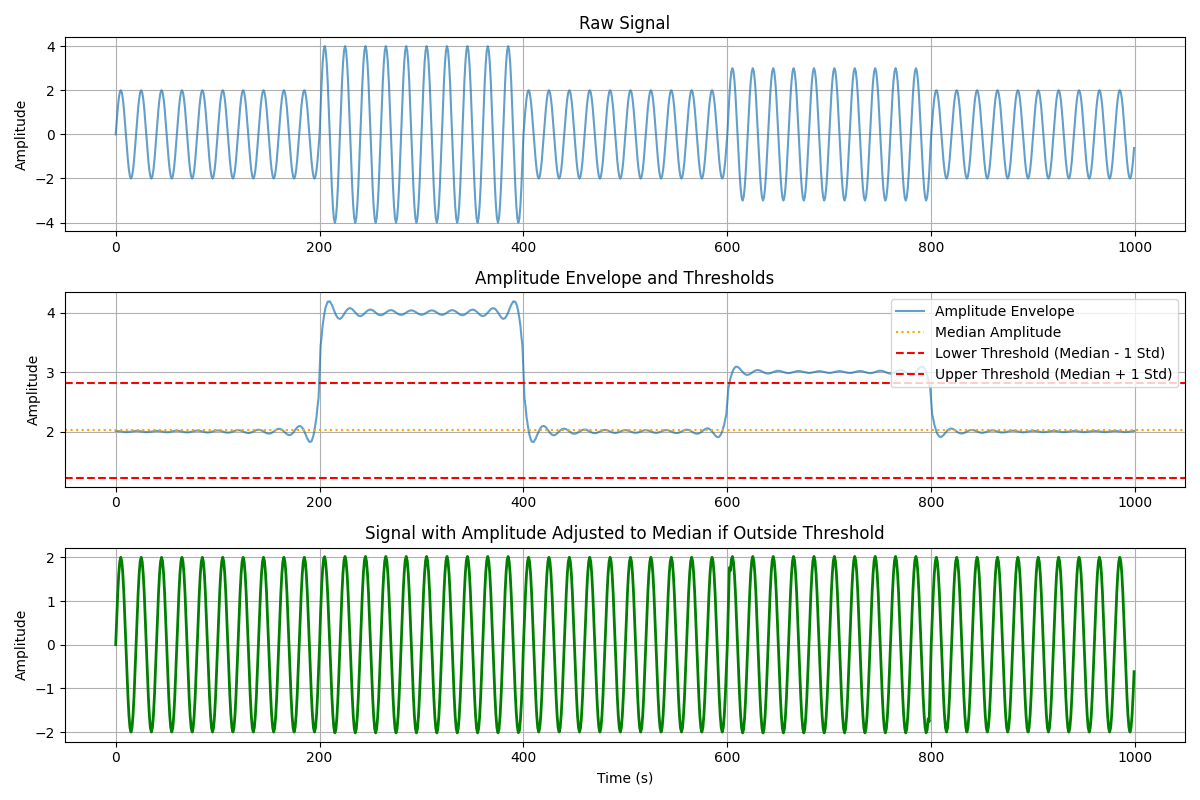}
    \caption{Threshold-based smoothing of an artificial signal with the threshold level reduced to 1 standard deviation for illustrative purposes. The top plot shows the raw, unprocessed signal. The middle plot shows the amplitude envelope with an empirically derived threshold. The bottom plot shows the processed signal, where out-of-threshold amplitudes are normalised to the median value, preserving in-range dynamics.}
    \label{fig:smoothing_supplementary}
\end{figure}

\begin{figure}[!h]
    \centering
    \includegraphics[angle=0,origin=c,width=110mm]{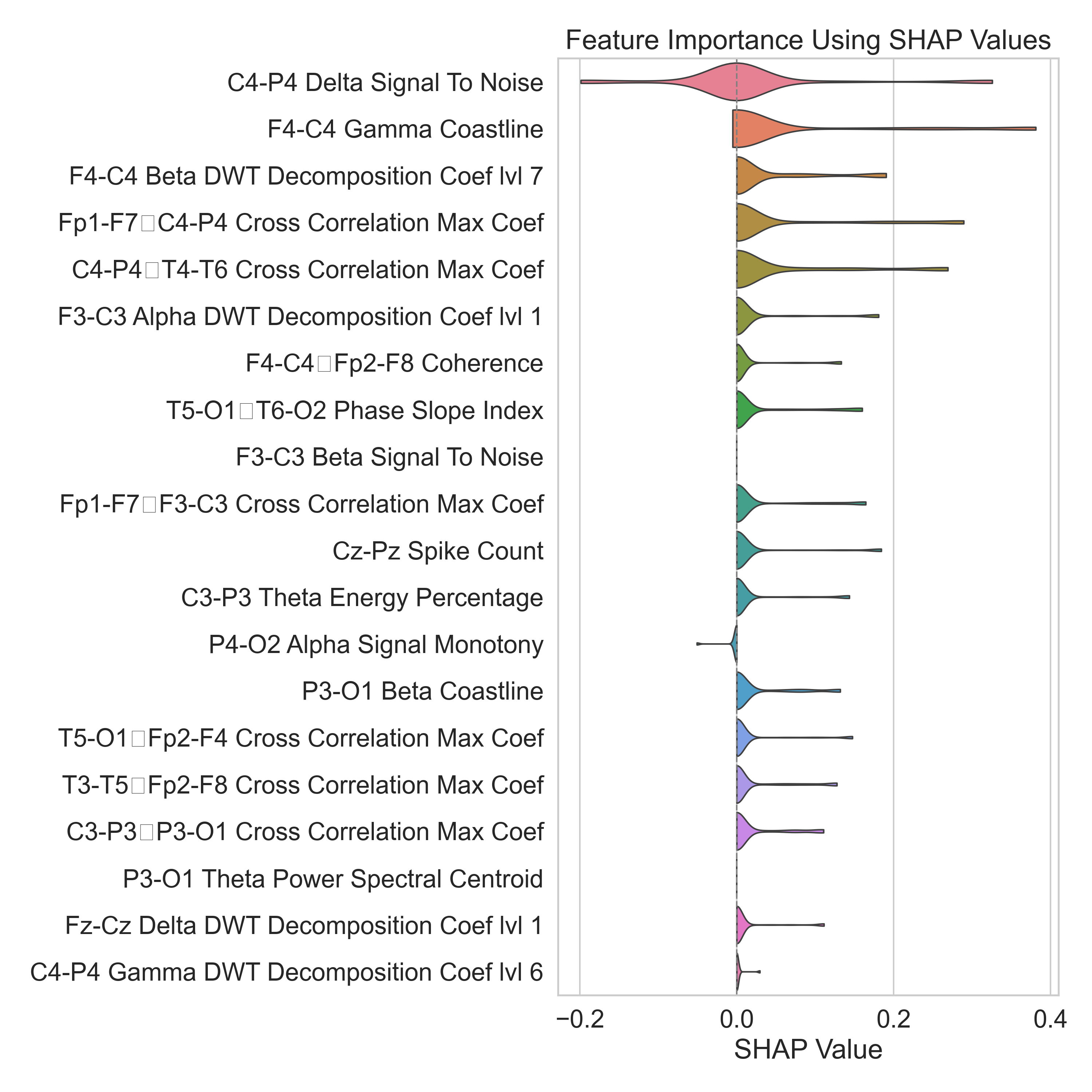}
    \caption{Global feature importance derived from \acrshort{shap} values, showing the top twenty most influential features for the \acrshort{lr} model trained on \Acrfull{tusz} dataset and evaluated on \Acrshort{tusz} using engineered features.}
    \label{fig:feature-shap_LR_tusz_tusz}
\end{figure}
\begin{figure}[!h]
    \centering
    \includegraphics[angle=0,origin=c,width=110mm]{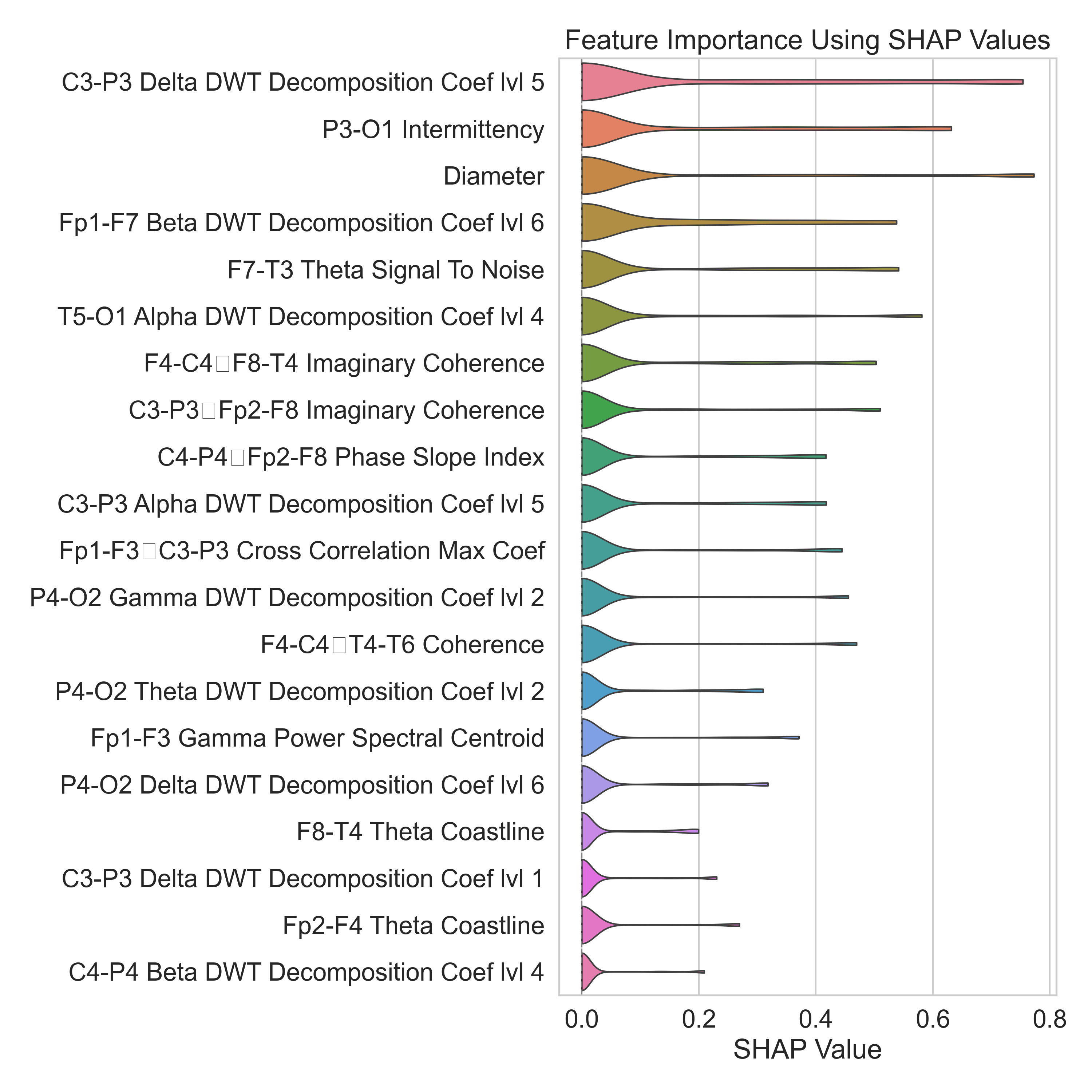}
    \caption{Global feature importance derived from \acrshort{shap} values, showing the top twenty most influential features for the \acrshort{lr} model trained on \Acrfull{chb-mit} dataset and evaluated on \Acrshort{chb-mit} using engineered features.}
    \label{fig:feature-shap_LR_chbmit_chbmit}
\end{figure}
\begin{figure}[!h]
    \centering
    \includegraphics[angle=0,origin=c,width=110mm]{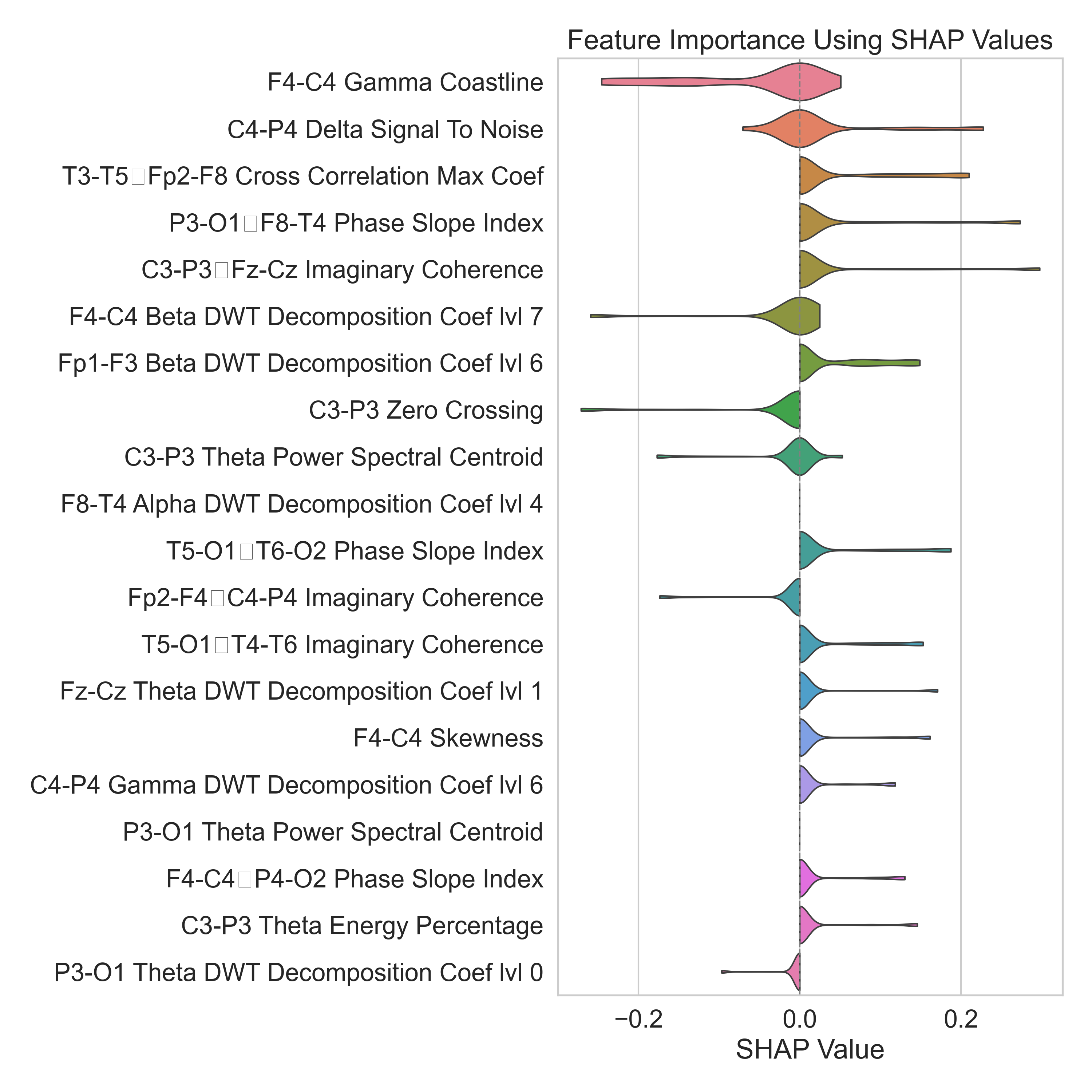}
    \caption{Global feature importance derived from \acrshort{shap} values, showing the top twenty most influential features for the \acrshort{lr} model trained on \Acrfull{tusz} dataset and evaluated on \Acrfull{chb-mit} using engineered features.}
    \label{fig:feature-shap_LR_tusz_chmit}
\end{figure}
\begin{figure}[!h]
    \centering
    \includegraphics[angle=0,origin=c,width=110mm]{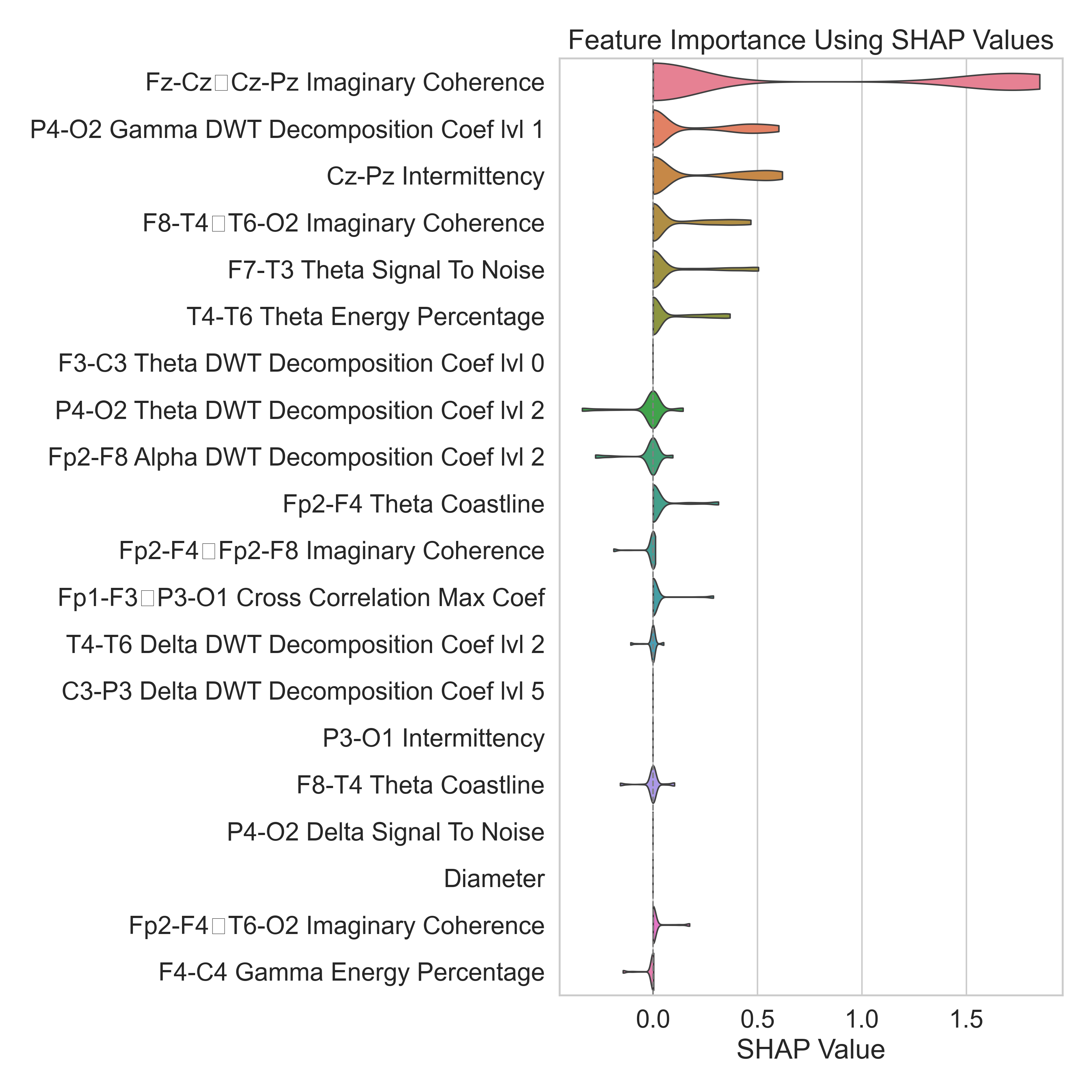}
    \caption{Global feature importance derived from \acrshort{shap} values, showing the top twenty most influential features for the \acrshort{lr} model trained on \Acrfull{chb-mit} dataset and evaluated on \Acrfull{tusz} using engineered features.}
    \label{fig:feature-shap_LR_chbmit_tusz}
\end{figure}

\begin{figure}[!h]
    \centering
    \includegraphics[angle=0,origin=c,width=110mm]{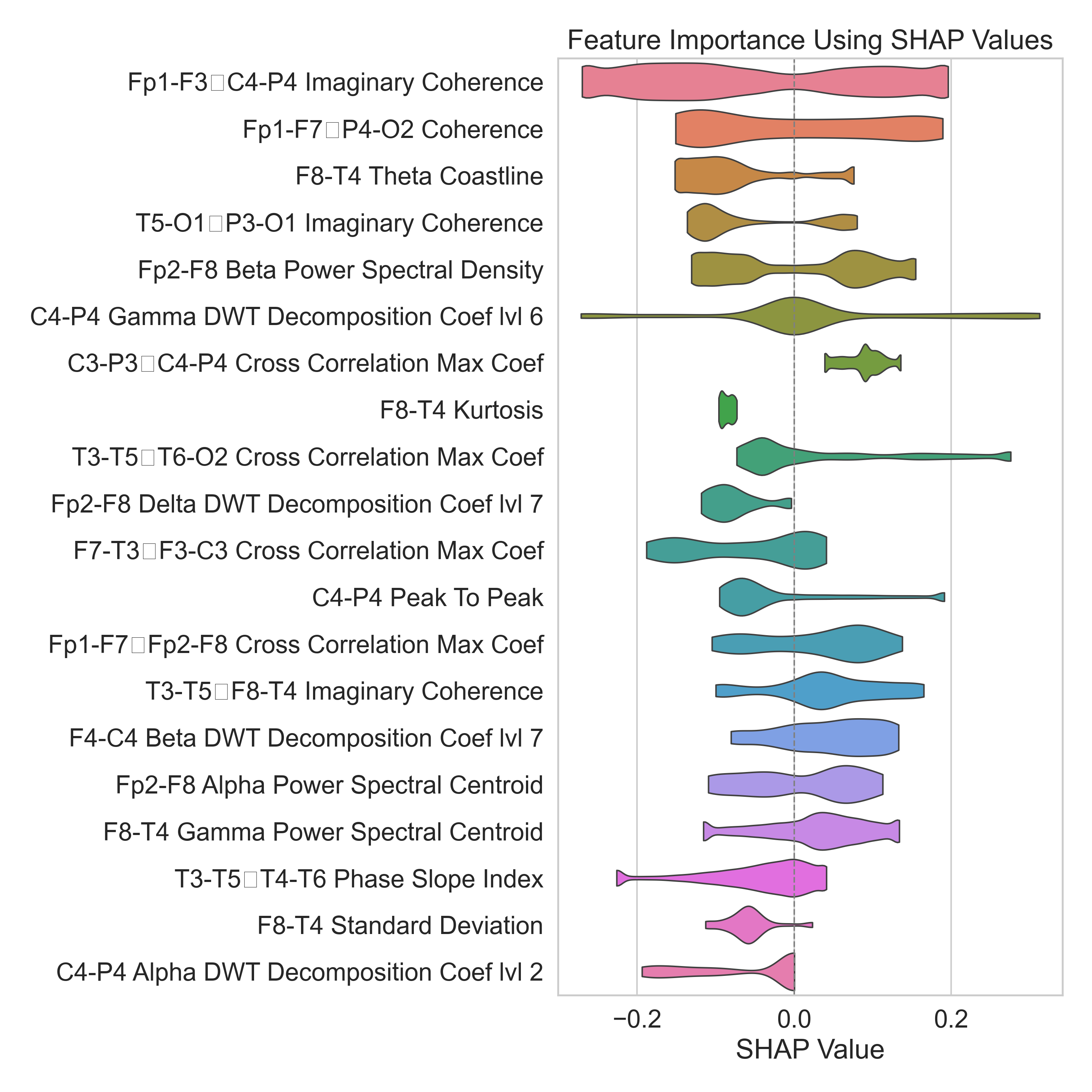}
    \caption{Global feature importance derived from \acrshort{shap} values, showing the top twenty most influential features for the \acrshort{xgb} model trained on \Acrfull{tusz} dataset and evaluated on \Acrshort{tusz} using engineered features.}
    \label{fig:feature-shap_XGB_tusz_tusz}
\end{figure}
\begin{figure}[!h]
    \centering
    \includegraphics[angle=0,origin=c,width=110mm]{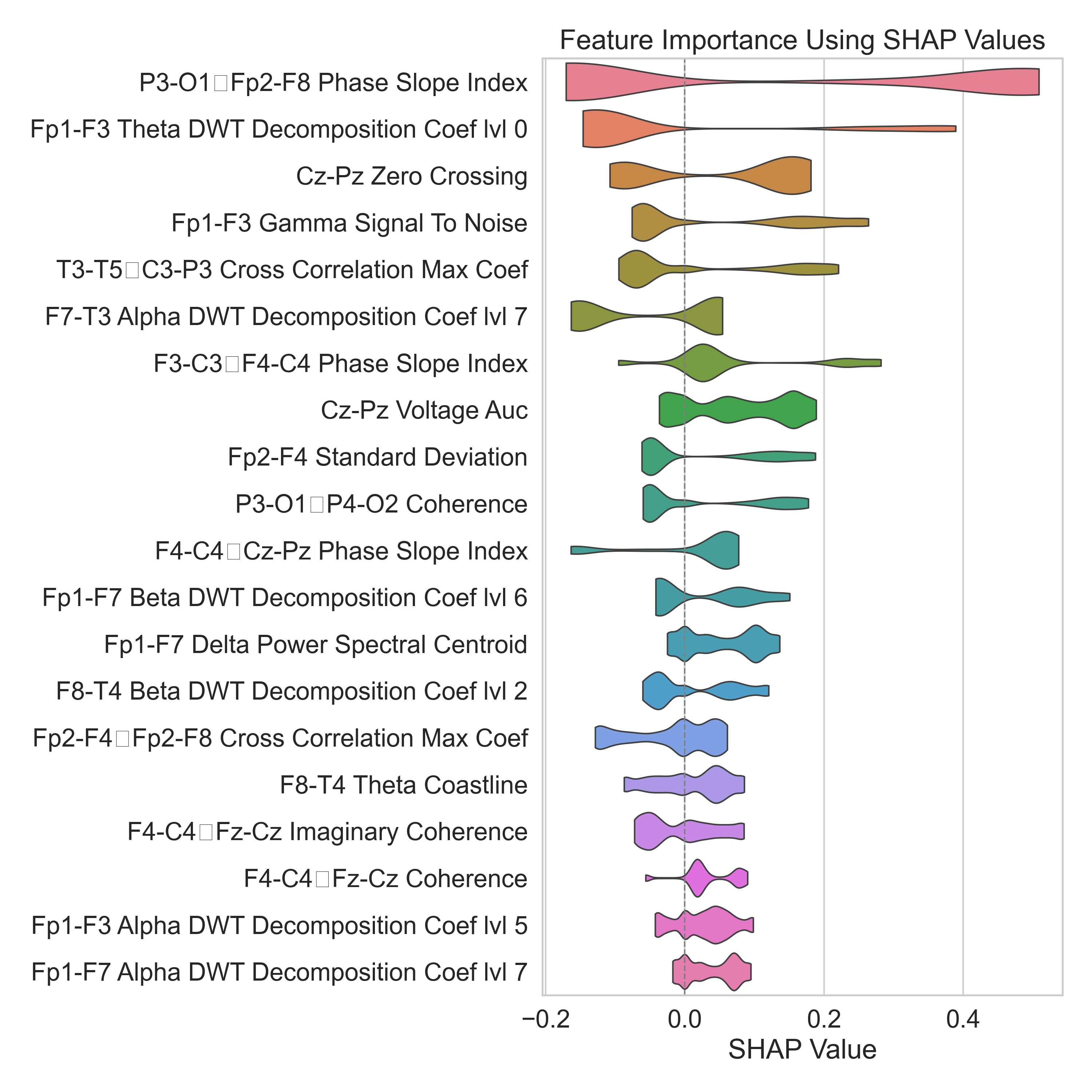}
    \caption{Global feature importance derived from \acrshort{shap} values, showing the top twenty most influential features for the \acrshort{xgb} model trained on \Acrfull{chb-mit} dataset and evaluated on \Acrshort{chb-mit} using engineered features.}
    \label{fig:feature-shap_XGB_chbmit_chbmit}
\end{figure}
\begin{figure}[!h]
    \centering
    \includegraphics[angle=0,origin=c,width=110mm]{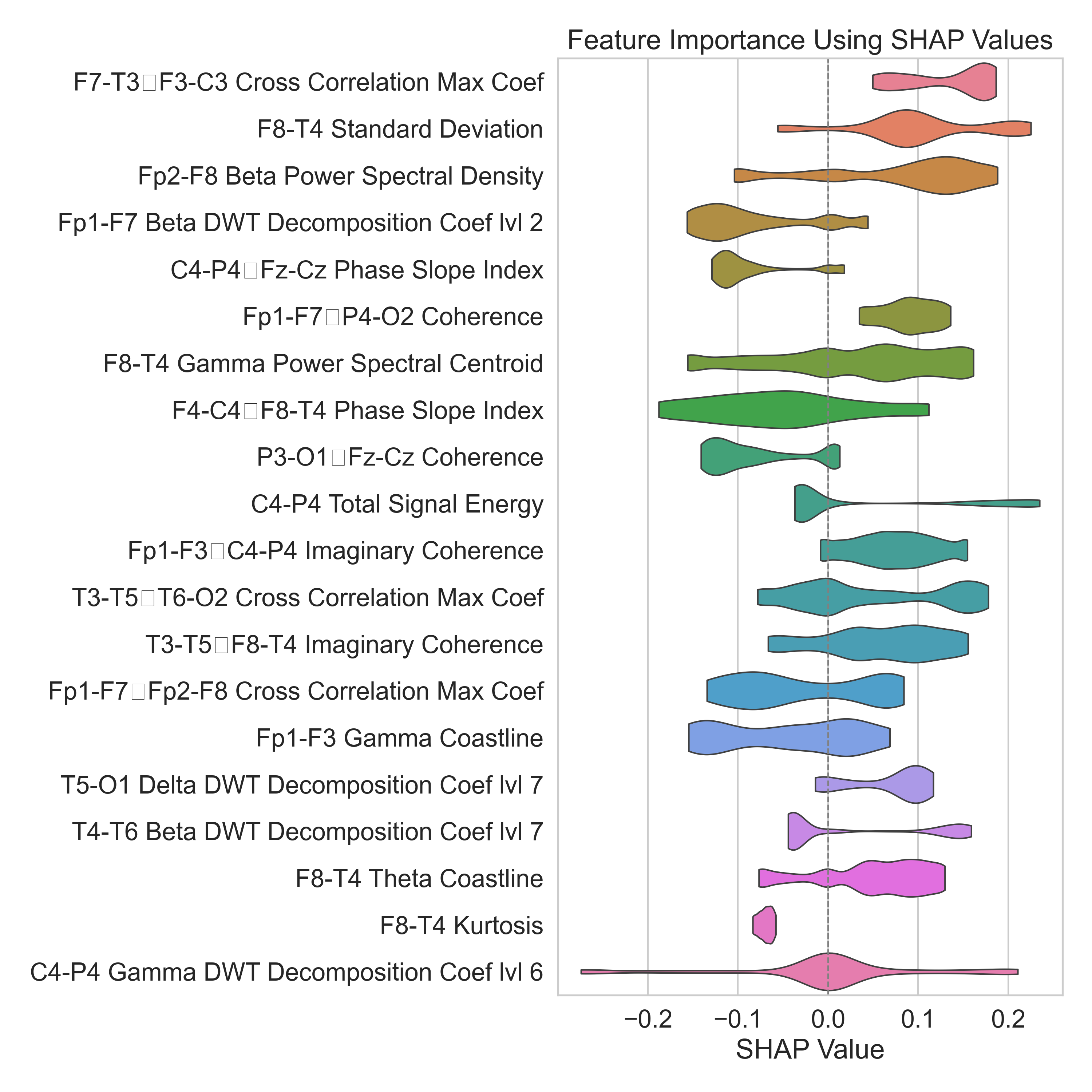}
    \caption{Global feature importance derived from \acrshort{shap} values, showing the top twenty most influential features for the \acrshort{xgb} model trained on \Acrfull{tusz} dataset and evaluated on \Acrfull{chb-mit} using engineered features.}
    \label{fig:feature-shap_XGB_tusz_chmit}
\end{figure}
\begin{figure}[!h]
    \centering
    \includegraphics[angle=0,origin=c,width=110mm]{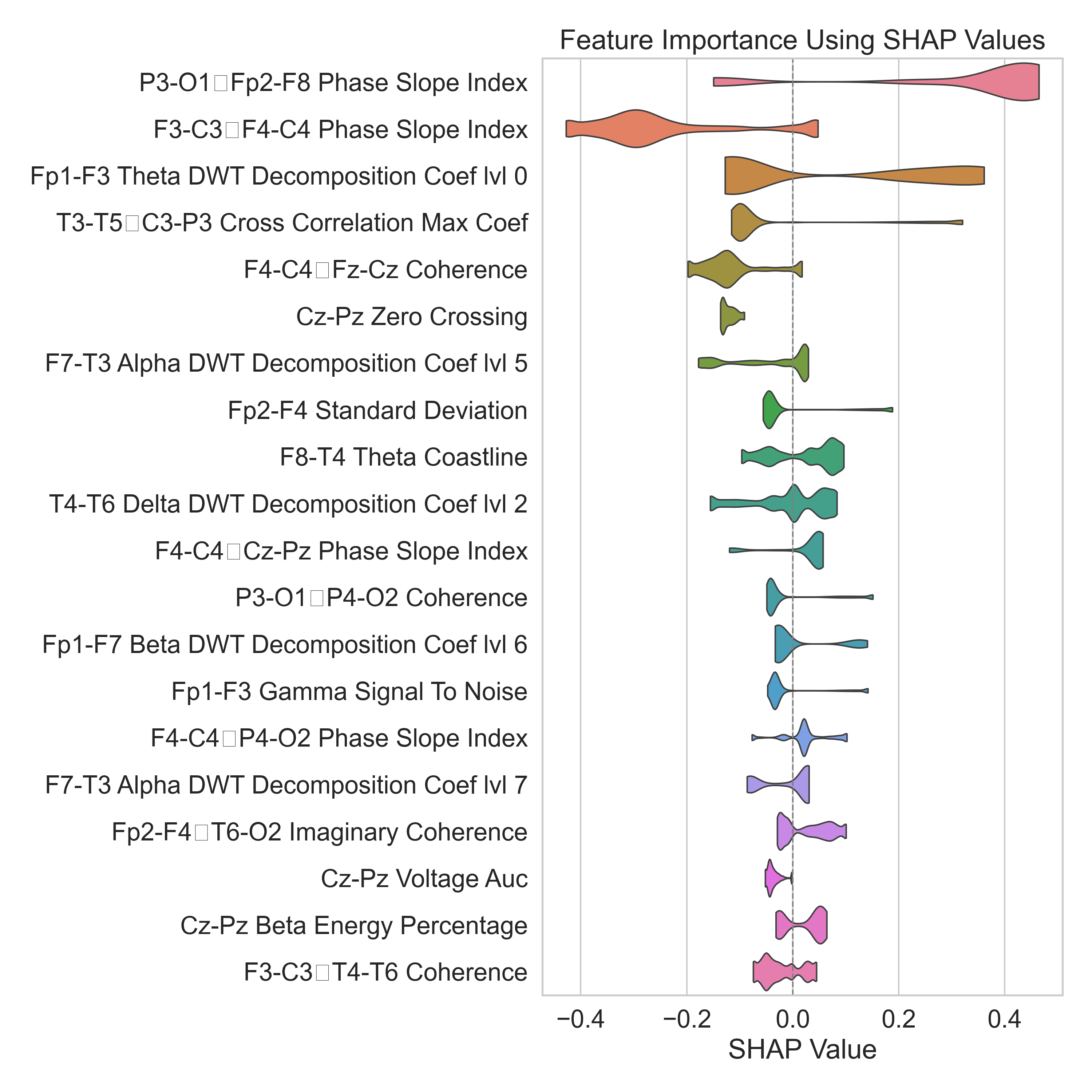}
    \caption{Global feature importance derived from \acrshort{shap} values, showing the top twenty most influential features for the \acrshort{xgb} model trained on \Acrfull{chb-mit} dataset and evaluated on \Acrfull{tusz} using engineered features.}
    \label{fig:feature-shap_XGB_chbmit_tusz}
\end{figure}

\begin{figure}[!h]
    \centering
    \includegraphics[angle=0,origin=c,width=110mm]{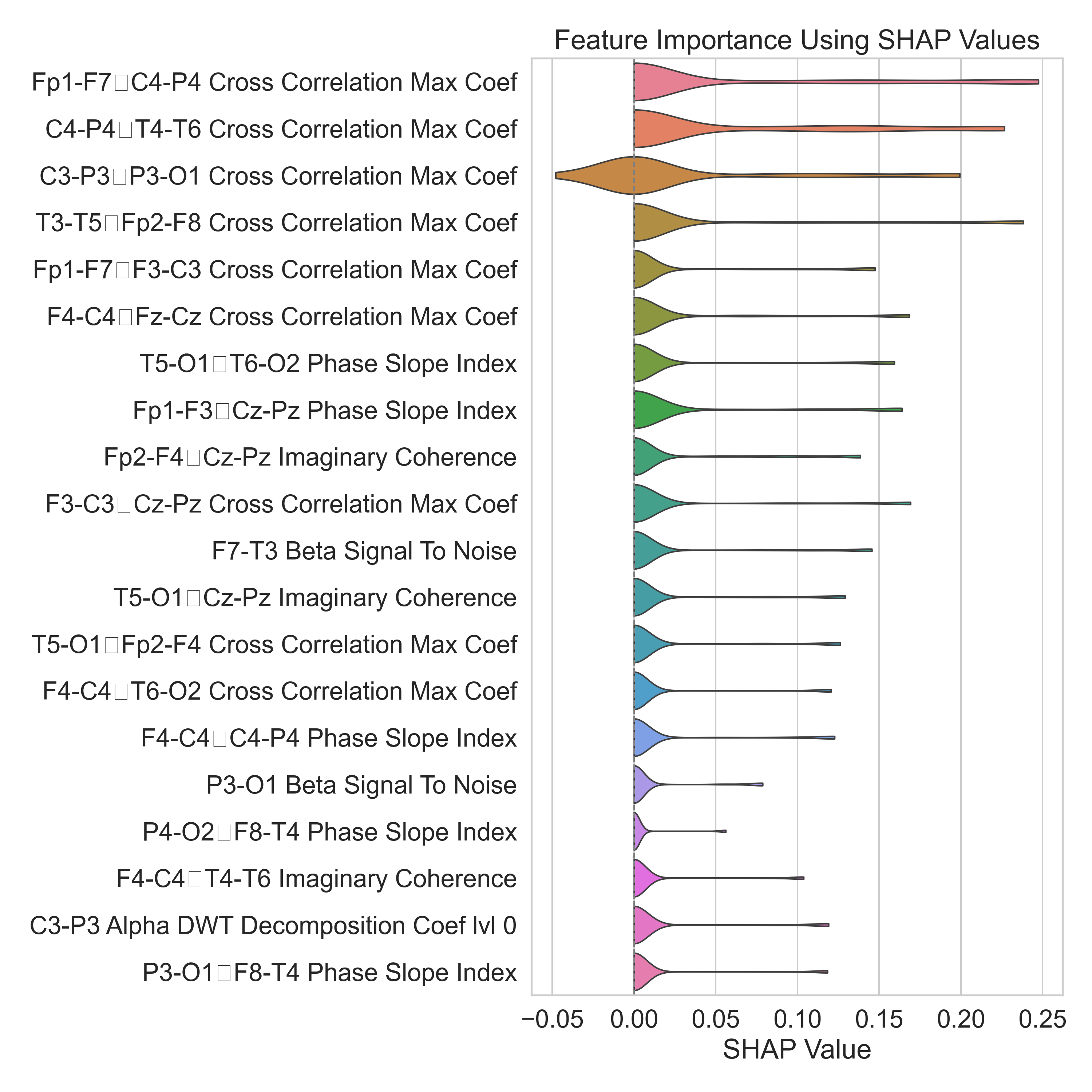}
    \caption{Global feature importance derived from \acrshort{shap} values, showing the top twenty most influential features for the \acrshort{mlp} model trained on \Acrfull{tusz} dataset and evaluated on \Acrshort{tusz} using engineered features.}
    \label{fig:feature-shap_MLP_tusz_tusz}
\end{figure}
\begin{figure}[!h]
    \centering
    \includegraphics[angle=0,origin=c,width=110mm]{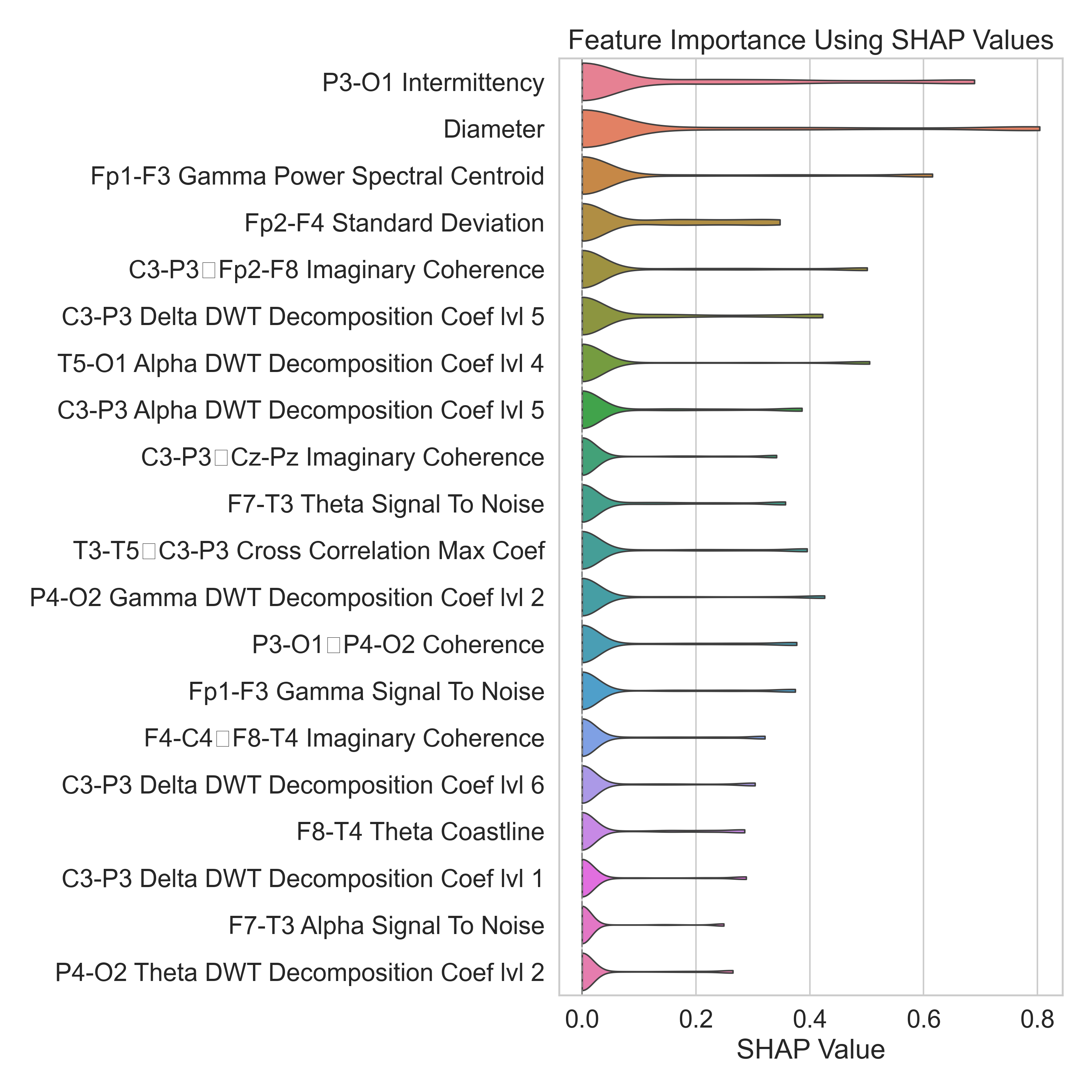}
    \caption{Global feature importance derived from \acrshort{shap} values, showing the top twenty most influential features for the \acrshort{mlp} model trained on \Acrfull{chb-mit} dataset and evaluated on \Acrshort{chb-mit} using engineered features.}
    \label{fig:feature-shap_MLP_chbmit_chbmit}
\end{figure}
\begin{figure}[!h]
    \centering
    \includegraphics[angle=0,origin=c,width=110mm]{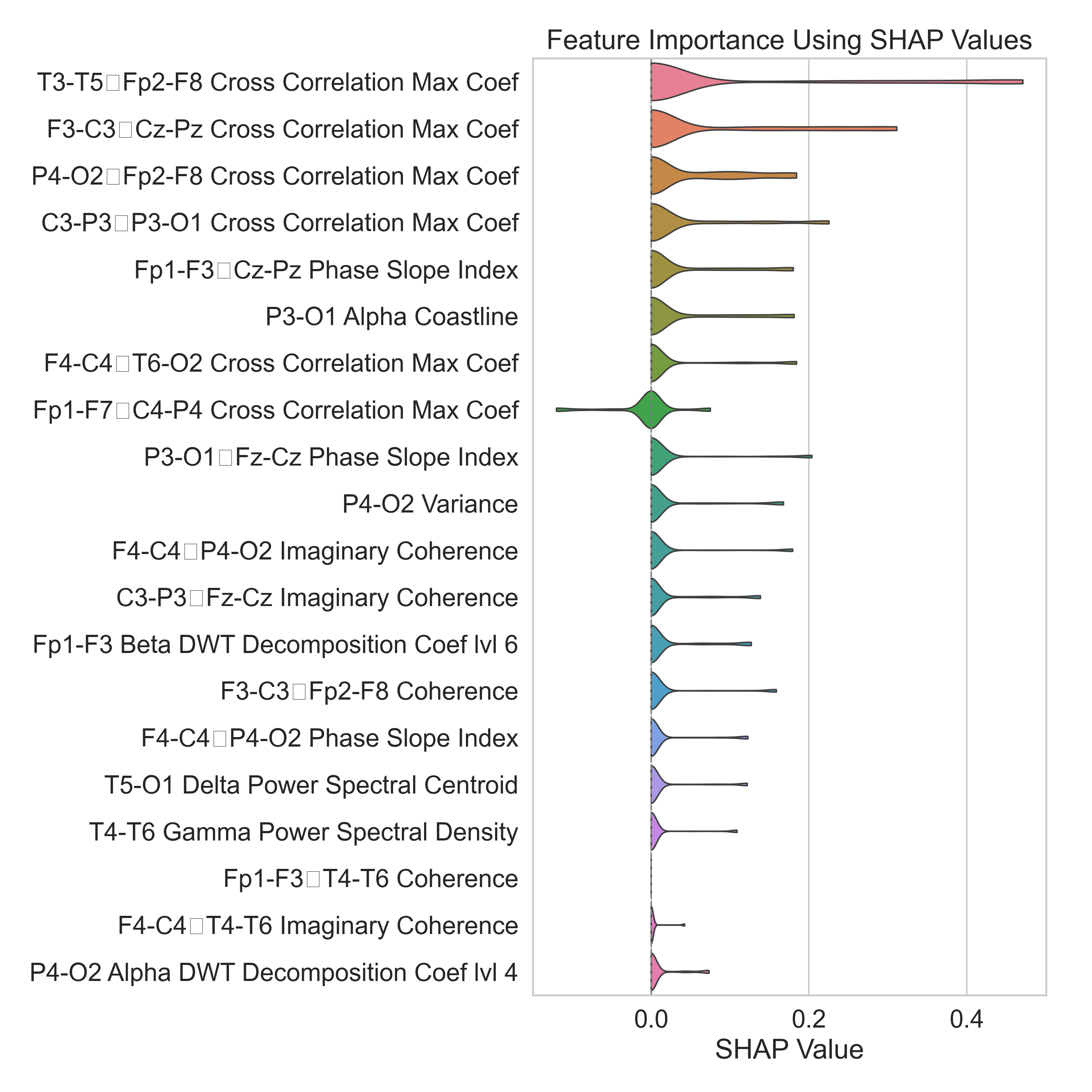}
    \caption{Global feature importance derived from \acrshort{shap} values, showing the top twenty most influential features for the \acrshort{mlp} model trained on \Acrfull{tusz} dataset and evaluated on \Acrfull{chb-mit} using engineered features.}
    \label{fig:feature-shap_MLP_tusz_chmit}
\end{figure}
\begin{figure}[!h]
    \centering
    \includegraphics[angle=0,origin=c,width=110mm]{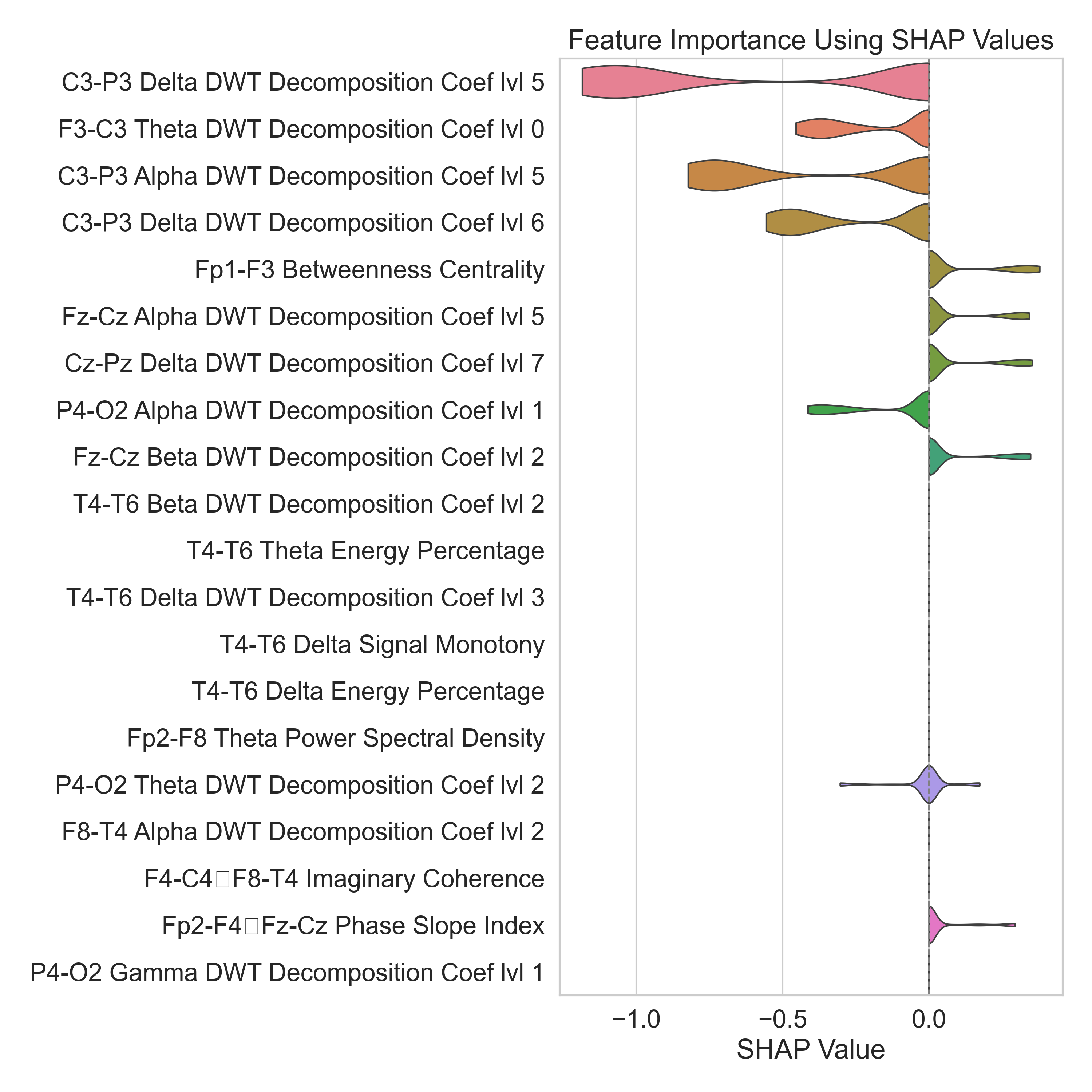}
    \caption{Global feature importance derived from \acrshort{shap} values, showing the top twenty most influential features for the \acrshort{mlp} model trained on \Acrfull{chb-mit} dataset and evaluated on \Acrfull{tusz} using engineered features.}
    \label{fig:feature-shap_MLP_chbmit_tusz}
\end{figure}

\begin{figure}[!h]
    \centering
    \includegraphics[angle=0,origin=c,width=110mm]{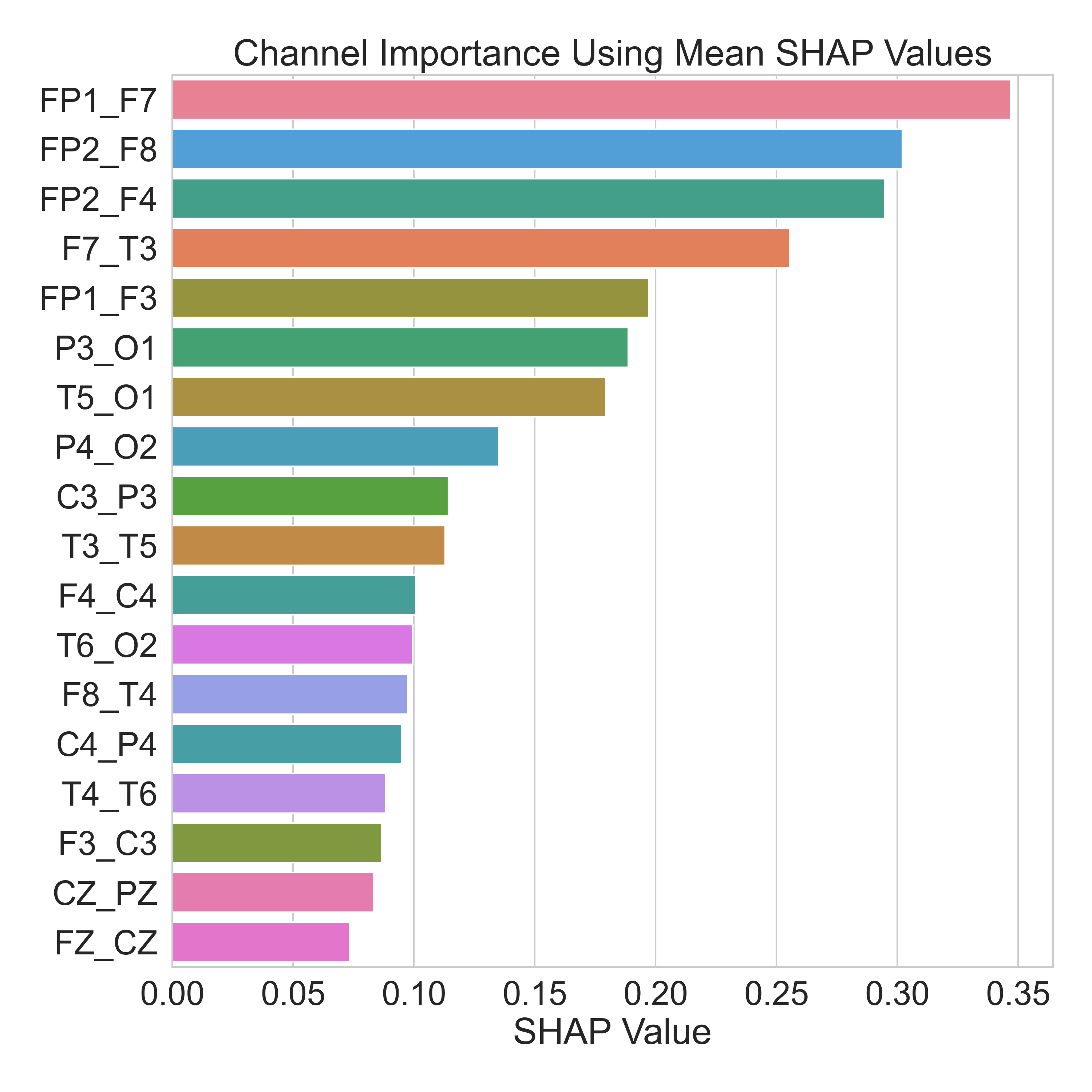}
    \caption{Channel importance using \acrshort{shap} values for the \acrshort{cnn} model trained on \Acrfull{tusz} dataset and evaluated on \Acrshort{tusz} using unprocessed data.}
    \label{fig:feature-shap_CNN_tusz_tusz}
\end{figure}
\begin{figure}[!h]
    \centering
    \includegraphics[angle=0,origin=c,width=110mm]{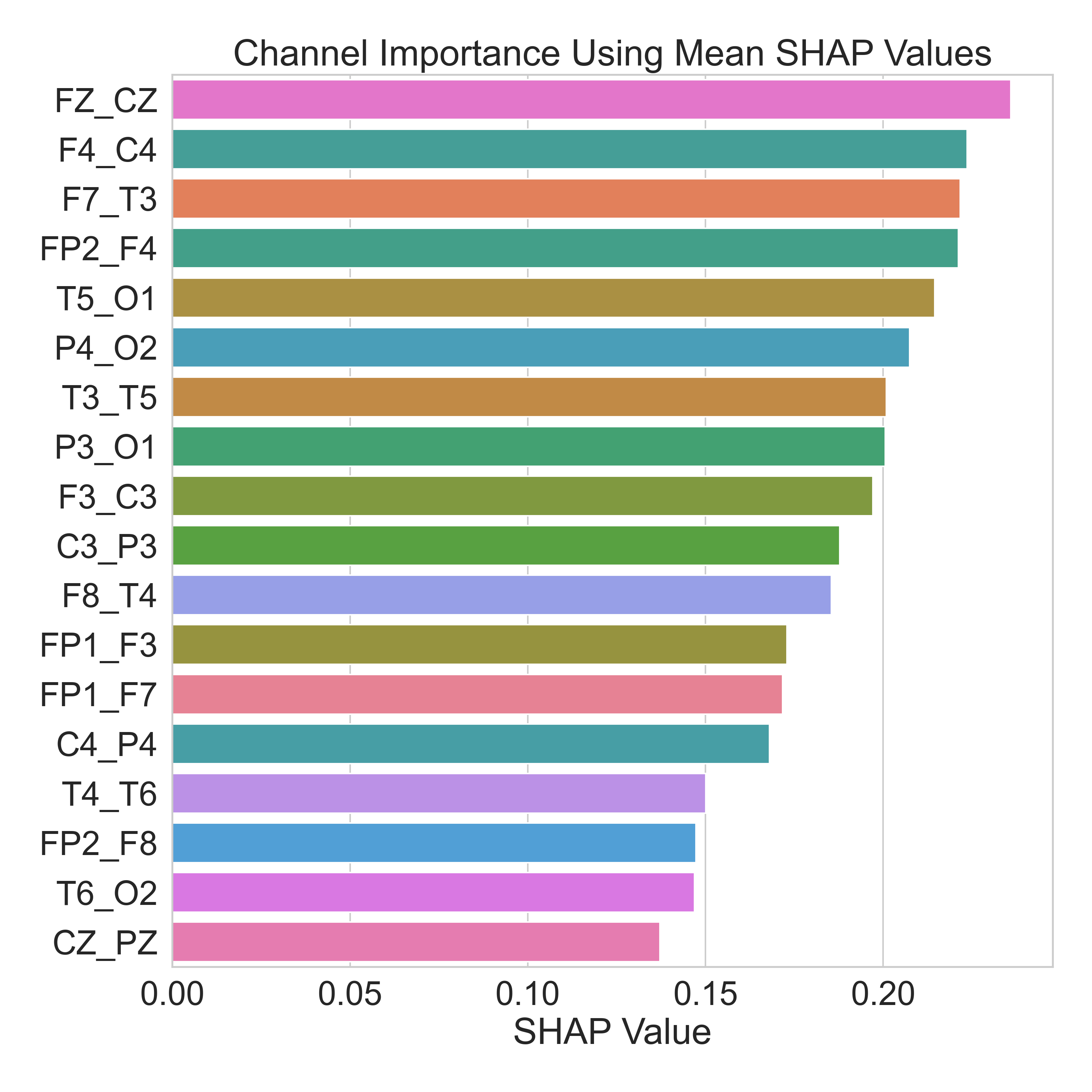}
    \caption{Channel importance using \acrshort{shap} values for the \acrshort{cnn} model trained on \Acrfull{chb-mit} dataset and evaluated on \Acrshort{chb-mit} using unprocessed data.}
    \label{fig:feature-shap_CNN_chbmit_chbmit}
\end{figure}
\begin{figure}[!h]
    \centering
    \includegraphics[angle=0,origin=c,width=110mm]{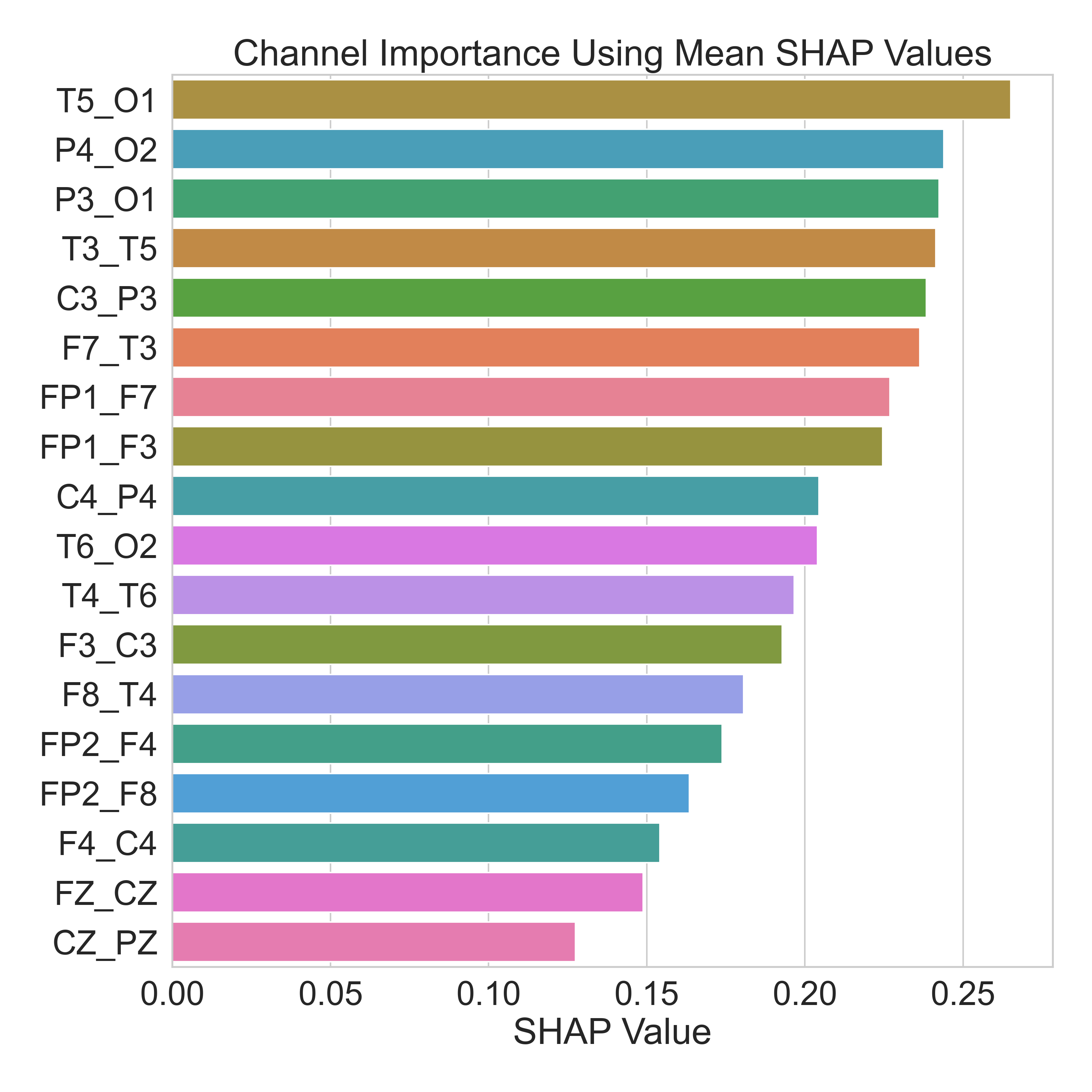}
    \caption{Channel importance using \acrshort{shap} values for the \acrshort{cnn} model trained on \Acrfull{tusz} dataset and evaluated on \Acrfull{chb-mit} using unprocessed data.}
    \label{fig:feature-shap_CNN_tusz_chmit}
\end{figure}
\begin{figure}[!h]
    \centering
    \includegraphics[angle=0,origin=c,width=110mm]{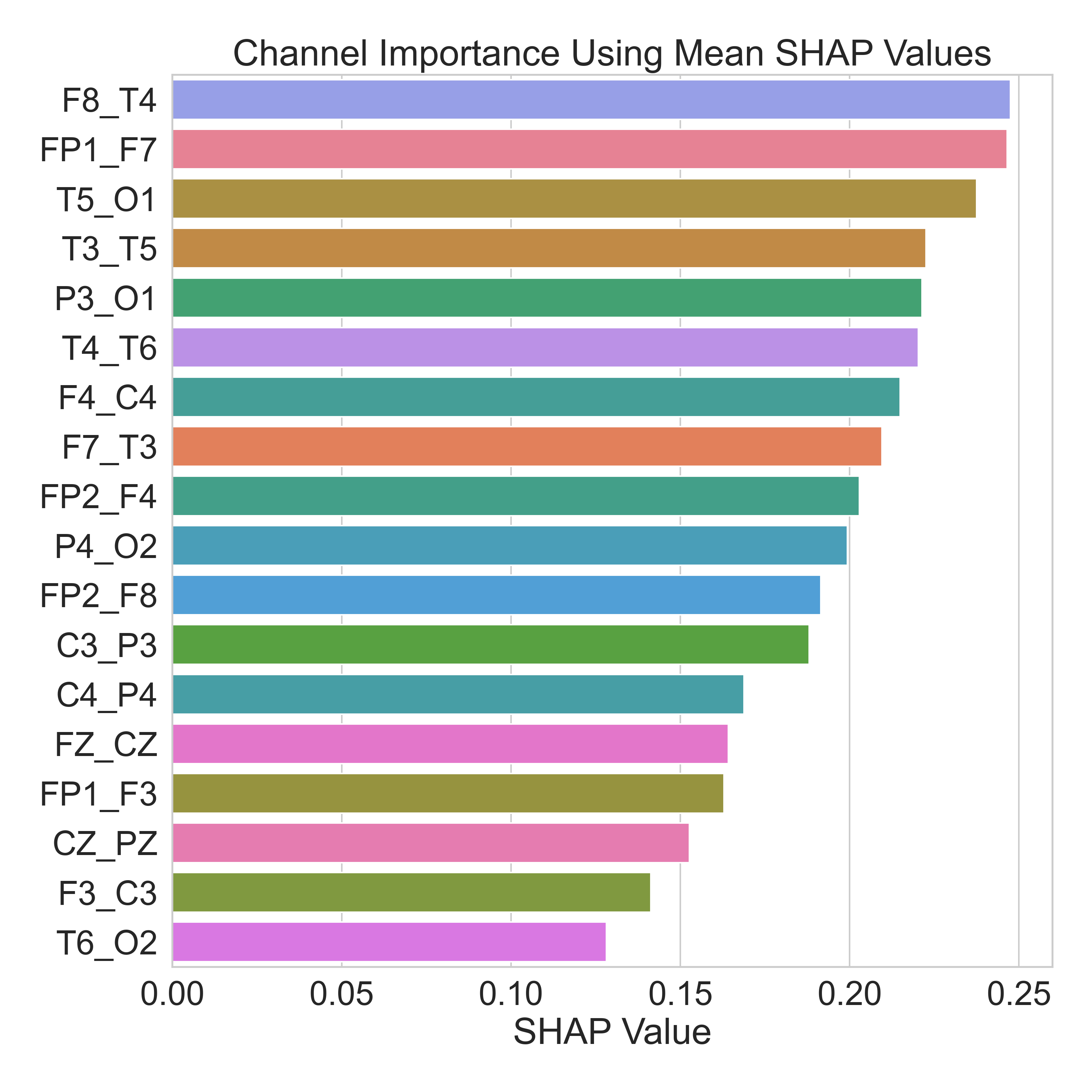}
    \caption{Channel importance using \acrshort{shap} values for the \acrshort{cnn} model trained on \Acrfull{chb-mit} dataset and evaluated on \Acrfull{tusz} using unprocessed data.}
    \label{fig:feature-shap_CNN_chbmit_tusz}
\end{figure}

\begin{figure}[!h]
    \centering
    \includegraphics[angle=0,origin=c,width=110mm]{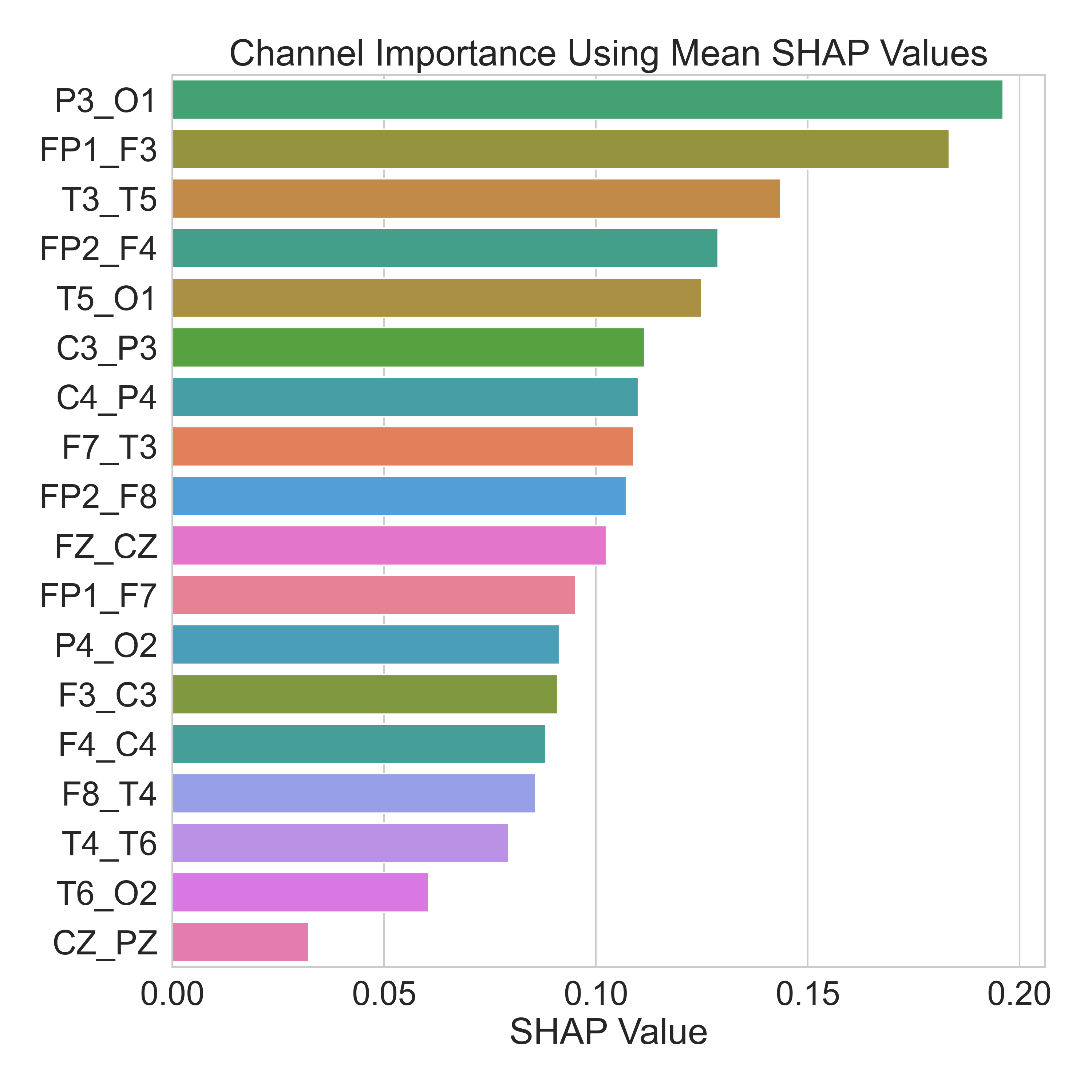}
    \caption{Channel importance using \acrshort{shap} values for the \acrshort{eegnet} model trained on \Acrfull{tusz} dataset and evaluated on \Acrshort{tusz} using unprocessed data.}
    \label{fig:feature-shap_EEGNet_tusz_tusz}
\end{figure}
\begin{figure}[!h]
    \centering
    \includegraphics[angle=0,origin=c,width=110mm]{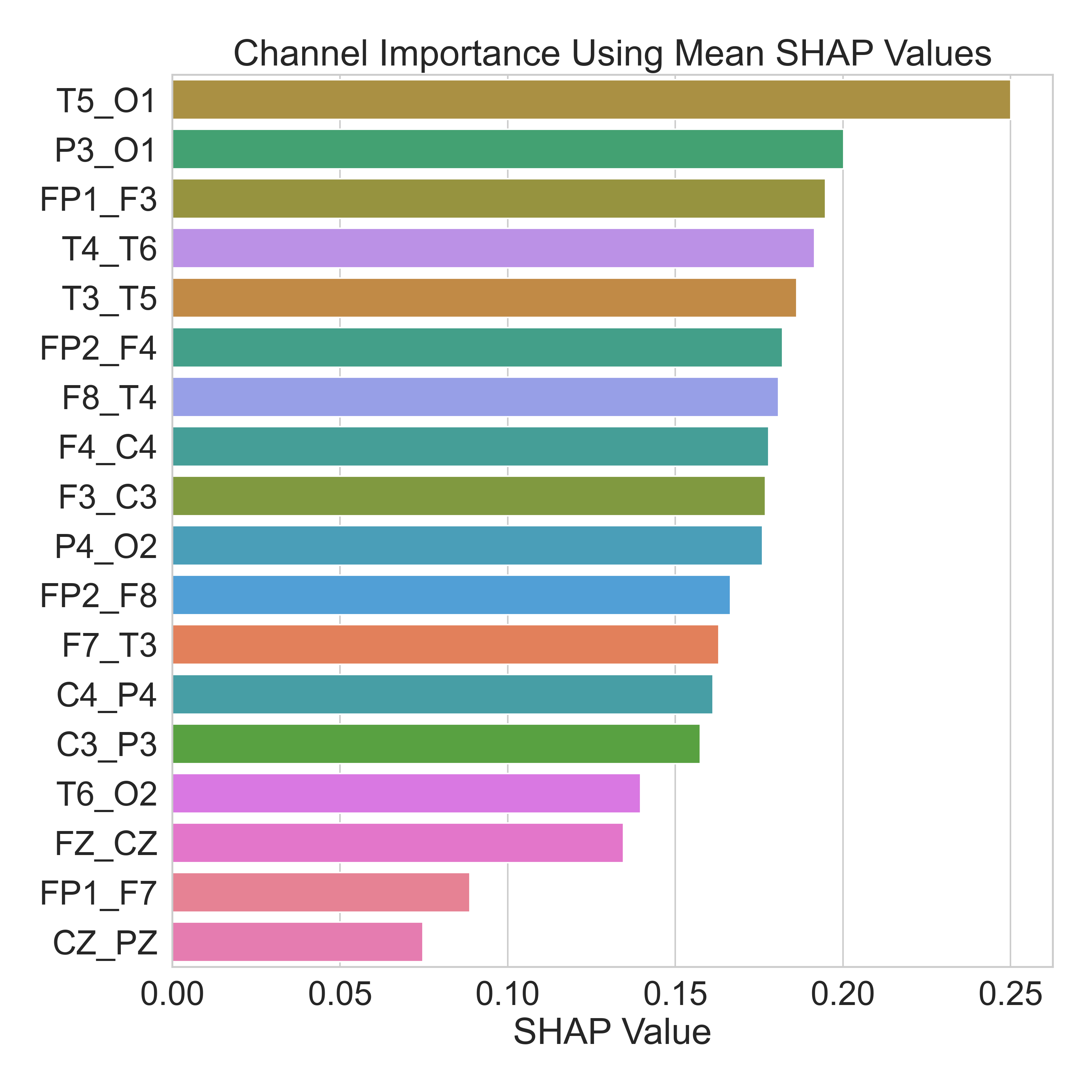}
    \caption{Channel importance using \acrshort{shap} values for the \acrshort{eegnet} model trained on \Acrfull{chb-mit} dataset and evaluated on \Acrshort{chb-mit} using unprocessed data.}
    \label{fig:feature-shap_EEGNet_chbmit_chbmit}
\end{figure}
\begin{figure}[!h]
    \centering
    \includegraphics[angle=0,origin=c,width=110mm]{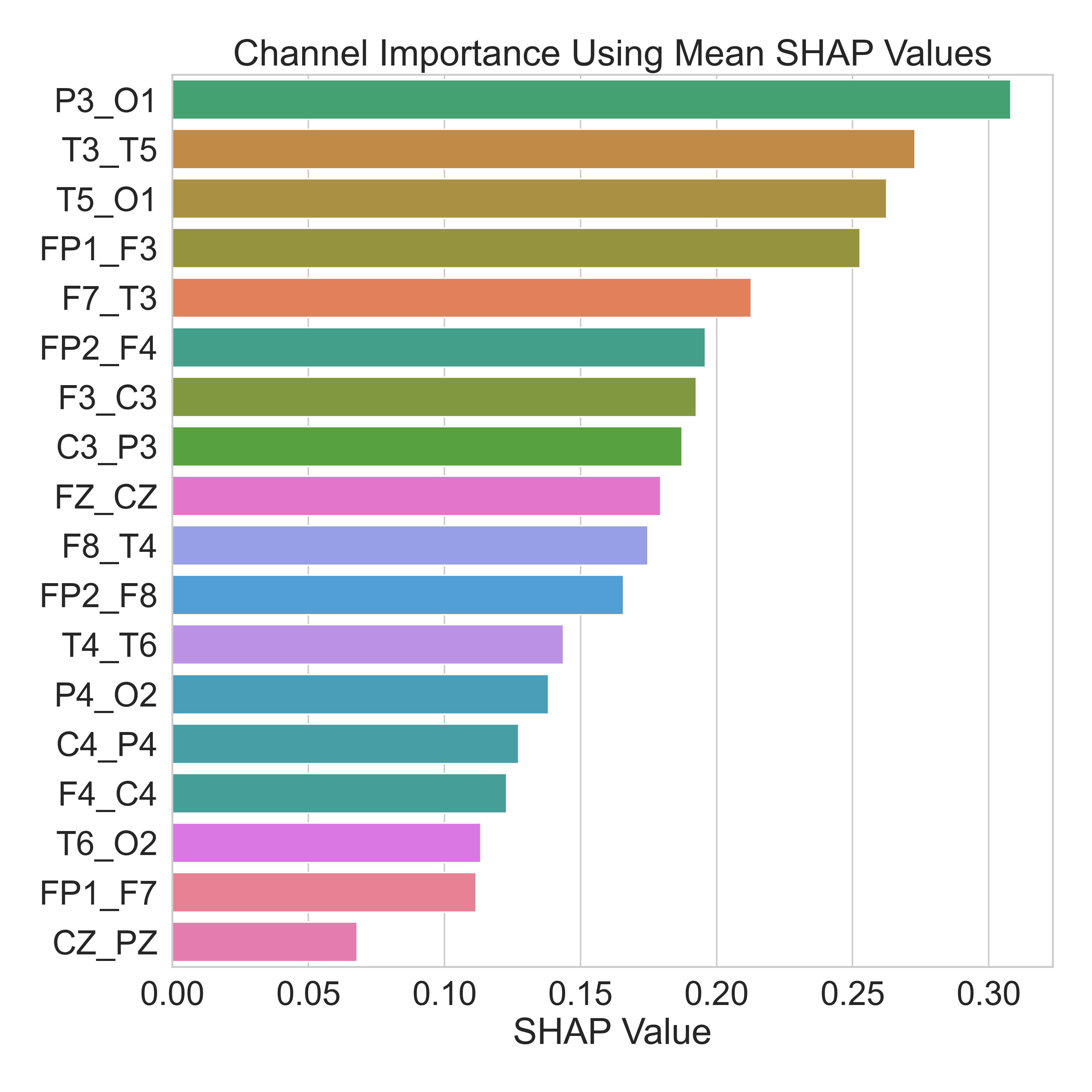}
    \caption{Channel importance using \acrshort{shap} values for the \acrshort{eegnet} model trained on \Acrfull{tusz} dataset and evaluated on \Acrfull{chb-mit} using unprocessed data.}
    \label{fig:feature-shap_EEGNet_tusz_chmit}
\end{figure}
\begin{figure}[!h]
    \centering
    \includegraphics[angle=0,origin=c,width=110mm]{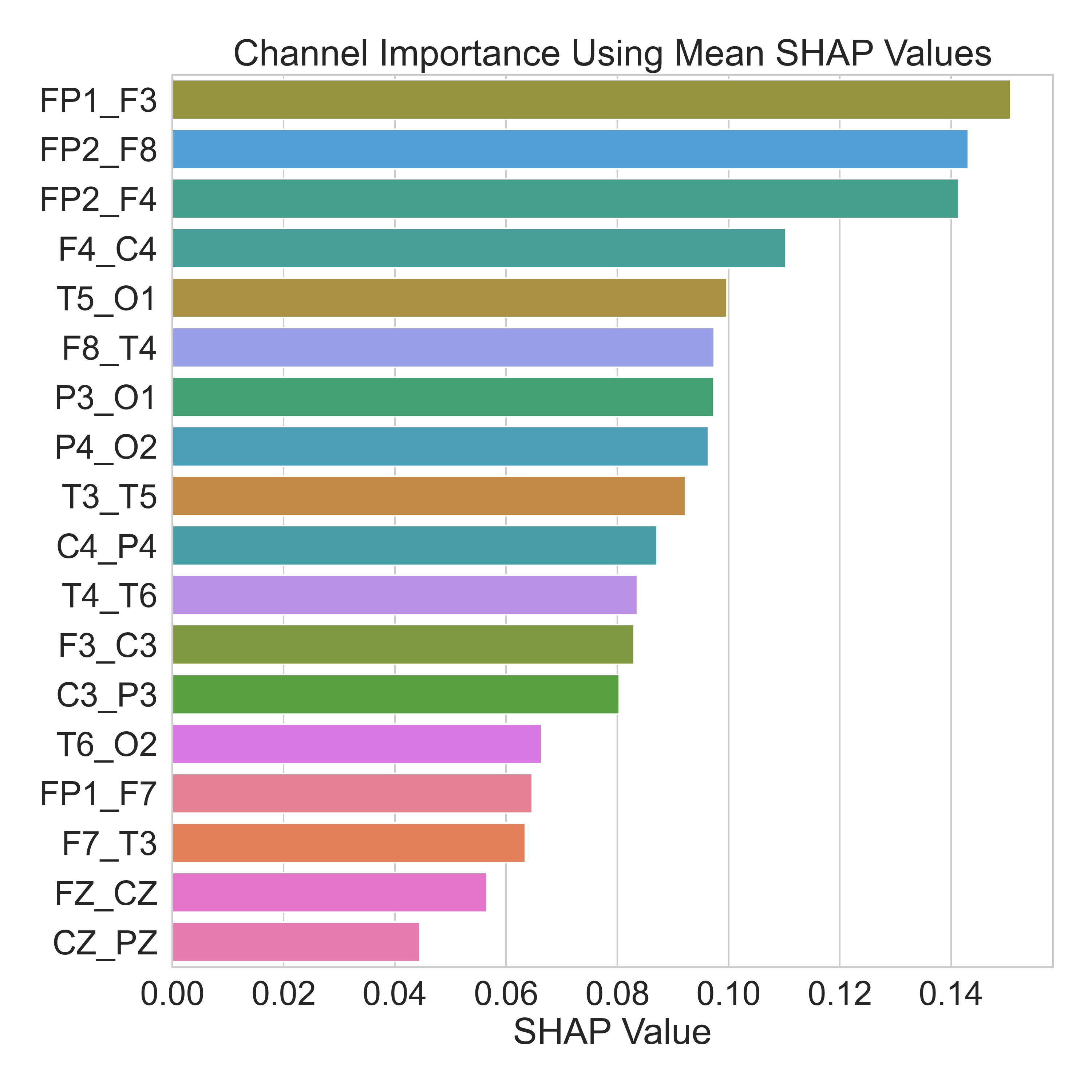}
    \caption{Channel importance using \acrshort{shap} values for the \acrshort{eegnet} model trained on \Acrfull{chb-mit} dataset and evaluated on \Acrfull{tusz} using unprocessed data.}
    \label{fig:feature-shap_EEGNet_chbmit_tusz}
\end{figure}

\begin{figure}[!h]
    \centering
    \includegraphics[angle=0,origin=c,width=110mm]{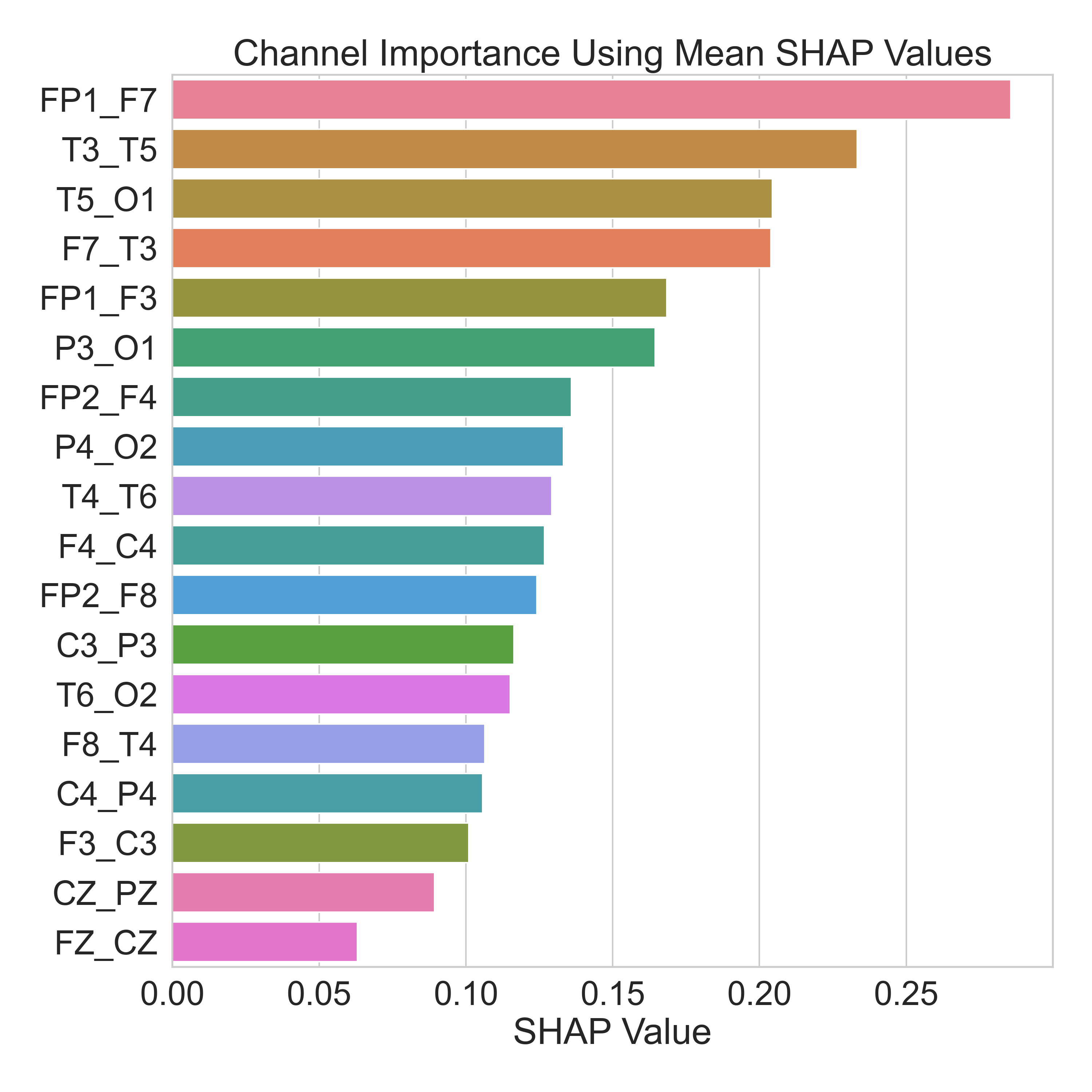}
    \caption{Channel importance using \acrshort{shap} values for the \acrshort{convlstm} model trained on \Acrfull{tusz} dataset and evaluated on \Acrshort{tusz} using unprocessed data.}
    \label{fig:feature-shap_ConvLSTM_tusz_tusz}
\end{figure}
\begin{figure}[!h]
    \centering
    \includegraphics[angle=0,origin=c,width=110mm]{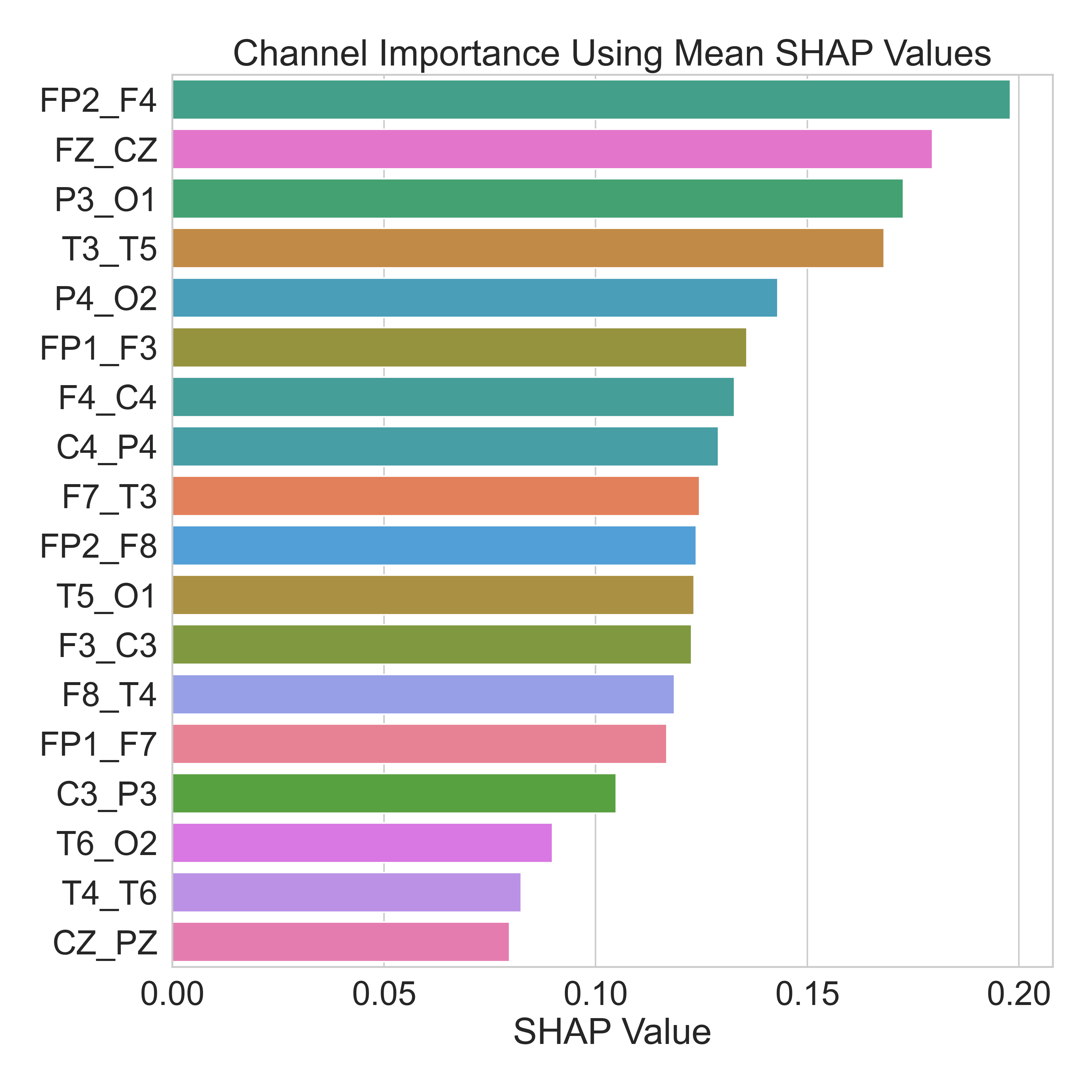}
    \caption{Channel importance using \acrshort{shap} values for the \acrshort{convlstm} model trained on \Acrfull{chb-mit} dataset and evaluated on \Acrshort{chb-mit} using unprocessed data.}
    \label{fig:feature-shap_ConvLSTM_chbmit_chbmit}
\end{figure}
\begin{figure}[!h]
    \centering
    \includegraphics[angle=0,origin=c,width=110mm]{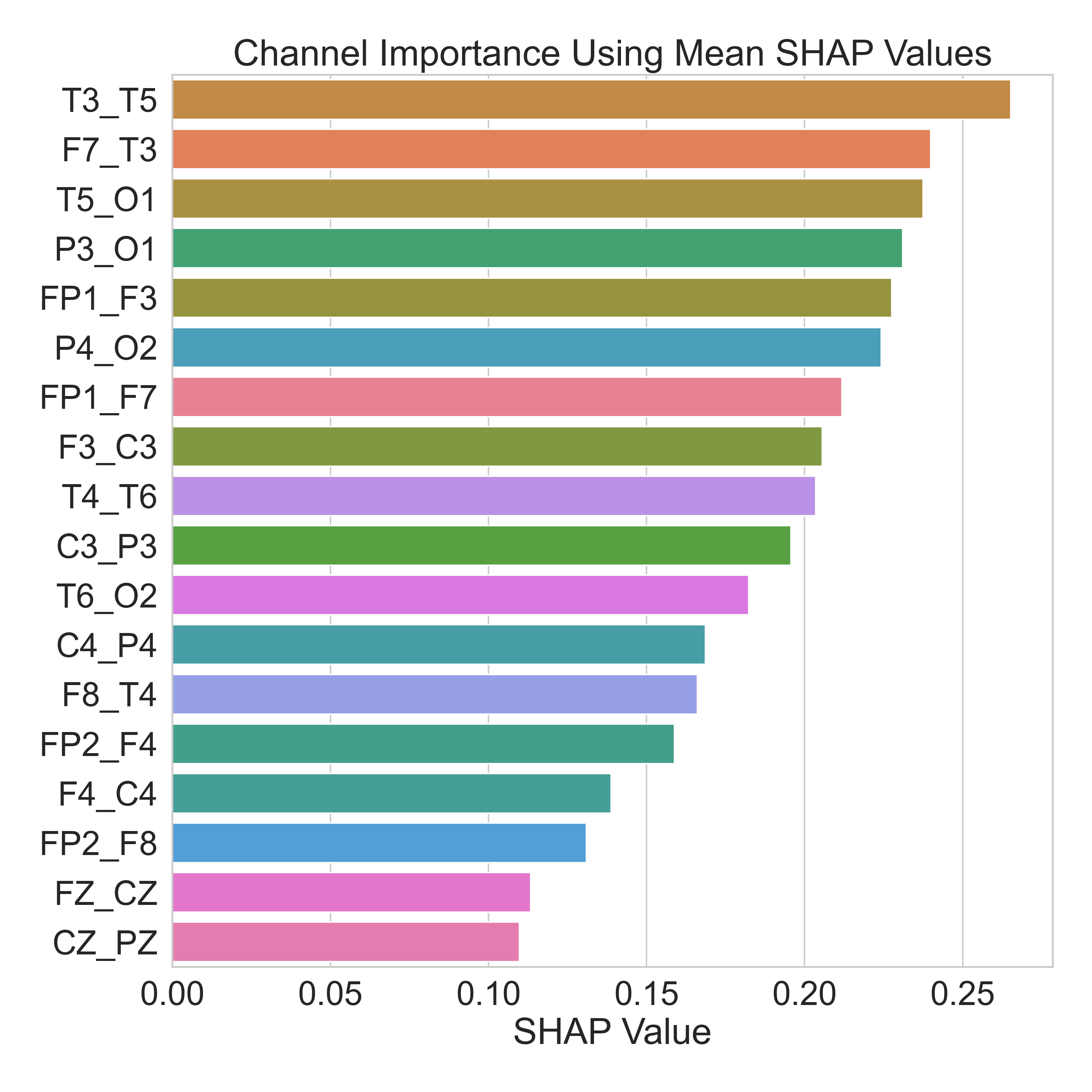}
    \caption{Channel importance using \acrshort{shap} values for the \acrshort{convlstm} model trained on \Acrfull{tusz} dataset and evaluated on \Acrfull{chb-mit} using unprocessed data.}
    \label{fig:feature-shap_ConvLSTM_tusz_chmit}
\end{figure}
\begin{figure}[!h]
    \centering
    \includegraphics[angle=0,origin=c,width=110mm]{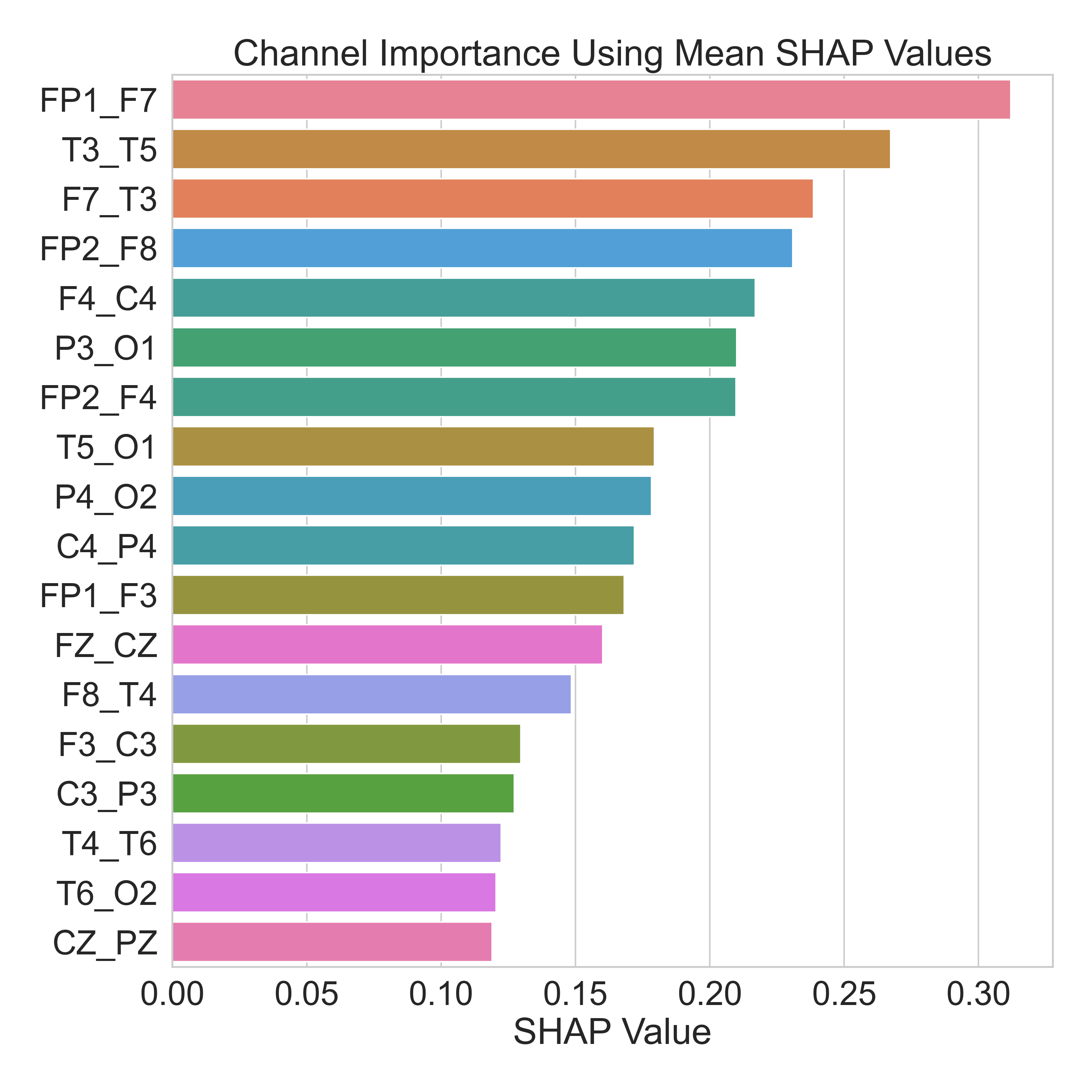}
    \caption{Channel importance using \acrshort{shap} values for the \acrshort{convlstm} model trained on \Acrfull{chb-mit} dataset and evaluated on \Acrfull{tusz} using unprocessed data.}
    \label{fig:feature-shap_ConvLSTM_chbmit_tusz}
\end{figure}

\begin{figure}[!h]
    \centering
    \includegraphics[angle=0,origin=c,width=110mm]{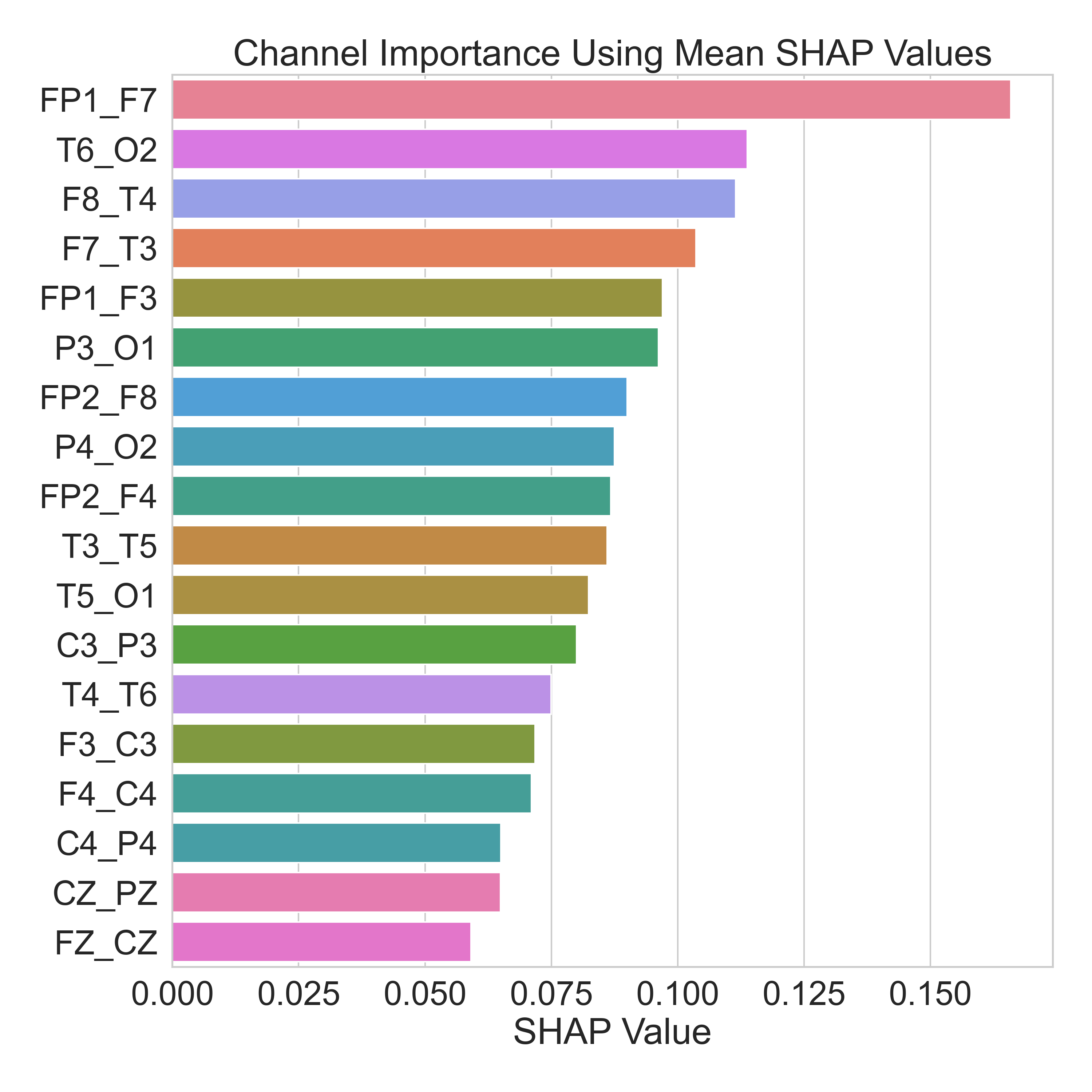}
    \caption{Channel importance using \acrshort{shap} values for the \acrshort{convtransformer} model trained on \Acrfull{tusz} dataset and evaluated on \Acrshort{tusz} using unprocessed data.}
    \label{fig:feature-shap_ConvTransformer_tusz_tusz}
\end{figure}
\begin{figure}[!h]
    \centering
    \includegraphics[angle=0,origin=c,width=110mm]{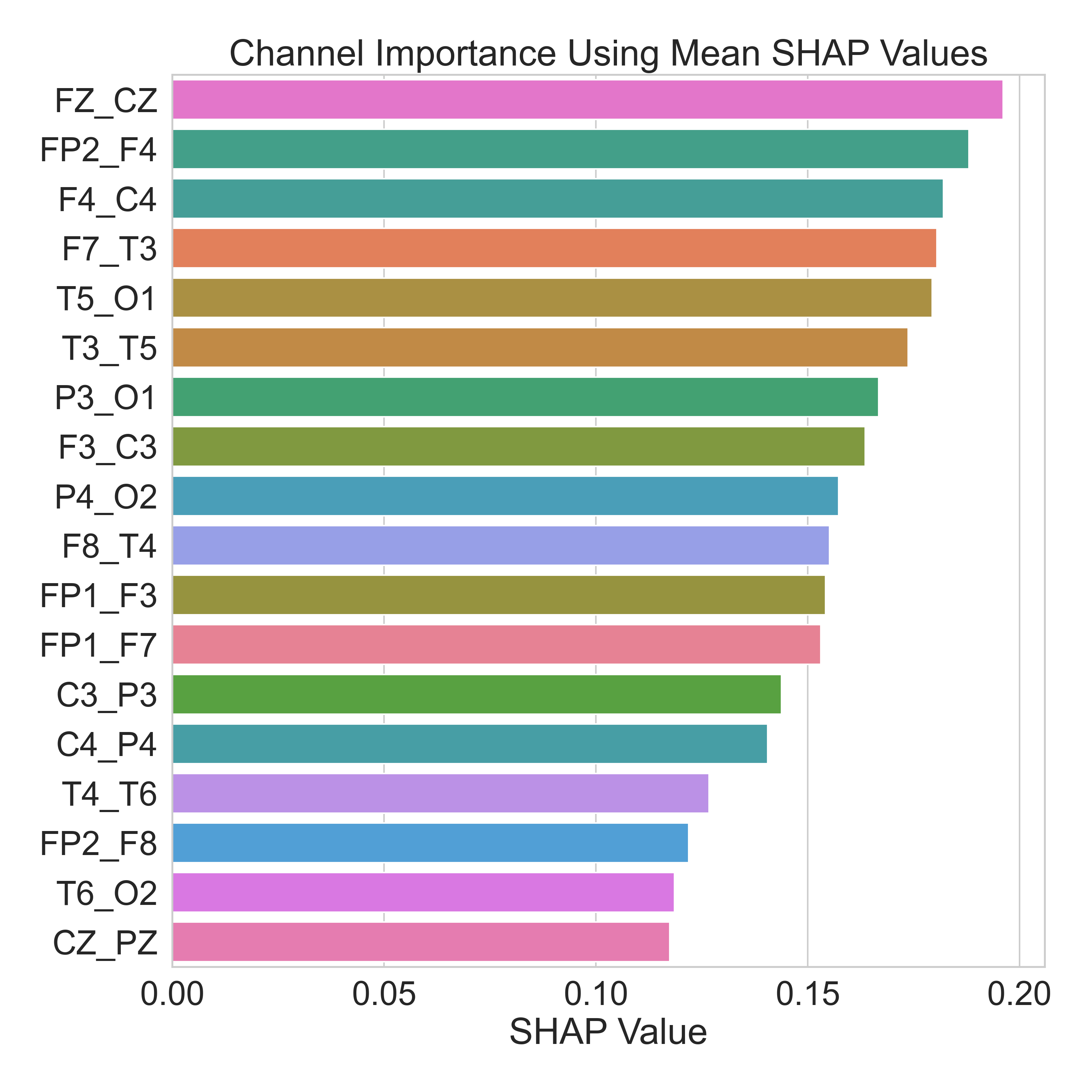}
    \caption{Channel importance using \acrshort{shap} values for the \acrshort{convtransformer} model trained on \Acrfull{chb-mit} dataset and evaluated on \Acrshort{chb-mit} using unprocessed data.}
    \label{fig:feature-shap_ConvTransformer_chbmit_chbmit}
\end{figure}
\begin{figure}[!h]
    \centering
    \includegraphics[angle=0,origin=c,width=110mm]{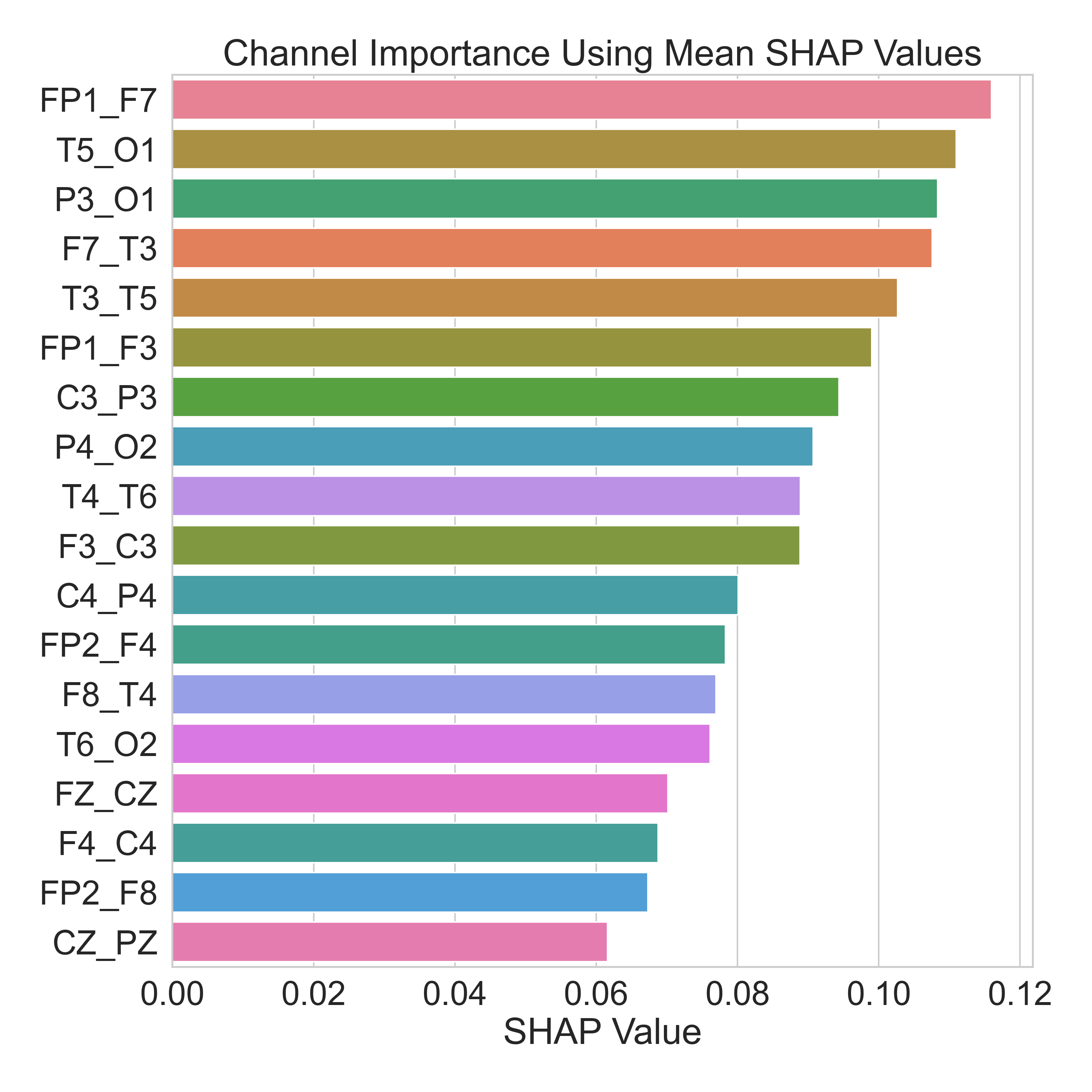}
    \caption{Channel importance using \acrshort{shap} values for the \acrshort{convtransformer} model trained on \Acrfull{tusz} dataset and evaluated on \Acrfull{chb-mit} using unprocessed data.}
    \label{fig:feature-shap_ConvTransformer_tusz_chmit}
\end{figure}
\begin{figure}[!h]
    \centering
    \includegraphics[angle=0,origin=c,width=110mm]{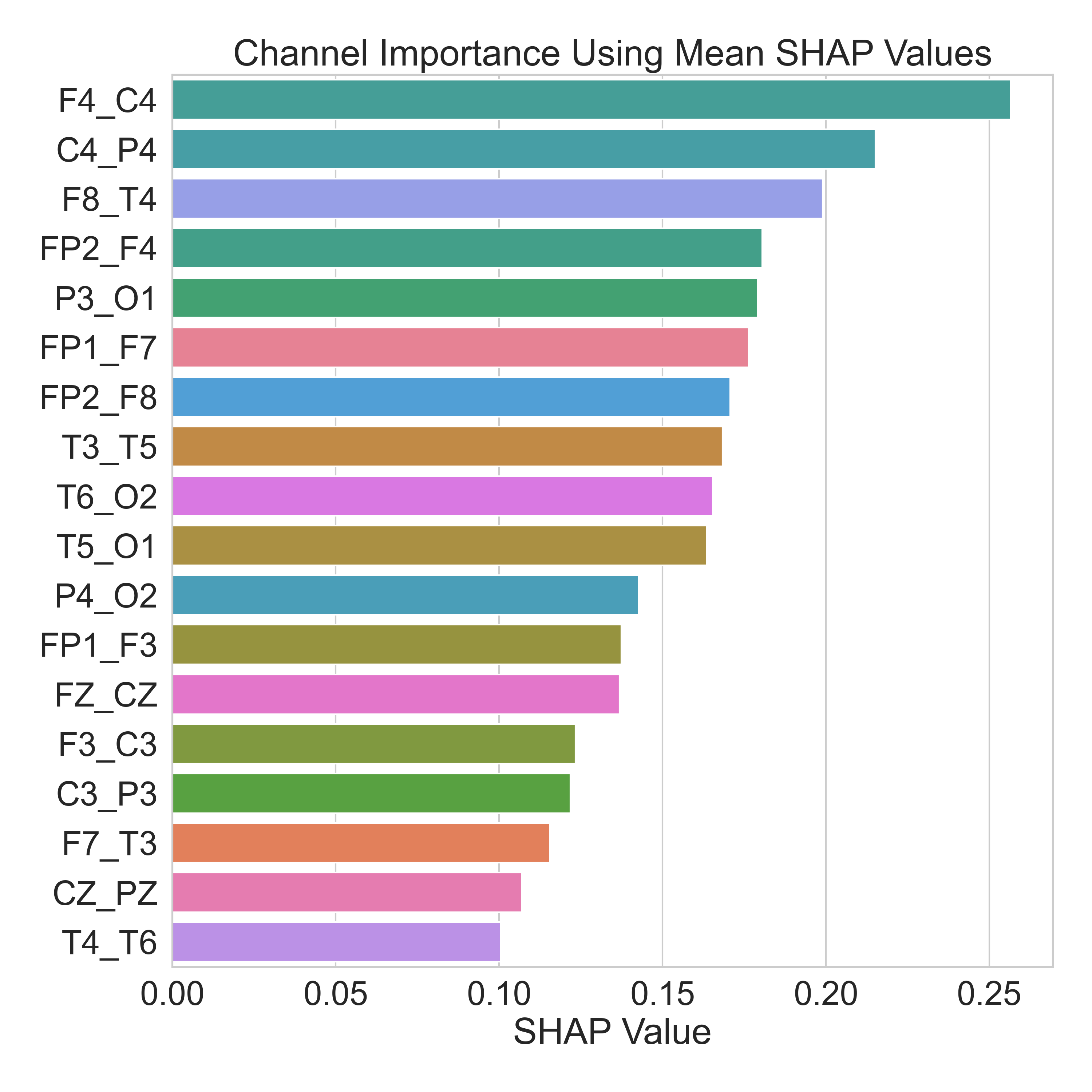}
    \caption{Channel importance using \acrshort{shap} values for the \acrshort{convtransformer} model trained on \Acrfull{chb-mit} dataset and evaluated on \Acrfull{tusz} using unprocessed data.}
    \label{fig:feature-shap_ConvTransformer_chbmit_tusz}
\end{figure}

\clearpage

\section{Tables}\label{sec:supplementary:tables}

\begin{table}[!h]
\caption{Overview of the hyperparameter search space used during model tuning for all evaluated classifiers.}
\label{tab:search-space}
\begin{tabular}{r|ll}
    \textbf{Model} & \textbf{Parameter} & \textbf{Search space} \\
    \hline\\
    \multirow{11}{*}{XGB} & learning rate & 0.001, 0.01, 0.05, 0.1, 0.2, 0.3 \\
     & batch size & 32 - 256 (linear spacing, step: 32) \\
     & max depth & 0.7 - 1.6 (log-scale, 10 steps) \\
     & min child weight & 0.5 - 4 (log-scale, 10 steps) \\
     & reg alpha & 0.001 - 10 (geometric spacing, 10 steps) \\
     & reg lambda & 0.001 - 10 (geometric spacing, 10 steps) \\
     & reg gamma & 0.000001 - 0.2 (geometric spacing, 10 steps) \\
     & max delta step & 0.1 - 10 (geometric spacing, 10 steps) \\
     & colsample bytree & 0.3 - 10 (geometric spacing, 10 steps) \\
     & num parallel tree & 100 - 900 (linear spacing, step: 100) \\
     & n estimators ranges & 200 - 1000 (linear spacing, step: 100) \\
     & subsample & 0.55, 0.6, 0.70, 0.85 \\
    \rule{0pt}{2em}
    \multirow{4}{*}{\begin{tabular}[c]{@{}r@{}}LR, \\ MLP, \\ CNN, \\ ConvLSTM, \\ ConvTransformer\end{tabular}} & learning rate & $10^{-6}, 10^{-5}, 10^{-4}, 10^{-3}, 10^{-2}$ \\
     & weight decay & $10^{-7}, 10^{-6}, 10^{-5}, 10^{-4}, 10^{-3}, 10^{-2}, 10^{-1}$ \\
     & batch size & 32 - 128 (linear spacing, step: 32) \\
     & optimiser & ``Adam'', ``AdamW'', ``SGD'' \\
    \rule{0pt}{2em}
    \multirow{5}{*}{\begin{tabular}[c]{@{}r@{}}CNN, \\ ConvLSTM, \\ ConvTransformer\end{tabular}} & convolution layer 1 & $2^{8}, 2^{9}, 2^{10}$ \\
     & convolution layer 2 & $2^{7}, 2^{8}$ \\
     & convolution layer 3 & $2^{6}, 2^{7}$ \\
     & convolution kernel 2 & 3, 4, 5 \\
     & max pool & 2 \\
    \rule{0pt}{2em}
    \multirow{3}{*}{MLP} & fully-connected layer 1 & $2^{8}, 2^{8}, 2^{10}$ \\
     & fully-connected layer 2 & $2^{7}, 2^{8}$ \\
     & fully-connected layer 3 & $2^{6}, 2^{7}$ \\
    \rule{0pt}{2em}
    CNN & fully-connected layer & $2^{5}, 2^{6}, 2^{7}, 2^{8}$ \\
    \rule{0pt}{2em}
    \multirow{4}{*}{ConvLSTM} & LSTM layer & $2^{6}, 2^{7}$ \\
     & number of LSTM layers & 2, 3, 4, 5 \\
     & fully-connected layer 1 & $2^{6}, 2^{7}$ \\
     & fully-connected layer 2 & $2^{5}, 2^{6}$ \\
    \rule{0pt}{2em}
    \multirow{5}{*}{ConvTransformer} & vocabulary size & 1300, 2100, 3400, 5500, 8900 \\
     & model dimension & 100, 200, 300, 500, 800 \\
     & number of heads & 2, 3, 4, 5, 6 \\
     & number of encoder layers & 1, 2, 3, 4, 5, 6 \\
     & fully-connected layer & $2^{7}, 2^{8}, 2^{9}, 2^{10}$ \\
\end{tabular}
\end{table}

\begin{table}[!h]
\caption{The average improvement in the metric after post-processing, with the p-value indicating whether the difference is statistically significant. Asterisks denote significance level: * $p<0.05$; ** $p<0.01$; *** $p<0.001$. Values are aggregated for each metric across all folds and all models. The arrow symbol ($\rightarrow$) denotes that models were trained on the dataset indicated before the arrow and evaluated on the dataset indicated after the arrow.}
\label{tab:post-processing-all}
\setlength{\tabcolsep}{3pt}
\begin{tabular}{r|lS[table-format=-1.1e-1]l}
    & \textbf{Dataset} & \textbf{Mean improvement} &\textbf{ p-value} \\
    \hline\\
    \multirow{4}{*}{Accuracy} & CHB-MIT $\rightarrow$ CHB-MIT & 2.30e-02 & 3.32e-05$^{***}$ \\
     & CHB-MIT $\rightarrow$ TUSZ & 1.59e-02 & 6.47e-05$^{***}$ \\
     & TUSZ $\rightarrow$ CHB-MIT & 4.22e-02 & 1.19e-07$^{***}$ \\
     & TUSZ $\rightarrow$ TUSZ & 1.89e-02 & 3.86e-05$^{***}$ \\
     \rule{0pt}{2em}
    \multirow{4}{*}{F1} & CHB-MIT $\rightarrow$ CHB-MIT & 3.33e-02 & 7.83e-05$^{***}$ \\
     & CHB-MIT $\rightarrow$ TUSZ & 1.54e-03 & 9.29e-01 \\
     & TUSZ $\rightarrow$ CHB-MIT & 2.25e-02 & 1.19e-07$^{***}$ \\
     & TUSZ $\rightarrow$ TUSZ & 1.27e-02 & 2.71e-03$^{**}$ \\
     \rule{0pt}{2em}
    \multirow{4}{*}{MSE} & CHB-MIT $\rightarrow$ CHB-MIT & -1.21e-02 & 3.32e-05$^{***}$ \\
     & CHB-MIT $\rightarrow$ TUSZ & -5.91e-03 & 9.15e-05$^{***}$ \\
     & TUSZ $\rightarrow$ CHB-MIT & -2.35e-02 & 1.19e-07$^{***}$ \\
     & TUSZ $\rightarrow$ TUSZ & -6.47e-03 & 6.66e-04$^{***}$ \\
     \rule{0pt}{2em}
    \multirow{4}{*}{PRAUC} & CHB-MIT $\rightarrow$ CHB-MIT & 3.19e-02 & 2.24e-04$^{***}$ \\
     & CHB-MIT $\rightarrow$ TUSZ & 7.49e-03 & 7.81e-03$^{**}$ \\
     & TUSZ $\rightarrow$ CHB-MIT & 2.07e-03 & 5.06e-03$^{**}$ \\
     & TUSZ $\rightarrow$ TUSZ & 1.04e-02 & 2.39e-03$^{**}$ \\
     \rule{0pt}{2em}
    \multirow{4}{*}{Precision} & CHB-MIT $\rightarrow$ CHB-MIT & 2.93e-02 & 6.72e-05$^{***}$ \\
     & CHB-MIT $\rightarrow$ TUSZ & 1.89e-02 & 2.60e-04$^{***}$ \\
     & TUSZ $\rightarrow$ CHB-MIT & 4.54e-03 & 3.32e-05$^{***}$ \\
     & TUSZ $\rightarrow$ TUSZ & 3.05e-02 & 3.32e-05$^{***}$ \\
     \rule{0pt}{2em}
    \multirow{4}{*}{ROC} & CHB-MIT $\rightarrow$ CHB-MIT & 7.30e-04 & 8.70e-02 \\
     & CHB-MIT $\rightarrow$ TUSZ & -1.27e-03 & 3.92e-01 \\
     & TUSZ $\rightarrow$ CHB-MIT & -3.63e-03 & 8.56e-01 \\
     & TUSZ $\rightarrow$ TUSZ & -7.63e-03 & 4.08e-02 \\
     \rule{0pt}{2em}
    \multirow{4}{*}{Sensitivity} & CHB-MIT $\rightarrow$ CHB-MIT & 2.30e-02 & 3.32e-05$^{***}$ \\
     & CHB-MIT $\rightarrow$ TUSZ & 1.59e-02 & 6.47e-05$^{***}$ \\
     & TUSZ $\rightarrow$ CHB-MIT & 4.22e-02 & 1.19e-07$^{***}$ \\
     & TUSZ $\rightarrow$ TUSZ & 1.89e-02 & 3.86e-05$^{***}$ \\
     \rule{0pt}{2em}
    \multirow{4}{*}{Specificity} & CHB-MIT $\rightarrow$ CHB-MIT & 9.66e-03 & 7.20e-02 \\
     & CHB-MIT $\rightarrow$ TUSZ & 2.65e-03 & 2.62e-01 \\
     & TUSZ $\rightarrow$ CHB-MIT & 1.30e-02 & 2.39e-03$^{**}$ \\
     & TUSZ $\rightarrow$ TUSZ & 4.67e-03 & 1.22e-01
\end{tabular}
\end{table}

\begin{table}[h]
\caption{
  A complete list of available features, broken down by type and channel scope. Legend:
  $\mathcal{T}$ = Temporal,\ 
  $\rho$ = Connectivity,\ 
  $\mathcal{G}$ = Graph Theory Derived,\ 
  $\delta,\theta,\alpha,\beta,\gamma$ = Frequency bands (\textit{delta, theta, alpha, beta, gamma});\ 
  $\bullet$ = Single channel,\ 
  $\bullet\bullet$ = Channel pair,\ 
  $\bigodot$ = All channels.
}
\label{tab:features-list}
\setlength{\tabcolsep}{3pt}
\begin{tabular}{lcc}
    \multicolumn{1}{c}{\textbf{Name}} &
    \multicolumn{1}{c}{\textbf{Type}} &
    \multicolumn{1}{c}{\textbf{Channels}} \\
    \hline\\
    \textbf{Mean} & $\mathcal{T}$ & $\bullet$ \\
    \textbf{Variance} & $\mathcal{T}$ & $\bullet$ \\
    \textbf{Skewness} & $\mathcal{T}$ & $\bullet$ \\
    \textbf{Kurtosis} & $\mathcal{T}$ & $\bullet$ \\
    \textbf{Interquartile range} & $\mathcal{T}$ & $\bullet$ \\
    \textbf{Min} & $\mathcal{T}$ & $\bullet$ \\
    \textbf{Max} & $\mathcal{T}$ & $\bullet$ \\
    \textbf{Hjorth complexity} & $\mathcal{T}$ & $\bullet$ \\
    \textbf{Hjorth mobility} & $\mathcal{T}$ & $\bullet$ \\
    \textbf{Petrosian fractal dimension} & $\mathcal{T}$ & $\bullet$ \\
    \textbf{Intermittency} & $\mathcal{T}$ & $\bullet$ \\
    \textbf{Voltage auc} & $\mathcal{T}$ & $\bullet$ \\
    \textbf{Spikiness} & $\mathcal{T}$ & $\bullet$ \\
    \textbf{Standard deviation} & $\mathcal{T}$ & $\bullet$ \\
    \textbf{Zero crossing} & $\mathcal{T}$ & $\bullet$ \\
    \textbf{Peak to peak} & $\mathcal{T}$ & $\bullet$ \\
    \textbf{Absolute area under signal} & $\mathcal{T}$ & $\bullet$ \\
    \textbf{Total signal energy} & $\mathcal{T}$ & $\bullet$ \\
    \textbf{Spike count} & $\mathcal{T}$ & $\bullet$ \\
    \textbf{Coastline} & $\mathcal{T}$, $\delta,\theta,\alpha,\beta,\gamma$ & $\bullet$ \\
    \textbf{Power spectral density} & $\delta,\theta,\alpha,\beta,\gamma$ & $\bullet$ \\
    \textbf{Power spectral centroid} & $\delta,\theta,\alpha,\beta,\gamma$ & $\bullet$ \\
    \textbf{Signal monotony} & $\delta,\theta,\alpha,\beta,\gamma$ & $\bullet$ \\
    \textbf{Signal to noise} & $\delta,\theta,\alpha,\beta,\gamma$ & $\bullet$ \\
    \textbf{Energy percentage} & $\delta,\theta,\alpha,\beta,\gamma$ & $\bullet$ \\
    \textbf{Discrete wavelet transform} & $\delta,\theta,\alpha,\beta,\gamma$ & $\bullet$ \\
    \textbf{Cross correlation max coef} & $\rho$ & $\bullet\bullet$ \\
    \textbf{Coherence} & $\rho$ & $\bullet\bullet$ \\
    \textbf{Imaginary coherence} & $\rho$ & $\bullet\bullet$ \\
    \textbf{Phase slope index} & $\rho$ & $\bullet\bullet$ \\
    \textbf{Eccentricity} & $\mathcal{G}$ & $\bigodot$ \\
    \textbf{Clustering coefficient} & $\mathcal{G}$ & $\bigodot$ \\
    \textbf{Betweenness centrality} & $\mathcal{G}$ & $\bigodot$ \\
    \textbf{Local efficiency} & $\mathcal{G}$ & $\bigodot$ \\
    \textbf{Global efficiency} & $\mathcal{G}$ & $\bigodot$ \\
    \textbf{Diameter} & $\mathcal{G}$ & $\bigodot$ \\
    \textbf{Radius} & $\mathcal{G}$ & $\bigodot$ \\
    \textbf{Characteristic path} & $\mathcal{G}$ & $\bigodot$ \\
\end{tabular}
\end{table}

\begin{landscape}
\begin{tiny}
    \begin{table}[h!]
    \caption{Average performance of proposed models on \Acrfull{chb-mit} dataset.}
    \label{tab:supplementary-chb-results}
    \setlength{\tabcolsep}{2pt}
    \begin{tabular}{r|cccccccc}
     &
      \multicolumn{1}{c}{\textbf{MSE}} &
      \multicolumn{1}{c}{\textbf{ROC}} &
      \multicolumn{1}{c}{\textbf{Accuracy}} &
      \multicolumn{1}{c}{\textbf{Precision}} &
      \multicolumn{1}{c}{\textbf{F1}} &
      \multicolumn{1}{c}{\textbf{PRAUC}} &
      \multicolumn{1}{c}{\textbf{Sensitivity}} &
      \multicolumn{1}{c}{\textbf{Specificity}} \\
        \hline\rule{0pt}{1em}
        \textbf{LR} & 0.1345 \!\!$\pm$\!\! 0.0610 & 0.8675 \!\!$\pm$\!\! 0.0913 & 0.8163 \!\!$\pm$\!\! 0.0907 & 0.5391 \!$\pm$\! 0.0245 & 0.5167 \!$\pm$\! 0.0642 & 0.3522 \!$\pm$\! 0.0407 & 0.8163 \!$\pm$\! 0.0907 & 0.7990 \!$\pm$\! 0.0471 \\
        \textbf{XGB} & 0.0297 \!$\pm$\! 0.0102 & 0.8769 \!$\pm$\! 0.0395 & 0.9695 \!$\pm$\! 0.0114 & 0.6490 \!$\pm$\! 0.0291 & 0.7070 \!$\pm$\! 0.0392 & 0.5462 \!$\pm$\! 0.0549 & 0.9695 \!$\pm$\! 0.0114 & 0.8605 \!$\pm$\! 0.0627 \\
        \textbf{MLP} & 0.0898 \!$\pm$\! 0.0204 & 0.9266 \!$\pm$\! 0.0536 & 0.8906 \!$\pm$\! 0.0260 & 0.5505 \!$\pm$\! 0.0103 & 0.5624 \!$\pm$\! 0.0242 & 0.4534 \!$\pm$\! 0.0382 & 0.8906 \!$\pm$\! 0.0260 & 0.8574 \!$\pm$\! 0.0640 \\
        \textbf{CNN} & 0.1317 \!$\pm$\! 0.1498 & 0.8861 \!$\pm$\! 0.0574 & 0.8025 \!$\pm$\! 0.2322 & 0.5246 \!$\pm$\! 0.0280 & 0.4784 \!$\pm$\! 0.1154 & 0.1078 \!$\pm$\! 0.0883 & 0.8025 \!$\pm$\! 0.2322 & 0.7655 \!$\pm$\! 0.0789 \\
        \textbf{EEGNet} & 0.0796 \!$\pm$\! 0.0799 & 0.8059 \!$\pm$\! 0.0654 & 0.9023 \!$\pm$\! 0.1018 & 0.5257 \!$\pm$\! 0.0306 & 0.5155 \!$\pm$\! 0.0691 & 0.1882 \!$\pm$\! 0.1554 & 0.9023 \!$\pm$\! 0.1018 & 0.7245 \!$\pm$\! 0.0660 \\
        \textbf{ConvLSTM} & 0.0585 \!$\pm$\! 0.0531 & 0.9067 \!$\pm$\! 0.0414 & 0.9286 \!$\pm$\! 0.0694 & 0.5401 \!$\pm$\! 0.0503 & 0.5432 \!$\pm$\! 0.0873 & 0.2447 \!$\pm$\! 0.1736 & 0.9286 \!$\pm$\! 0.0694 & 0.7570 \!$\pm$\! 0.0984 \\
        \textbf{ConvTransformer} & 0.1069 \!$\pm$\! 0.0926 & 0.7848 \!$\pm$\! 0.0557 & 0.8306 \!$\pm$\! 0.1691 & 0.5117 \!$\pm$\! 0.0132 & 0.4708 \!$\pm$\! 0.0712 & 0.0513 \!$\pm$\! 0.0398 & 0.8306 \!$\pm$\! 0.1691 & 0.6823 \!$\pm$\! 0.0149 \\
        \textbf{Binary voting} & 0.0316 \!$\pm$\! 0.0235 & 0.5829 \!$\pm$\! 0.0530 & 0.9684 \!$\pm$\! 0.0235 & 0.6914 \!$\pm$\! 0.1414 & 0.5763 \!$\pm$\! 0.0775 & 0.2966 \!$\pm$\! 0.1480 & 0.9684 \!$\pm$\! 0.0235 & 0.5829 \!$\pm$\! 0.0530 \\
        \textbf{Mean voting} & 0.0382 \!$\pm$\! 0.0258 & 0.9044 \!$\pm$\! 0.0586 & 0.9752 \!$\pm$\! 0.0142 & 0.7139 \!$\pm$\! 0.1243 & 0.6034 \!$\pm$\! 0.0739 & 0.3802 \!$\pm$\! 0.1874 & 0.9752 \!$\pm$\! 0.0142 & 0.6309 \!$\pm$\! 0.0998    
        \end{tabular}
    \end{table}
    
    \begin{table}[h!]
    \caption{Average performance of proposed models on \Acrfull{tusz} dataset.}
    \label{tab:supplementary-tusz-results}
    \setlength{\tabcolsep}{2pt}
    \begin{tabular}{r|cccccccc}
     &
      \multicolumn{1}{c}{\textbf{MSE}} &
      \multicolumn{1}{c}{\textbf{ROC}} &
      \multicolumn{1}{c}{\textbf{Accuracy}} &
      \multicolumn{1}{c}{\textbf{Precision}} &
      \multicolumn{1}{c}{\textbf{F1}} &
      \multicolumn{1}{c}{\textbf{PRAUC}} &
      \multicolumn{1}{c}{\textbf{Sensitivity}} &
      \multicolumn{1}{c}{\textbf{Specificity}} \\
    \hline\rule{0pt}{1em}
    \textbf{LR} & 0.2314 \!$\pm$\! 0.0274 & 0.7132 \!$\pm$\! 0.0502 & 0.6517 \!$\pm$\! 0.0386 & 0.6138 \!$\pm$\! 0.0286 & 0.6038 \!$\pm$\! 0.0350 & 0.4153 \!$\pm$\! 0.0739 & 0.6517 \!$\pm$\! 0.0386 & 0.6574 \!$\pm$\! 0.0404 \\
    \textbf{XGB} & 0.1621 \!$\pm$\! 0.0357 & 0.7602 \!$\pm$\! 0.0415 & 0.8180 \!$\pm$\! 0.0330 & 0.7543 \!$\pm$\! 0.0561 & 0.7338 \!$\pm$\! 0.0240 & 0.6354 \!$\pm$\! 0.0636 & 0.8180 \!$\pm$\! 0.0330 & 0.7250 \!$\pm$\! 0.0084 \\
    \textbf{MLP} & 0.1802 \!$\pm$\! 0.0434 & 0.7532 \!$\pm$\! 0.1048 & 0.7282 \!$\pm$\! 0.0727 & 0.6521 \!$\pm$\! 0.0735 & 0.6581 \!$\pm$\! 0.0826 & 0.5202 \!$\pm$\! 0.1748 & 0.7282 \!$\pm$\! 0.0727 & 0.6866 \!$\pm$\! 0.0911 \\
    \textbf{CNN} & 0.1600 \!$\pm$\! 0.0287 & 0.7952 \!$\pm$\! 0.0668 & 0.7908 \!$\pm$\! 0.0454 & 0.7041 \!$\pm$\! 0.0555 & 0.6885 \!$\pm$\! 0.0753 & 0.5555 \!$\pm$\! 0.1152 & 0.7908 \!$\pm$\! 0.0454 & 0.6923 \!$\pm$\! 0.0832 \\
    \textbf{EEGNet} & 0.2325 \!$\pm$\! 0.0953 & 0.7067 \!$\pm$\! 0.0675 & 0.6775 \!$\pm$\! 0.1760 & 0.6346 \!$\pm$\! 0.0649 & 0.5149 \!$\pm$\! 0.0952 & 0.4060 \!$\pm$\! 0.0631 & 0.6775 \!$\pm$\! 0.1760 & 0.5683 \!$\pm$\! 0.0478 \\
    \textbf{ConvLSTM} & 0.1137 \!$\pm$\! 0.0274 & 0.8594 \!$\pm$\! 0.0445 & 0.8549 \!$\pm$\! 0.0342 & 0.8345 \!$\pm$\! 0.0228 & 0.7624 \!$\pm$\! 0.0498 & 0.7430 \!$\pm$\! 0.0434 & 0.8549 \!$\pm$\! 0.0342 & 0.7365 \!$\pm$\! 0.0486 \\
    \textbf{ConvTransformer} & 0.1308 \!$\pm$\! 0.0323 & 0.8265 \!$\pm$\! 0.0594 & 0.8287 \!$\pm$\! 0.0352 & 0.7993 \!$\pm$\! 0.0369 & 0.7021 \!$\pm$\! 0.0413 & 0.6519 \!$\pm$\! 0.0460 & 0.8287 \!$\pm$\! 0.0352 & 0.6749 \!$\pm$\! 0.0344 \\
    \textbf{Binary voting} & 0.1582 \!$\pm$\! 0.0464 & 0.7313 \!$\pm$\! 0.0420 & 0.8418 \!$\pm$\! 0.0464 & 0.7953 \!$\pm$\! 0.0615 & 0.7547 \!$\pm$\! 0.0496 & 0.6799 \!$\pm$\! 0.0629 & 0.8418 \!$\pm$\! 0.0464 & 0.7313 \!$\pm$\! 0.0420 \\
    \textbf{Mean voting} & 0.1217 \!$\pm$\! 0.0228 & 0.8638 \!$\pm$\! 0.0603 & 0.8531 \!$\pm$\! 0.0425 & 0.8200 \!$\pm$\! 0.0569 & 0.7688 \!$\pm$\! 0.0443 & 0.7341 \!$\pm$\! 0.0738 & 0.8531 \!$\pm$\! 0.0425 & 0.7413 \!$\pm$\! 0.0378    
    \end{tabular}
    \end{table}
\end{tiny}
\end{landscape}

\begin{landscape}
\begin{tiny}
    \begin{table}[h!]
    \caption{The average performance of proposed models trained on \Acrfull{chb-mit} and tested on \Acrfull{tusz} dataset.}
    \label{tab:supplementary-chb-on-tusz-results}
    \setlength{\tabcolsep}{2pt}
    \begin{tabular}{r|cccccccc}
     &
      \multicolumn{1}{c}{\textbf{MSE}} &
      \multicolumn{1}{c}{\textbf{ROC}} &
      \multicolumn{1}{c}{\textbf{Accuracy}} &
      \multicolumn{1}{c}{\textbf{Precision}} &
      \multicolumn{1}{c}{\textbf{F1}} &
      \multicolumn{1}{c}{\textbf{PRAUC}} &
      \multicolumn{1}{c}{\textbf{Sensitivity}} &
      \multicolumn{1}{c}{\textbf{Specificity}} \\
    \hline\rule{0pt}{1em}
    \textbf{LR} & 0.2368 \!$\pm$\! 0.0288 & 0.5895 \!$\pm$\! 0.0074 & 0.6590 \!$\pm$\! 0.0791 & 0.5577 \!$\pm$\! 0.0142 & 0.5285 \!$\pm$\! 0.0158 & 0.2830 \!$\pm$\! 0.0467 & 0.6590 \!$\pm$\! 0.0791 & 0.5511 \!$\pm$\! 0.0129 \\
    \textbf{XGB} & 0.2212 \!$\pm$\! 0.0412 & 0.5510 \!$\pm$\! 0.0417 & 0.7660 \!$\pm$\! 0.0267 & 0.6196 \!$\pm$\! 0.0100 & 0.4928 \!$\pm$\! 0.0268 & 0.3557 \!$\pm$\! 0.0477 & 0.7660 \!$\pm$\! 0.0267 & 0.5231 \!$\pm$\! 0.0080 \\
    \textbf{MLP} & 0.2282 \!$\pm$\! 0.0250 & 0.6171 \!$\pm$\! 0.0088 & 0.6930 \!$\pm$\! 0.0457 & 0.5825 \!$\pm$\! 0.0279 & 0.5502 \!$\pm$\! 0.0109 & 0.3132 \!$\pm$\! 0.0489 & 0.6930 \!$\pm$\! 0.0457 & 0.5648 \!$\pm$\! 0.0162 \\
    \textbf{CNN} & 0.2687 \!$\pm$\! 0.0692 & 0.6059 \!$\pm$\! 0.0837 & 0.6704 \!$\pm$\! 0.0835 & 0.5532 \!$\pm$\! 0.0213 & 0.5080 \!$\pm$\! 0.0164 & 0.2963 \!$\pm$\! 0.0389 & 0.6704 \!$\pm$\! 0.0835 & 0.5383 \!$\pm$\! 0.0287 \\
    \textbf{EEGNet} & 0.2317 \!$\pm$\! 0.0293 & 0.5125 \!$\pm$\! 0.0426 & 0.7609 \!$\pm$\! 0.0307 & 0.5162 \!$\pm$\! 0.0730 & 0.4507 \!$\pm$\! 0.0183 & 0.2544 \!$\pm$\! 0.0087 & 0.7609 \!$\pm$\! 0.0307 & 0.5015 \!$\pm$\! 0.0057 \\
    \textbf{ConvLSTM} & 0.2339 \!$\pm$\! 0.0355 & 0.6157 \!$\pm$\! 0.0681 & 0.7118 \!$\pm$\! 0.0690 & 0.6183 \!$\pm$\! 0.1044 & 0.5172 \!$\pm$\! 0.0618 & 0.3186 \!$\pm$\! 0.0270 & 0.7118 \!$\pm$\! 0.0690 & 0.5372 \!$\pm$\! 0.0475 \\
    \textbf{ConvTransformer} & 0.2845 \!$\pm$\! 0.0666 & 0.5660 \!$\pm$\! 0.0729 & 0.6233 \!$\pm$\! 0.1209 & 0.5065 \!$\pm$\! 0.0079 & 0.4668 \!$\pm$\! 0.0233 & 0.2670 \!$\pm$\! 0.0411 & 0.6233 \!$\pm$\! 0.1209 & 0.5021 \!$\pm$\! 0.0026 \\
    \textbf{Binary voting} & 0.2448 \!$\pm$\! 0.0256 & 0.5257 \!$\pm$\! 0.0200 & 0.7552 \!$\pm$\! 0.0256 & 0.5720 \!$\pm$\! 0.0168 & 0.5004 \!$\pm$\! 0.0404 & 0.3357 \!$\pm$\! 0.0444 & 0.7552 \!$\pm$\! 0.0256 & 0.5257 \!$\pm$\! 0.0200 \\
    \textbf{Mean voting} & 0.1841 \!$\pm$\! 0.0249 & 0.6148 \!$\pm$\! 0.0398 & 0.7562 \!$\pm$\! 0.0274 & 0.5734 \!$\pm$\! 0.0149 & 0.4979 \!$\pm$\! 0.0410 & 0.3096 \!$\pm$\! 0.0100 & 0.7562 \!$\pm$\! 0.0274 & 0.5244 \!$\pm$\! 0.0200
    \end{tabular}
    \end{table}
\end{tiny}
\begin{tiny}
    \begin{table}[h!]
    \caption{The average performance of proposed models trained on \Acrfull{tusz} and tested on\Acrfull{chb-mit} dataset.}
    \label{tab:supplementary-tusz-on-chb-results}
    \setlength{\tabcolsep}{2pt}
    \begin{tabular}{r|cccccccc}
     &
      \multicolumn{1}{c}{\textbf{MSE}} &
      \multicolumn{1}{c}{\textbf{ROC}} &
      \multicolumn{1}{c}{\textbf{Accuracy}} &
      \multicolumn{1}{c}{\textbf{Precision}} &
      \multicolumn{1}{c}{\textbf{F1}} &
      \multicolumn{1}{c}{\textbf{PRAUC}} &
      \multicolumn{1}{c}{\textbf{Sensitivity}} &
      \multicolumn{1}{c}{\textbf{Specificity}} \\
    \hline\rule{0pt}{1em}
    \textbf{LR} & 0.2440 \!$\pm$\! 0.0703 & 0.7758 \!$\pm$\! 0.1120 & 0.6418 \!$\pm$\! 0.1082 & 0.5130 \!$\pm$\! 0.0063 & 0.4177 \!$\pm$\! 0.0500 & 0.1378 \!$\pm$\! 0.0551 & 0.6418 \!$\pm$\! 0.1082 & 0.7029 \!$\pm$\! 0.0951 \\
    \textbf{XGB} & 0.4447 \!$\pm$\! 0.1103 & 0.5931 \!$\pm$\! 0.1621 & 0.4781 \!$\pm$\! 0.1547 & 0.5045 \!$\pm$\! 0.0094 & 0.3344 \!$\pm$\! 0.0758 & 0.3008 \!$\pm$\! 0.2043 & 0.4781 \!$\pm$\! 0.1547 & 0.6061 \!$\pm$\! 0.1440 \\
    \textbf{MLP} & 0.2372 \!$\pm$\! 0.0354 & 0.7724 \!$\pm$\! 0.1527 & 0.6070 \!$\pm$\! 0.1075 & 0.5116 \!$\pm$\! 0.0071 & 0.4023 \!$\pm$\! 0.0492 & 0.1186 \!$\pm$\! 0.0397 & 0.6070 \!$\pm$\! 0.1075 & 0.6979 \!$\pm$\! 0.1183 \\
    \textbf{CNN} & 0.1346 \!$\pm$\! 0.1377 & 0.7686 \!$\pm$\! 0.1029 & 0.7822 \!$\pm$\! 0.2513 & 0.5148 \!$\pm$\! 0.0109 & 0.4536 \!$\pm$\! 0.1077 & 0.0733 \!$\pm$\! 0.0534 & 0.7822 \!$\pm$\! 0.2513 & 0.6512 \!$\pm$\! 0.0229 \\
    \textbf{EEGNet} & 0.6912 \!$\pm$\! 0.1986 & 0.4219 \!$\pm$\! 0.1970 & 0.0937 \!$\pm$\! 0.0671 & 0.4960 \!$\pm$\! 0.0036 & 0.0829 \!$\pm$\! 0.0581 & 0.0238 \!$\pm$\! 0.0305 & 0.0937 \!$\pm$\! 0.0671 & 0.4220 \!$\pm$\! 0.0915 \\
    \textbf{ConvLSTM} & 0.0983 \!$\pm$\! 0.1061 & 0.7902 \!$\pm$\! 0.1182 & 0.8838 \!$\pm$\! 0.1331 & 0.5154 \!$\pm$\! 0.0116 & 0.4943 \!$\pm$\! 0.0574 & 0.0411 \!$\pm$\! 0.0227 & 0.8838 \!$\pm$\! 0.1331 & 0.6665 \!$\pm$\! 0.0392 \\
    \textbf{ConvTransformer} & 0.0804 \!$\pm$\! 0.0861 & 0.8650 \!$\pm$\! 0.1006 & 0.8842 \!$\pm$\! 0.1292 & 0.5175 \!$\pm$\! 0.0148 & 0.4988 \!$\pm$\! 0.0600 & 0.0428 \!$\pm$\! 0.0173 & 0.8842 \!$\pm$\! 0.1292 & 0.6760 \!$\pm$\! 0.0446 \\
    \textbf{Binary voting} & 0.3440 \!$\pm$\! 0.1390 & 0.6639 \!$\pm$\! 0.1149 & 0.6560 \!$\pm$\! 0.1390 & 0.5102 \!$\pm$\! 0.0076 & 0.4176 \!$\pm$\! 0.0619 & 0.3527 \!$\pm$\! 0.1318 & 0.6560 \!$\pm$\! 0.1390 & 0.6639 \!$\pm$\! 0.1149 \\
    \textbf{Mean voting} & 0.1833 \!$\pm$\! 0.0421 & 0.7618 \!$\pm$\! 0.1753 & 0.7875 \!$\pm$\! 0.1320 & 0.5190 \!$\pm$\! 0.0146 & 0.4783 \!$\pm$\! 0.0597 & 0.1383 \!$\pm$\! 0.1232 & 0.7875 \!$\pm$\! 0.1320 & 0.6997 \!$\pm$\! 0.1246
    \end{tabular}
    \end{table}
\end{tiny}
\end{landscape}
\clearpage

\end{appendices}

\end{document}